\documentclass[10pt]{article} 
\usepackage[accepted]{tmlr}

\usepackage{url}

\usepackage[utf8]{inputenc} 
\usepackage[T1]{fontenc}    
\usepackage[linkcolor=blue,allcolors=blue]{hyperref}
\usepackage{url}            
\usepackage{booktabs}       
\usepackage{amsfonts}       
\usepackage{nicefrac}       
\usepackage{microtype,nicefrac}      
\usepackage{xspace}
\usepackage{natbib}
\usepackage{pifont}

\usepackage{comment}
\usepackage{amsmath}
\usepackage{graphicx}
\usepackage{rotating}

\usepackage{amssymb}
\usepackage{autobreak}
\usepackage{mathtools}
\usepackage{xcolor}
\usepackage{bbm}
\usepackage{algorithm}
\usepackage[parfill]{parskip}
\usepackage{algpseudocode}
\usepackage[export]{adjustbox}
\usepackage{subcaption}
\usepackage[toc]{appendix}
\usepackage{minitoc}
\usepackage{wrapfig}
\newcommand{\cmark}{\text{\ding{51}}}%
\newcommand{\xmark}{\text{\ding{55}}}%
\makeatletter
\usepackage[flushleft]{threeparttable}
\newcommand{\G}{\mathcal{G}}

\newcommand{\alg}{\textsc{\small{OpFlow}}\xspace}

\newcommand{\U}{\mathcal{U}}

\newcommand{\F}{\mathcal F}

\newcommand{\A}{\mathcal A}

\newcommand{\wh}{\widehat}
\newcommand{\wb}{\overline}

\newtheorem{proof*}{Proof}[section]

\usepackage{xcolor}

\title{Universal Functional Regression with Neural Operator Flows}

\author{\name Yaozhong Shi \email yshi5@caltech.edu \\
      \addr California Institute of Technology
      \AND
      \name Angela F. Gao \email afgao@caltech.edu \\
      \addr California Institute of Technology
      \AND
      \name Zachary E. Ross \email zross@caltech.edu\\
      \addr California Institute of Technology
      \AND
      \name Kamyar Azizzadenesheli \email kamyara@nvidia.com\\
      \addr NVIDIA Corporation \\}

\def\month{10}  
\def\year{2024} 
\def\openreview{\url{https://openreview.net/forum?id=rHL329Xa3X}} 

\begin{document}

\maketitle

\begin{abstract}

Regression on function spaces is typically limited to models with Gaussian process priors. We introduce the notion of universal functional regression, in which we aim to learn a prior distribution over non-Gaussian function spaces that remains mathematically tractable for functional regression. To do this, we develop Neural Operator Flows (\alg), an infinite-dimensional extension of normalizing flows. \alg is an invertible operator that maps the (potentially unknown) data function space into a Gaussian process, allowing for exact likelihood estimation of functional point evaluations. \alg enables robust and accurate uncertainty quantification via drawing posterior samples of the Gaussian process and subsequently mapping them into the data function space. We empirically study the performance of \alg on regression and generation tasks with data generated from Gaussian processes with known posterior forms and non-Gaussian processes, as well as real-world earthquake seismograms with an unknown closed-form distribution. 

\end{abstract}

\section{Introduction}

The notion of inference on function spaces is essential to the physical sciences and engineering. In particular, it is often desirable to infer the values of a function everywhere in a physical domain given a sparse number of observation points. There are numerous types of problems in which functional regression plays an important role, such as inverse problems, time series forecasting, data imputation/assimilation. Functional regression problems can be particularly challenging for real world datasets because the underlying stochastic process is often unknown.

\begin{wrapfigure}{r}{.3\textwidth}
\vspace*{-.7cm}
    \begin{minipage}{0.45\linewidth}
        \includegraphics[height=.76\linewidth,trim={2cm 1cm 2cm 1.2cm},clip]{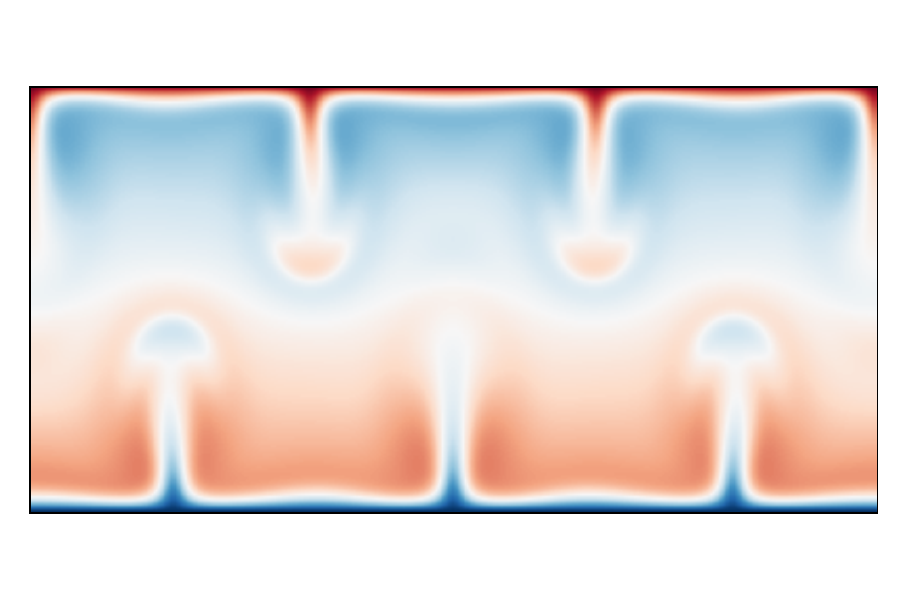}
        \vspace*{-.5cm}
        \caption*{(a)}
    \end{minipage}
    %
    \quad
    \begin{minipage}{0.45\linewidth}
        \includegraphics[height=.7\linewidth,trim={2cm 0.5cm 2cm 0.38cm},clip]{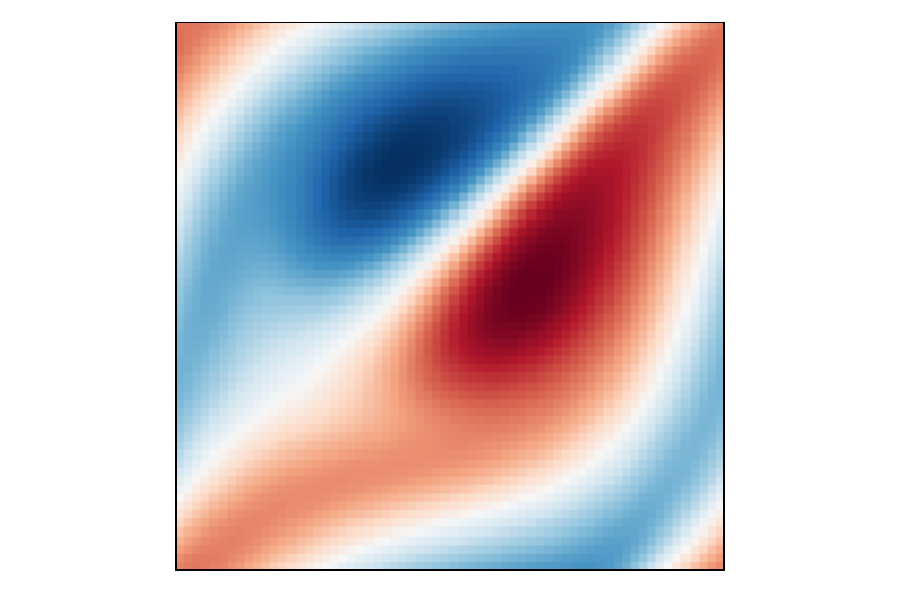}
        \caption*{(b)}
    \end{minipage}
    \caption{ Examples of data from non-Gaussian processes. (a) Temperature field from Rayleigh Bénard convection problem. (b) Vorticity field from Naiver-Stokes Equation.}
    \label{fig:non_gp_examples}
\end{wrapfigure}

Much of the work on functional regression and inference has relied on Gaussian processes (GPs) ~\citep{rasmussen_gaussian_2006}, a specific type of stochastic process in which any finite collection of points has a multivariate Gaussian distribution. Some of the earliest applications focused on analyzing geological data, such as the locations of valuable ore deposits, to identify where new deposits might be found~\citep{chiles_geostatistics_2012}.  GP regression (GPR) provides several advantages for functional inference including robustness and mathematical tractability for various problems. This has led to the use of GPR in an assortment of scientific and engineering fields, where precision and reliability in predictions and inferences can significantly impact outcomes~\citep{deringer_gaussian_2021, aigrain_gaussian_2023}. Despite widespread adoption, the assumption of a GP prior for functional inference problems can be rather limiting, particularly in scenarios where the data exhibit heavy-tailed or multimodal distributions, e.g. financial time series or environmental modeling. This underscores the need for models with greater expressiveness, allowing for regression on general function spaces, with data arising from potentially unknown stochastic processes and distributions (Fig. \ref{fig:non_gp_examples}). We refer to such a regression problem as universal functional regression (UFR).


We hypothesize that addressing the problem of UFR requires a model with two primary components. First, the model needs to be capable of learning priors over data that live in function spaces. Indeed, there has been much recent progress in machine learning on learning priors on function spaces~\citep{rahman_generative_2022,lim_score-based_2023,seidman_variational_2023,kerrigan_diffusion_2023,pidstrigach_infinite-dimensional_2023,baldassari_conditional_2023,hagemann_multilevel_2023,kerrigan_functional_2023,franzese_continuous-time_2023}. Second, the models need a framework for performing functional regression with the learned function space priors and learned likelihood--a component that is critically missing from the previous works on pure generative models described above. This includes the ability to compute the likelihood of an input data sample, which would allow for posterior estimation over the entire physical domain. Addressing these challenges not only expands the models available for functional regression but also enhances our capacity to extract meaningful insights from complex datasets. 

In this study, our contributions are as follows. We develop a viable formulation for UFR with a novel learnable bijective operator, \alg. This operator extends the classical normalizing flow, which maps between finite dimensional vector spaces, to the function space setting. \alg consists of an invertible neural operator, an invertible map between function spaces, trained to map the data point-process distribution to a known and easy-to-sample GP distribution. Given point value samples of a data function, \alg allows for computing the likelihood of the observed values, a principled property a priori held by GP. Using this property, we formally define the problem setting for UFR and develop an approach for posterior estimation with Stochastic Gradient Langevin Dynamics (SGLD)~\citep{welling_bayesian_2011}. Finally, we demonstrate UFR with \alg on a suite of experiments that use both Gaussian and non-Gaussian processes.

In comparison to GPR, \alg draws on the training dataset to learn the prior over a function space. When such datasets are only minimally available, kernel tuning methods propose to hand-tune the kernel and parameters of GPR, providing considerable performance, although still with a Gaussian assumption. In such limited data scenarios, learning accurate priors with \alg may be difficult and less suitable. When training data is non-existent, often neither the plain GPR nor \alg can be of help; when expert knowledge can be incorporated~\citep{aigrain_gaussian_2023}, there may be advantages to GPR. However, such a method can also be considered as hand-tuning the prior used in \alg by, for example, setting the invertible model to identity map, resembling strong prior over the prior learning process.

\section{Related work}
\textbf{Neural Operators.}
Neural Operator is a new paradigm in deep learning for learning maps between infinite-dimensional function spaces. Unlike traditional neural networks that primarily work with fixed-dimensional vectors, neural operators are designed to operate on functions, making them inherently suitable for a wide range of scientific computing and engineering tasks involving partial differential equations (PDEs)~\citep{kovachki_neural_2023, li_neural_2020}. A key feature of neural operators is their discretization agnostic (resolution invariant) property~\citep{kovachki_neural_2023}, which means that a neural operator can learn from data represented on various resolutions and predict outcomes on yet new resolutions. This property is particularly valuable in applications involving the natural sciences, PDEs, and complex physical simulations where functional data may come from variable discretizations of meshes or sensors. Within the family of neural operators, Fourier Neural Operator (FNO)~\citep{li_fourier_2021} stands out for its quasi-linear time complexity by defining the integral kernel in Fourier space and has been applied to many problems in engineering and science~\citep{azizzadenesheli_neural_2024}. In the conventional view of neural operators, provided discretization and point evaluation of functions resembles approximation of function for which increasing the resolution is connected to improved approximation error in the integration and differentiation operators~\citep{liu-schiaffini_neural_2024}. The proposed view in the current work provides a new perspective for which the point evaluations are instead considered as points samples of functions that are associated probably rather than finite bases and finite resolution approximation of a continuous function. Therefore, increasing the resolution is seen as providing more information for regression rather than a way to reduce, e.g., the approximation error in the integral operators used in neural operations. 

\textbf{Function space generative modeling with Neural Operators.} Several discretization invariant generative models on function spaces have been recently developed using Neural Operators~\citep{li_neural_2020, kovachki_neural_2023}. These models include Generative Adversarial Neural Operator (GANO)~\citep{rahman_generative_2022}, Denoising Diffusion models (DDO)~\citep{lim_score-based_2023} and Variational Autoencoding Neural Operator (VANO)~\citep{seidman_variational_2023}.  To be specific, GANO is the first proposed infinite-dimensional generative model that generalizes GAN, and learns probability measures on function spaces. However, GANO training relies on an adversarial game, which can be slow~\citep{goodfellow_generative_2014, rahman_generative_2022}; DDO is a resolution-invariant diffusion generative model for function spaces, leveraging denoising score matching within Hilbert spaces and employing infinite-dimensional Langevin dynamics for sampling. The inference of DDO requires a diffusion process, as with the finite-dimensional score-based model, which can be time-consuming~\citep{song_score-based_2021, lim_score-based_2023}; VANO offers a functional variational objective~\citep{kingma_auto-encoding_2022,seidman_variational_2023}, mapping input functional data to a finite-dimensional latent space through an encoder, which may be a sub-optimal formulation due to the fixed dimensionality of the latent space being decoupled from the discretization of the input function data.

All of these aforementioned models have exhibited superior performance over their finite-dimensional counterparts for functional data by directly learning mappings between function spaces, which show great potential in addressing the challenges in science and engineering problems~\citep{shi_development_2024,shi_broadband_2024,  yang_seismic_2021, azizzadenesheli_neural_2024}. Compared to these existing infinite-dimensional generative models, \alg shows advantages in fast training, fast inference, bijective architecture, and providing precise likelihood estimation as summarized in Table \ref{tab:opflow_comp}. These advantages of \alg can be explained by: (i) The training and inference processes of \alg only involve likelihood estimation, and don't require an adversarial game or a diffusion process. (ii) \alg directly generalizes normalizing flows to function space without finite-dimensional simplicity, and enables precise likelihood estimation for functional point evaluations on irregular grids. (iii) As an invertible operator, \alg can be useful for tasks that require the bijectivity between function spaces~\citep{zhou_neural_2023, furuya_globally_2023}.

\begin{table}[h]
\centering
\caption{Comparison of \alg with other infinite-dimensional generative models}
\begin{tabular}{lcccc}
\hline
Models & Fast training & Fast inference & Bijective architecture & Exact likelihood estimation\\ \hline
GANO & $\xmark$ & $\cmark$ & $\xmark$ & $\xmark$ \\
DDO & $\xmark$ & $\xmark$ & $\xmark$ & $\xmark$ \\
VANO & $\cmark$ & $\cmark$ & $\xmark$ &$\xmark$  \\
\alg [Ours] & $\cmark$ & $\cmark$ & $\cmark$ &$\cmark$  \\ \hline
\end{tabular}
\label{tab:opflow_comp}
\end{table}

\textbf{Neural Processes.} Neural Process (NP)~\citep{garnelo_neural_2018} leverages neural networks to address the constraints associated with GPs, particularly the computational demands and the restrictive Gaussian assumptions for priors and posteriors~\citep{dutordoir_neural_2022}. While NP aims to model functional distributions, several fundamental flaws suggest NP might not be a good candidate for learning function data~\citep{rahman_generative_2022, dupont_generative_2022}. First, NP lacks true generative capability; it focuses on maximizing the likelihood of observed data points through an amortized inference approach, neglecting the metric space of the data. This approach can misinterpret functions sampled at different resolutions as distinct functions, undermining the NP's ability to learn from diverse function representations. As pointed out in prior works~\citep{rahman_generative_2022}, if a dataset comprises mainly low-resolution functions alongside a single function with a somewhat high resolution, NP only tries to maximize the likelihood for the point evaluation of the high-resolution function during training and ignores information from all other lower-resolution functions. A detailed experiment to show the failure of NP for learning simple function data can be found in Appendix A.1 of~\citep{rahman_generative_2022} . What's more, NP relies on an encoder to map the input data pairs to finite-dimensional latent variables, which are assumed to have Gaussian distributions~\citep{garnelo_neural_2018}, and then project the finite-dimensional vector to an infinite-dimensional space; this approach results in diminished consistency at elevated resolutions, and thus disables NP to scale to large datasets~\citep{dupont_generative_2022}. The other fundamental limitation is the Bayesian framework of NP is also defined on a set of points, rather than functions themselves. This method results in the dilution of prior information as the number of evaluated points increases, further constraining NP's ability to generalize from function data.

\textbf{Normalizing Flows.} Normalizing flows are a class of flow-based finite-dimensional generative models, that usually composed of a sequence of invertible transformations~\citep{kingma_glow_2018,dinh_density_2017, chen_neural_2019}. By gradually transforming a simple probability distribution into a more complex target distribution with the invertible architecture, normalizing flows enable exact likelihood evaluation, direct sampling, and good interpretability~\citep{kobyzev_normalizing_2021} with applications from image and audio generations~\citep{kingma_glow_2018,ping_waveflow_2020}, reinforcement learning~\citep{mazoure_leveraging_2020} to Bayesian inference~\citep{gao_deepgem_2021,whang_composing_2021}. Despite the advantages of normalizing flows, they still face challenges with high-dimensional data due to computational complexity, memory demands, scalability, and cannot be directly used for zero-shot super-resolution~\citep{lugmayr_srflow_2020}. Moreover, traditional normalizing flows are defined on a finite-dimensional spaces, and constrained to evaluating likelihood on fixed-size, regular grids. This inherent limits their applications in Bayesian inverse problems, where the data may not conform to fixed and regular grids~\citep{ghanem_bayesian_2017}. These limitations underline the need for a new generation of normalizing flows designed on function space, providing a solution that could potentially transcend the limitations of traditional models.

\textbf{Nonparametric Bayesian and Gaussian Process Regression.} Nonparametric Bayesian models are defined on an infinite-dimensional parameter space, yet they can be evaluated using only a finite subset of parameter dimensions to explain observed data samples~\citep{orbanz_bayesian_2010}. These models provide a flexible regression framework by offering analytically tractable posteriors and enable adjusting their complexity by allowing the model size to grow when more observations are available. Further, nonparametric Bayesian leverages a rich theoretical foundation of parametric Bayesian modeling and can be evaluated using common inference techniques like Markov Chain Monte Carlo (MCMC) and variational inference, etc. Despite the great advantages of the nonparametric Bayesian models, such models heavily depend on the chosen prior, which could restrict their adaptability and scalability in real-world complex scenarios~\citep{barry_nonparametric_1986,wasserman_all_2006}. The emergence of learning-based Bayesian models, propelled by advancements in deep learning, offers a potential solution by marrying the flexibility and scalability of neural networks with Bayesian principles; these modern approaches learn directly from data rather than rely on the chosen priors.

Within the family of nonparmateric Bayesian models, GPR is distinguished for its widespread applications, characterized by its analytical posteriors within the Bayesian framework~\citep{rasmussen_gaussian_2006}. GPR offers a robust framework for capturing uncertainties in predictions given noisy observations. However, GPR assumes data can be modeled using a Gaussian process, both the prior and posterior should be Gaussian. Such assumption, while simplifying the analytical process, might not always align with the complex nature, thus the universal applicability of GPR is constrained by the diversity and complexity of real-world datasets.

\textbf{Deep and Warped Gaussian Processes.}
Deep GP stacks multiple layers of Gaussian process~\citep{damianou_deep_2013}, which enables modeling of highly nonlinear and potentially non-Gaussian data, as well as providing uncertainty quantification at each layer~\citep{jakkala_deep_2021,liu_when_2020,de_souza_thin_2024}. However, as~\cite{ustyuzhaninov_compositional_2020,havasi_inference_2018} have pointed out, achieving exact Bayesian inference in Deep GP is challenged by its inherent intractability, necessitating the use of simplifying approximations. Unfortunately, these approximations could lead to the layers of a Deep GP collapsing to near-deterministic transformations~\citep{ustyuzhaninov_compositional_2020}, significantly reducing the model's probabilistic nature and its ability to quantify uncertainty accurately.
Separately, Warped GP presents an alternative for modeling non-Gaussian data by either transforming the likelihood of the outputs of GP with nonlinear transformations~\citep{snelson_warped_2003} or transforming the prior. While Warped GP has shown efficacy in handling real-world non-Gaussian data~\citep{kou_sparse_2013}, the selection of an appropriate warping function is critical and demands extensive domain expertise, where an ill-suited warping function can severely degrade model performance~\citep{rios_compositionally-warped_2019}. Besides, similar to Deep GP, Warped GP also suffers from analytical intractability, often requiring the employment of variational inference methods for practical implementation~\citep{lazaro-gredilla_bayesian_2012}. Recently, inspired by the success of normalizing flows, \citep{maronas_transforming_2021} proposed Transformed GP, which utilizes a marginal normalizing flow to learn the non-Gaussian prior. This approach not only matches the capability of traditional Deep GP in modeling complex data, but is also computationally efficient. However, Transformed GP use, yet again, pointwise wrapping of GP point evaluation, resulting in marginal flows that perform diagonal mapping, limiting its applicability to learn more general stochastic process. Additionally, the inference of Transformed GP relies on sparse variational objective which is considered as a sub-optimal approach in terms of accuracy.

Regression with \alg addresses the limitations of existing Gaussian process models discussed above, and offer two key improvements over the Transformed GP framework~\citep{maronas_transforming_2021}. To be specific, the \alg enables modeling a more general stochastic process compared to the marginal flow which is based on pointwise operator. The regression with the \alg enables true Bayesian uncertainty quantification via drawing posterior samples of the Gaussian process with SGLD and subsequently mapping them into the data function space. Such regression formulation is valid in the sense of correctness, imposing no approximation in the formulation, which enables us to capture true Bayesian uncertainty. The mentioned conceptual advancements are empirically shown beneficial in this paper. In contrast, variational inference formulation often  relies on optimizing the Evidence Lower Bound (ELBO) and yields an approximate posterior with an approximation gap “in the formulation”, rather than the true posterior. Despite the benefit, the proposed regression framework with SGLD requires significantly more time for sampling compared to variational inference, where the latter is more practical for time-sensitive application. Last, we should note that the regression framework of \alg slightly differs from that of the generic Deep GP. In Deep GP or Transformed GP, the model is fitted to the entire training dataset. For example, when analyzing a dataset containing rainfall measurements over a ten-year period, a Deep GP or Transformed GP would be used to directly fit all the data, enabling predictions of rainfall at any point within these ten years. In contrast, the regression with \alg can be analogous to image inpainting or reconstruction from pixels with generative models in a Bayesian manner ~\citep{marinescu_bayesian_2021}. In the \alg framework, each data point in the training set represents a distinct function, potentially having its unique discretization. During training, the goal of \alg is to learn a bijective mapping between a Gaussian process and the data function space. Once trained, \alg will be frozen and serve as the prior, when provided with new observations (i.e., point evaluations of an unknown function drawn from the data function measure), the task is to reconstruct the entire function, aligning \alg closely with the traditional settings of standard GP regression. The regression setup with \alg is formally defined in the subsequent section.

\section{Preliminaries}

\textbf{Universal functional regression.} Consider a function space, denoted as $\mathcal{U}$, such that for any $u \in \mathcal{U}, u : \mathcal{D}_{\mathcal{U}} \rightarrow \mathbb{R}^{d_{\mathcal{U}}}$. Here, $\mathcal{U}$ represents the space of the data. Let $\mathbb{P}_{\mathcal{U}}$ be a probability measure defined on $\mathcal{U}$. A finite collection of observations $u|_D$ is a set of point evaluations of function $u$ on a few points in its domain $D \subset \mathcal{D}_{\mathcal{U}}$, where $D$ is the collection of co-location positions $\{x_i\}_{i=1}^{l}, x_i \in \mathcal{D}_{\mathcal{U}}$. Note that in practice, $u \in \mathcal{U}$ is usually not observed in its entirety over $\mathcal{D}_{\mathcal{U}}$, but instead is discretized on a mesh. In UFR, independent and identically distributed samples of $\{u_j\}_{j=1}^{m} \sim \mathbb{P}_{\mathcal{U}}$ are accessed on the discretization points $\{D_j\}_{j=1}^{m}$, constituting a dataset of $\{u_j |_{D_j}\}_{j=1}^{m}$. The point evaluation of the function draws, i.e., $u \in \mathcal{U}$, can also be interpreted as a stochastic process where the process is discretely observed. The behavior of the stochastic process is characterized by the finite-dimensional distributions, such that for any discretization $D^{\dagger}$, $p(u|_{D^{\dagger}})$ denotes the probability distribution of points evaluated on a collection $D^{\dagger}$. What's more, $p(u\big|u|_{D^{\dagger}})$ is $p(u ~ \textit{given} ~ u|_{D^{\dagger}})$, that is the posterior. Similarly, $p(u|_{D'} \big| u|_{D^{\dagger}})$ is the posterior on $D'$ collocation points. In the UFR setting, each data sample $u_j$ is provided on a specific collection of domain points $D_j$, and for the sake of simplicity of notation, we consider all the collected domain points to be the same and referred to by $D$.

The main task in UFR is to learn the measure $\mathbb{P}_{\mathcal{U}}$ from the data $\{u_j|_{D_j}\}_{j=1}^{m}$ such that, at the inference time, for a given $u|_{D^{\dagger}}$ of a new sample $u$, we aim to find its posterior and the $u$ that maximizes the posterior, 
\begin{align*}
    \max_{u\in\mathcal{U}}  p(u\big|u|_{D^{\dagger}}),
\end{align*}
as well as sampling from the posterior distribution. Please note that, when $\mathbb{P}_{\mathcal{U}}$ is known to be Gaussian, then UFR reduces to GPR. 

\textbf{Invertible operators on function spaces.} Consider two Polish function spaces, denoted $\mathcal{A}$ and $\mathcal{U}$, such that for any $a \in \mathcal{A}, a : \mathcal{D}_{\mathcal{A}} \rightarrow \mathbb{R}^{d_{\mathcal{A}}}$, and for any $u \in \mathcal{U}, u : \mathcal{D}_{\mathcal{U}} \rightarrow \mathbb{R}^{d_{\mathcal{U}}}$. Here, $\mathcal{U}$ represents the space of the data, whereas $\mathcal{A}$ represents the latent space. Let $\mathbb{P}_{\mathcal{U}}$ and $\mathbb{P}_{\mathcal{A}}$ be probability measures defined on $\mathcal{U}$ and $\mathcal{A}$, respectively. In this work, we take samples of functions in $\mathcal{A}$, i.e., $a_j \in \mathcal{A}$ to be drawn from a known GP, and consequently, the evaluation of $a_j|_{D_{\mathcal{A}}}$ is a Gaussian random variable equipt with a Gaussian probability measure. Under these construction, and for a suitable Gaussian process on $\mathcal{A}$, we may establish an invertible (bijective) operator between $\mathcal{A}$ and $\mathcal{U}$\footnote{We only require homeomorphism between $\mathcal{D}_{\mathcal{A}}$ and $ \mathcal{D}_{\mathcal{U}}$ where the point alignment is through the underlying homeomorphic map, and use $D_{\mathcal{A}}$ to represent the discretized domain of $\mathcal{D}_{\mathcal{A}}$. For simplicity, we consider the case where $\mathcal{D}_{\mathcal{U}} = \mathcal{D}_{\mathcal{A}}$ and the trivial identity map is the homeomorphic map in our experimental study.}. We denote the forward operator as 
$\mathcal{G} : \mathcal{A} \rightarrow \mathcal{U}$, such that for any $a_j$, $\mathcal{G}(a_j) = u_j $, and its corresponding inverse operator $\mathcal{G}^{-1}$ as $\mathcal{F}$, where $\mathcal{F} = \mathcal{G}^{-1}$ and $\mathcal{F} : \mathcal{U} \rightarrow \mathcal{A}$, such that for any $u_j$, $\mathcal{F}(u_j) = a_j $. We aim to approximate $\mathcal{F}$  and correspondingly its inverse $\mathcal{G}$ with an invertible learnable model $\mathcal{F_{\theta}} \approx \mathcal{F}$,  and $\mathcal{G_{\theta}}=\mathcal{F_{\theta}}^{-1}$, from a finite collection of observations $\{u_j |_D \}_{j=1}^{m}$. Finally, to have well-defined invertible maps between the two measures, we assume absolute continuity of measures on $\A$ and $\U$ with respect to each other. Given the GP nature of the measure on $\A$, such continuity is obtained by adding a small amount of GP noise to the measure on $\U$. A similar procedure is envisioned and incorporated by adding Gaussian noise to images in the prior normalizing flow in finite-dimensional spaces~\citep{huang_augmented_2020,shorten_survey_2019}.
\section{Neural Operator Flow}\label{Sec:SV}

\subsection{Invertible Normalized Architecture}
\begin{figure*}
    \centering
    \includegraphics[width=0.75\textwidth]{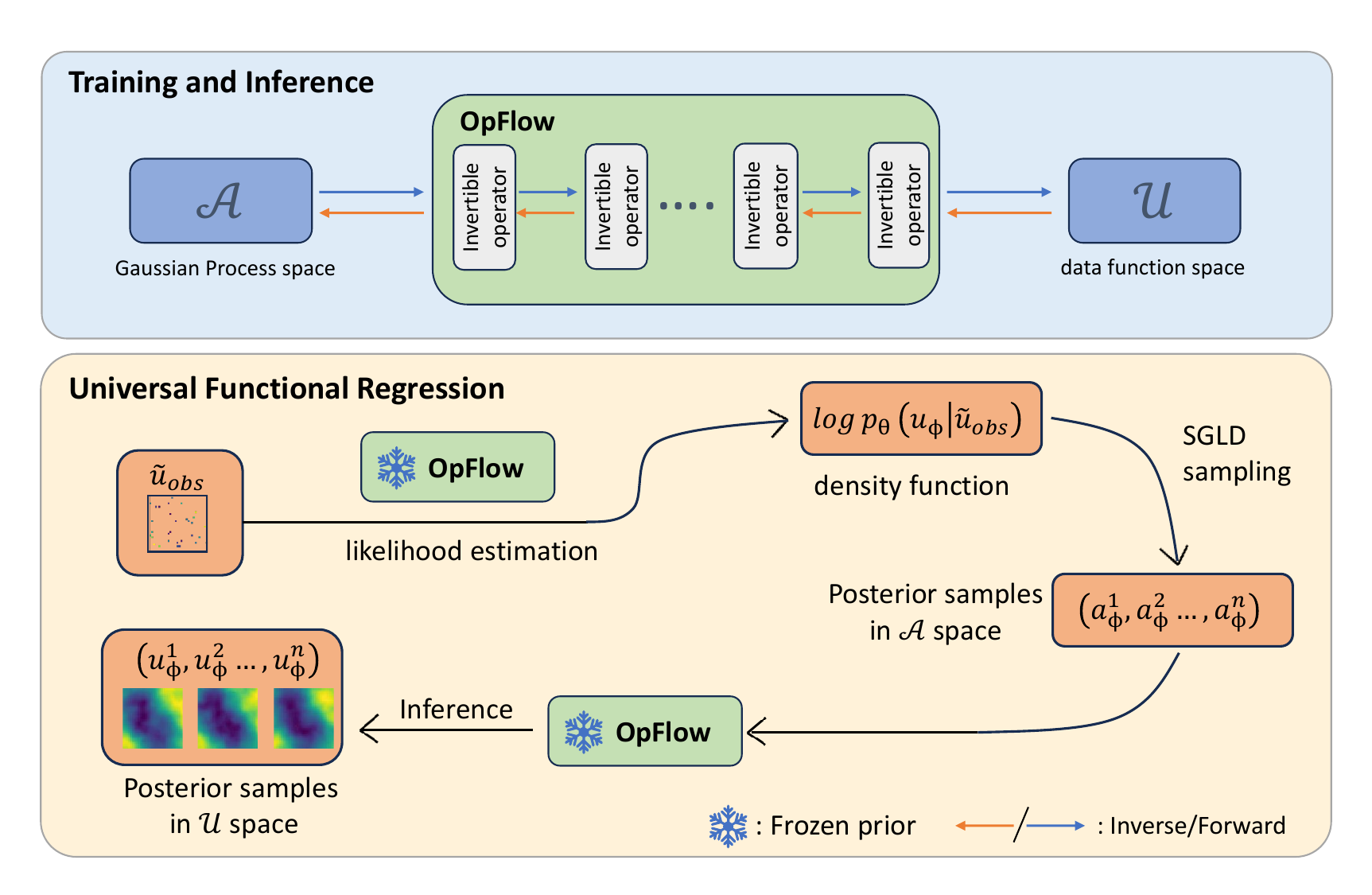}
    \caption{Model architecture of \alg, \alg is composed of a list of invertible operators. For the universal function regression task, \alg is the learnt prior, which is able to provide exact likehood estimation for function point evaluation. $\tilde{u}_{obs}$ is the noisy observation, $u_{\phi}$ is the posterior function of interest and $u_{\phi} = \mathcal{G}_{\theta}(a_\phi)$, where $\mathcal{G}_{\theta}$ is the learnt forward operator.}
    \label{fig:OpFlow_model}
\end{figure*}

We introduce the Neural Operator Flow (\alg), an innovative framework that extends the principle of the finite-dimensional normalizing flow into the realm of infinite-dimensional function spaces. The architecture is shown schematically in Fig. \ref{fig:OpFlow_model}. \alg retains the invertible structure of normalizing flows, while directly operating on function spaces. It allows for exact likelihood estimation for point estimated functions.
\alg is composed of a sequence of layers, each containing the following parts:

\textbf{Actnorm.} Channel-wise normalization with trainable scale and bias parameters is often effective in facilitating affine transformations in normalizing flow neural network training~\citep{kingma_glow_2018}. We implement a function space analog by representing the scale and bias as learnable constant-valued functions. Specifically, let $(v')^i$ denote the input function to the $i$th actnorm layer of \alg and let $ s_{\theta}^i$ and $b_{\theta}^i$ represent the learnable constant-valued scale and bias functions, respectively, in this layer. The output of the actnorm layer is then given by $ v^i = s_{\theta}^i \odot (v')^i + b_{\theta}^i $, where $ \odot $ denotes pointwise multiplication in the function space.

\textbf{Domain and codomain partitioning.} The physical domain refers to the space in which the data resides, such as the time domain for 1D data or the spatial domain for 2D data. Conversely, the channel domain is defined as the codomain of the function data, such as the 3 channels in RGB images or temperature, velocity vector, pressure, and precipitation in weather forecast~\citep{pathak_fourcastnet_2022,hersbach_era5_2020}. In normalizing flows, a bijective structure requires dividing the input domain into two segments~\citep{dinh_density_2017, kingma_glow_2018}. Based on the method of splitting the data domain, we propose two distinct \alg architectures.

\textit{(i) Domain partitioning}: In this architecture, we apply checkerboard masks to the physical domain, a technique widely used in existing normalizing flow models~\citep{dinh_density_2017, kingma_glow_2018} which we extend to continuous domains. Our experiments in the following sections reveal that the domain decomposed \alg is efficient and expressive with a minor downside in introducing checkerboard pattern artifacts in zero-shot super-resolution tasks.

\textit{(ii) Codomain partitioning}: This approach partitions the data along the codomain of the input function data, while maintaining the integrity of the entire physical domain. We show that, unlike its domain decomposed counterpart, the codomain \alg does not produce artifacts in zero-shot super-resolution experiments. However, it exhibits a lower level of expressiveness compared to the domain decomposed \alg.

\textbf{Affine coupling.}
Let $v^i$ represent the input function data to the $i$th affine coupling layer of \alg. Then, we split $v^i$ either along the physical domain or codomain, and we have two halves $h_1^i, h_2^i = \mathbf{split}(v^i)$. Let $\mathcal{T}_{\theta}^i$ denote the $i$th affine coupling layer of \alg, where $\mathcal{T}_{\theta}^i$ is a FNO layer~\citep{li_fourier_2021}, which ensures the model is resolution agnostic. Then, we have $ \log(s^i), b^i = \mathcal{T}^i_{\theta}(h_1^i)$, where $s^i$ and $b^i$ are scale and shift functions, respectively, output by the $i$th affine coupling layer. From here, we update $h_2^i$ through $(h^{\prime})^i_2 = s^i \odot h_2^i + b^i$. Finally, we output the function  $(v^{\prime})^{i+1}$ for the next layer where $(v^{\prime})^{i+1} = \mathbf{concat}(h_1^i, (h^{\prime})_2^i)$. The $\mathbf{concat}$ (concatenation) operation corresponds to the reversing the split operation, and is either in the physical domain or codomain. 

\textbf{Inverse process.}
The inverse process for the $i$the layer of \alg is summarized in Algorithm~\ref{algo:forward_inverse}. First, we split the input function $(v^{\prime})^{i+1}$ with $h_1^i, (h^{\prime})_2^i = \mathbf{split}((v')^{i+1})$. Then the inverse process of the affine coupling layer also takes $h_1^i$ as input, and generates the scale and shift functions $ \log(s^i), b^i = \mathcal{T}^i_{\theta}(h_1^i)$. Thus, both forward and inverse processes utilize the $h_1^i$ as input for the affine coupling layer, which implies that the scale and shift functions derived from the forward process and inverse process are identical from the affine coupling layer. Then we can reconstruct $h_2^i$ and $v^i$ through $h_2^i = ((h^{\prime})^i_2 - b^i)/s^i$; $v^i = \mathbf{concat}(h^i_1, h^i_2)$. Finally, we have the inverse of the actnorm layer with $(v')^i = (v^i - b_{\theta}^i)/s_{\theta}^i $. Similar inverse processes of normalizing flows are described in~\citep{kingma_glow_2018, dinh_density_2017}.

\subsection{Training}
Since \alg is a bijective operator, we only need to learn the inverse mapping $\F_{\theta} : \mathcal{U} \rightarrow \mathcal{A}$, as the forward operator is immediately available. We train \alg in a similar way to other infinite-dimensional generative models~\citep{rahman_generative_2022, seidman_variational_2023,lim_score-based_2023} with special addition of domain alignment. The space of functions for $\mathcal{A}$ is taken to be drawn from a GP, which allows for exact likelihood estimation and efficient training of \alg by minimizing the negative log-likelihood. Since \alg is discretization agnostic~\citep{rahman_generative_2022, kovachki_neural_2023}, it can be trained on various discretizations of $\mathcal{D}_{\mathcal{U}}$ and be later applied to a new set of discretizations, and in the limit, to the whole of $\mathcal{D_U}$~\citep{kovachki_neural_2023}. We now define the training objective $\mathcal{L} $ for model training as follows,
\begin{equation}
    \mathcal{L} = -\min_{\theta \in \Theta} \mathbb{E}_{u \sim \mathbb{P}_\mathcal{U}}[\log p_{\theta}(u|_D)], ~~ \textit{where,} ~~ \log p_{\theta}(u|_D) =  \log p(a|_{D_{\mathcal{A}}}) + \sum_{i=1}^{s} \log |\det(\frac{\partial(v^i|_{D_{\mathcal{V}}^{i}})}{\partial(v^{i-1}|_{D_{\mathcal{V}}^{i-1}})})|.
    \label{loss:basic}
\end{equation}

%
The inverse operator is composed of $s$ invertible operator layers with $\mathcal{F}_{\theta} :=  \mathcal{F}_{\theta}^s \circ \mathcal{F}_{\theta}^{s-1} \cdots \mathcal{F}_{\theta}^{0} $, and gradually transforms $v^0$ to $v^1, \cdots v^{s-1}, v^s$, where $v^0 = u , v^s = a$ and the alignment  naturally holds within the domains associated with $v^0, v^1, \cdots v^s$ as each layer of $\mathcal{F}_{\theta}$ is an invertible operator. $\mathcal{V}^{i} $ is the collection of positions for function $v^{i}$ with $D_\mathcal{V}^{0} = D $, $D_\mathcal{V}^{s} = D_\mathcal{A}$, and each data point has its own discretization $D$~\footnote{Unlike finite-dimensional normalizing flows on fixed-size and regular grids, \alg can provide likelihood estimation $\log p_{\theta}(u|_{D})$ on arbitrary grids $D$, which represents a potentially irregular discretization of \(\mathcal{D}_{\mathcal{U}}\). To compute $\log p_{\theta}(u|_{D})$, we need the homeomorphic mapping between $\mathcal{D}_{\mathcal{A}}$ and $\mathcal{D}_{\mathcal{U}}$ to find $D_{\mathcal{A}}$, corresponding to $D$. The density function in Eq. \ref{loss:basic} holds for new points of evaluation. Our implementation of \alg incorporates FNO that requires inputs on regular grids. Consequently, although \alg is designed to handle functions for irregular grids, our specific implementation of \alg only allows for likelihood estimation on regular girds with arbitrary resolution.}. 

Although the framework of \alg is well-defined, the negative log-likelihood for high-dimensional data may suffer from instabilities during training, leading to a failure to recover the true distribution. Through our experiments, we found that relying solely on the likelihood objective in Eq \ref{loss:basic} was insufficient for training \alg successfully. To address this, we introduce a regularization term, to ensure that \alg learns the true probability measure by helping to stabilize the training process and leading to faster convergence. Such regularization is inspired by the infinite-dimensional Wasserstein loss used in~\citep{rahman_generative_2022}. Unlike the implicit Wasserstein loss provided by the discriminator in GANO, we could potentially have a closed-form expression of the infinite-dimensional Wasserstein loss in the context of \alg. This is due to the learning process mapping function data onto a well-understood GP. The inherent properties of Gaussian Processes facilitate measuring the distance between the learned probability measure and the true probability measure, which is explained into detail in the subsequent sections.

A common way for measuring the distance between two probability measures is square 2-Wasserstein~\citep{mallasto_learning_2017} distance, which is defined as 
\begin{equation}
    W_2^2(\mathbb{P}_{\mathcal{A}}, \mathcal{F}_{\theta}{\sharp}\mathbb{P}_\mathcal{U}) = \inf_{\pi \in \Pi} \int_{\mathcal{A} \times \mathcal{A}} d_2^2(a_1, a_2) d\pi, \ (a_1, a_2) \in {\mathcal{A}} \times {\mathcal{A}},
    \label{eq:w2dist}
\end{equation}

where $\Pi$ is a space of joint measures defined on $\A\times\A$ such that for any $\pi\in\Pi$, the margin measures of $\pi(a_1,a_2)$ on the first and second arguments are $\mathbb{P}_{\mathcal{A}}$ and $\mathcal{F}_{\theta}{\sharp}\mathbb{P}_\mathcal{U}$, and $d_2$ is a metric induced by the $L^2$ norm. By construction, $\mathcal{A} \sim \mathcal{GP}$ and if sufficiently trained, $\mathcal{F_{\theta} \sharp U} \sim \mathcal{GP}$ . We can thus further simplify Eq. \ref{eq:w2dist}, which is named as $F^2ID$ score by ~\cite{rahman_pacmo_2022}, and is defined as:
\begin{equation}
    W_2^2(\mathbb{P}_{\mathcal{A}}, \mathcal{F}_{\theta}{\sharp}\mathbb{P}_\mathcal{U}) = d^2_2(m_1, m_2) + Tr(\mathcal{K}_1 + \mathcal{K}_2 - 2(\mathcal{K}_1^{\frac{1}{2}} \mathcal{K}_2 \mathcal{K}_1^{\frac{1}{2}})^{\frac{1}{2}}).
\label{eq:gp2dist}
\end{equation}

for which $\mathbb{P}_{\mathcal{A}}$ is chosen to be $\mathcal{GP}_1(m_1, \mathcal{K}_1)$ and $\mathcal{GP}_2(m_2, \mathcal{K}_2)$ is the GP fit to the push-forward measure $\mathcal{F}_{\theta}{\sharp}\mathbb{P}_\mathcal{U}$. The $m_1, \mathcal{K}_1$ are the mean function and covariance operator of $\mathcal{GP}_1(m_1, \mathcal{K}_1)$, whereas $m_2, \mathcal{K}_2$ are the mean function and covariance operator of $\mathcal{GP}_2(m_2, \mathcal{K}_2)$. We note that it is equivalent to say a function $f \sim \mathcal{GP}(m, k)$ or $f \sim \mathcal{GP}(m, \mathcal{K})$, where $k$ is the covariance function and $\mathcal{K}$ is the covariance operator~\citep{mallasto_learning_2017}. We take Eq. \ref{eq:gp2dist} as the regularization term. Furthermore, we inject noise with negligible amplitude that sampled from $\mathcal{GP}_1$ to the training dataset $\{u_j\}_{j=1}^m$. In this way, we can have the inversion map to be well-define and stable. From here, we can define a second objective function. 
\begin{equation}
  \mathcal{L} =  -\min_{\theta \in \Theta} \mathbb{E}_{u \sim \mathbb{P}_u} [ \log p_{\theta}(u |_D)] + \lambda     W_2^2(\mathbb{P}_{\mathcal{A}}|_D, \mathcal{F}_{\theta}{\sharp}\mathbb{P}_\mathcal{U}|_D),
  \label{eq:firststep}
\end{equation} where $\lambda$ is the weight for the regularization term. For the collections of position  $D_\mathcal{A} = \{y_i\}^l_{i=1}, y_i \in \mathcal{D}_{\mathcal{A}}$, we select a finite set of orthonormal basis $\{ e_1, e_2, .. e_l \}$ on the $L^2(\mathcal{D}_\mathcal{A})$ space  such that $\lVert e_i \rVert_2 = \frac{1}{\sqrt{l}}, i \in \{1,..l\}$. Thus the relationship between the covariance operator and the covariance function form can be expressed as:
\begin{equation}
    <\mathcal{K}e_i, e_j> = \int_{\mathcal{D}_{\mathcal{A}}} \frac{1}{\sqrt{l}}k(y, y_i)e_j(y)dy = \frac{1}{l}k(y_i, y_j)
    \label{eq:cov_to_matrix}
\end{equation}
Thus, given the chosen set of orthonormal basis functions, the covariance matrix (covariance function in matrix form) $K$ associated with the covariance operator $\mathcal{K}$ follows the relationship:
\begin{equation}
    K[i,j] = l<\mathcal{K}e_i, e_j> = k(y_i, y_j)
    \label{eq:operator_to_matrix}
\end{equation}
Where, $K[i,j]$ is the $(i_{th}, j_{th})$ component of the covariance matrix $K$. Detailed explanation of the chosen basis function can be found in the Appendix A.1 in~\citep{rahman_pacmo_2022}. Thus the matrix form of the Eq. \ref{eq:gp2dist} can be expressed as follows,
\begin{equation}
    \frac{1}{l} \lVert h_1 - h_2 \rVert^2_2 + \frac{1}{l}Tr(K_1 + K_2 - 2(K_1^{\frac{1}{2}}K_2 K_1^{\frac{1}{2}})^{\frac{1}{2}}) = \frac{1}{l} (\lVert h_1 - h_2 \rVert^2_2 + \lVert K_1^{\frac{1}{2}} - K_2^{\frac{1}{2}}\rVert^2_{f})
\label{eq:gp2dist_matrix}
\end{equation}
and $h1, h2$ are the matrix forms of means $m_1, m_2$ in Eq. \ref{eq:gp2dist}, $K_1, K_2$ are the covariance matrix, corresponding to $\mathcal{K}_1$ and $\mathcal{K}_2$. $\lVert \rVert_2^2$ is the square of $L^2$ norm and $\lVert \rVert_{f}^2$ is the square of Frobenius norm. For detailed proof of the simplification of the trace of covariance matrix in the commutative case for two Gaussian processes in Eq. \ref{eq:gp2dist_matrix}, please refer to~\citep{givens_class_1984}.
Equipped with the above preliminary knowledge, we introduce a two-step strategy for training as shown in Algorithm \ref{algo:train}. In the first step, we use the objective defined in Eq. \ref{eq:firststep} with $\lambda>0$, whereas in the second step, we fine-tune the model by setting $\lambda=0$ and a smaller learning rate is preferred. For step one, even though we have the closed and simplified form of the matrix form of square 2-Wasserstein distance shown in Eq. \ref{eq:gp2dist_matrix}, calculating the differentiable square root of the matrix is still time consuming and may be numerically unstable~\citep{pleiss_fast_2020}. Here, we directly calculate $\lVert K_1 - K_2\rVert^2_{f}$ instead of $\lVert K_1^{\frac{1}{2}} - K_2^{\frac{1}{2}}\rVert^2_{f}$, utilizing $\wh W_2^2=\frac{1}{l}(\lVert h_1 - h_2 \rVert_2^2 + \lVert K_1 - K_2 \rVert_{f}^2)$, and such approximation is acceptable due to the the regularization term only exists in the warmup phase and is not a tight preparation. In the Appendix~\ref{apx:ablation}, we show an ablation study that demonstrates the need for the regularization in the warmup phase despite the strong assumption that $\mathcal{GP}_1, \mathcal{GP}_2$ are commutative, which may not always satisfied. Furthermore, Eq. \ref{eq:gp2dist_matrix} implies we normalize the Wasserstein distance and makes sure the square of $L^2$ norm and Frobenius norm remain finite as the number of discretization points increases.

\begin{algorithm}
\caption{\alg Training}
\begin{algorithmic}[1]

\State 
\textbf{Input:} Training dataset $\{u_j | _{D}\}_{j=1}^{m}$; collection of positions $D = \{x_i, \cdots x_l\}$; mean vector $h_1$ and covariance matrix $K_1$; noise level $\gamma$; random noise samples $\{\nu_i\}_{i=1}^n \sim \mathcal{GP}_1$; regularization weight $\lambda$

\State
\textbf{Initialize the reverse \alg $\mathcal{F}_{\theta}$}

\While{warmup} 
    \State Random batch $\{u_i|_{D} \}_{i=1}^n$ from $\{u_j | _{D}\}_{j=1}^{m}$
    \State Inject $\mathcal{GP}$ noise:
    $\{u_i|_{D} \}_{i=1}^n$ = $\{u_i|_{D} + \gamma \cdot \nu_i|_{D} \}_{i=1}^n$

    \State Construct $h_2$, $K_2$: $h_2$ = $\frac{1}{n} \sum_{i=1}^{n} \mathcal{F}_{\theta}(u_i|_{D})$, $K_2 = cov(\{ \mathcal{F}_{\theta}(u_i|_{D}) \}_{i=1}^n)$

    \State Calculate $W_2^2$ approximate regularization term:
    $\wh W_2^2 = \frac{1}{l}(\lVert h_1 - h_2 \rVert_2^2 + \lVert K_1 - K_2 \rVert_{f}^2)$
    \State Compute the raining objective $\mathcal{L} = -\frac{1}{n} \sum_{i}^{n}\log p_{\theta}(u_i|_{D}) + \lambda \wh W_2^2 $
    \State Update $\mathcal{F}_{\theta}$ through gradient descent of $\mathcal{L}$
\EndWhile
\While{finetune} 
    \State Random batch $\{u_i|_{D} \}_{i=1}^n$ from $\{u_j | _{D}\}_{j=1}^{m}$
    \State Inject $\mathcal{GP}$ noise:
    $\{u_i|_{D} \}_{i=1}^n$ = $\{u_i|_{D} + \gamma \cdot \nu_i|_{D} \}_{i=1}^n$

    \State Computing training objective $\mathcal{L} = -\frac{1}{n} \sum_{i}^{n}\log p_{\theta}(u_i|_{D}) $
    \State Update $\mathcal{F}_{\theta}$ through gradient descent of $\mathcal{L}$
\EndWhile
\end{algorithmic}
\label{algo:train}
\end{algorithm}

\subsection{Universal Functional Regression with \alg}

Bayesian functional regression treats the mapping function itself as a stochastic process, placing a prior distribution over the mapping function space. This allows for predictions with uncertainty estimates throughout $\mathcal{D}$, most importantly away from the observation points. GPR places a GP prior on the mapping function space, a common choice because of mathematical tractability. However, there are many real world datasets and scenarios that are not well-represented by a GP~\citep{kindap_non-gaussian_2022}. To address this, we outline an algorithm for UFR with \alg in Bayesian inference problems.

In the following we assume that \alg has been trained previously and sufficiently approximates the true stochastic process of the data. We thus have an invertible mapping between function spaces $\mathcal{U}$ and $\mathcal{A}$. We now can use the trained \alg as a learned prior distribution over the space of possible mapping functions, fixing the parameters of \alg for functional regression.

Similar to GPR, suppose we are given $\tilde{u}_{obs}$, pointwise and potentially noisy observations of a function $u$, where the observation points are at the discretization set $D \subset \mathcal{D}_{\mathcal{U}}$. We refer to the evaluations at $D$ as $u_{obs}$, and the additive noise $\epsilon$ is the white noise with the law $\epsilon \sim \mathcal{N}(0,\sigma^2)$. Note that this is based on the conventions in GPRs; otherwise, the noise $\epsilon$ can also be treated as a function with a GP from. One more time, similar to GPR, we aim to infer the function's values on a set of new points $D'$ where $D\subset D'$ given the observation $\tilde{u}_{obs}$. The regression domain therefore is $D'$ at which we refer to the inferred function's values as $u_{\phi}$ and $a_\phi = \mathcal{F_{\theta}}(u_\phi)$. Ergo, for the likelihood of the values on points in $D'$, i.e., the likelihood of $u_{\phi}$, we have, 
\begin{align}
    \log p_{\theta}(u_{\phi}|\tilde{u}_{obs})     \label{eqn:reg1} &= \log p_{\theta}(\tilde{u}_{obs}|u_{\phi}) + \log p_{\theta}(u_{\phi}) - \log p_{\theta}(\tilde{u}_{obs}) \\
        \label{eqn:reg2} &= \log p_{\theta}(\tilde{u}_{obs}|u_{\phi|D}) + \log p_{\theta}(u_{\phi}) - \log p_{\theta}(\tilde{u}_{obs}) \\
        \label{eqn:reg3} &= -\frac{\lVert \tilde{u}_{obs} - u_{\phi|D}\rVert^2_2}{2\sigma^2} + \log p_{\theta}(u_{\phi}) - r\log\sigma - \frac{1}{2}r\log(2\pi) -\log p_{\theta}(\tilde{u}_{obs}),   
\end{align}
where $p_{\theta}(u|_D)$ denotes the probability of $u$ evaluated on the collection of points $D$ with the learned prior (\alg), and $r$ is the  cardinality of set $D$ with $r = |D|$. Maximizing the above equation results in maximum \textit{a posteriori} (MAP) estimate of $p_{\theta}(u_{\phi}|\tilde{u}_{obs})$. Since the last three terms in Eq.\ref{eqn:reg3} are constants, such maximization is carried using gradient descent focusing only on the first two terms. In the following, we refer to $\wb u_{\phi}$ as the MAP estimate given $\tilde{u}_{obs}$.



For sampling for the posterior in Eq. \ref{eqn:reg3}, we utilize the fact that \alg is a bijective operator and the target function $u_{\phi}$ uniquely defines $a_{\phi}$ in $\A$ space, therefore, drawing posterior samples of $u_{\phi}$ is equivalent to drawing posterior samples of $a_{\phi}$ in Gaussian space where SGLD with Gaussian random field perturbation is relevant. Our algorithm for sampling from the $p_{\theta}(u_{\phi}|\tilde{u}_{obs})$ is described in Algorithm \ref{algo:sgld}. We initialize $a_{\phi}|\tilde{u}_{obs}$ with the MAP estimate of $p_{\theta}(u_{\phi}|\tilde{u}_{obs})$, i.e., $\wb u_{\phi}$, and follow Langevin dynamics in $\A$ space with Gaussian structure to sample $u_{\phi}$. This choice helps to stabilize the functional regression and further reduces the number of burn-in iterations needed in SGLD. The burn-in time $b$ is chosen to warm up the induced Markov process in SGLD and is hyperparameter to our algorithm~\citep{ahn_stochastic_2015}. 
We initialize the SGLD with $a_{\phi}^0= \F_\theta(\wb u_{\phi})$ and  $u_{\phi}^0= \G_{\theta}(a_{\phi}^0) = \wb u_{\phi}$. Then, at each iteration, we draw a latent space posterior sample $a^t_{\phi}$ in $\mathcal{A}$ space, where $a^t_{\phi}$ is the $t^{th}$ sample of function $a_{\phi}$, and harvest the true posterior samples $u^t_{\phi}$ by calling $\mathcal{G_{\theta}}(a^t_{\phi})$. Since $p_{\theta}(a_{\phi}|\tilde{u}_{obs}) \sim \mathcal{GP}$, it is very efficient to sample from it, and we find that relatively little hyperparameter optimization is needed. Since the sequential samples are correlated, we only store samples every $t_N$ time step, resulting in representative samples of the posterior without the need to store all the steps of SGLD~\citep{riabiz_optimal_2022}.

\begin{algorithm}
\caption{Sample from posterior using SGLD}
\begin{algorithmic}[1]

\State 
\textbf{Input and Parameters:} \alg models $\mathcal{G}_{\theta},\F_\theta$, Langevin dynamics temperature $T$, $\mathcal{G}_{\theta}$, MAP $\wb u_{\phi}$.

\State \textbf{Initialization}: $a_{\phi}^0= \F_\theta(\wb u_{\phi})$ and  $u_{\phi}^0= \G(a_{\phi}^0)$,  
\For{$t = 0, 1, 2, \ldots, N$}
    \State Compute gradient of the posterior: $\nabla_{a_{\phi}} \log p_{\theta}(u^t_{\phi}|\tilde{u}_{obs})$
    \State Update $a_{\phi}^{t+1}$: $a_{\phi}^{t+1} = a_{\phi}^{t} + \frac{\eta_t}{2} \nabla \log p_{\theta}(u^t_{\phi}|\tilde{u}_{obs}) + \sqrt{\eta_t T} \mathcal{N}(0,I)$
    \If{$t\geq b$}
        \State Every $t_N$ iterations:
        obtain new sample $a_{\phi}^{t+1}$,
        $u_{\phi}^{t+1}= \mathcal{G_{\theta}}(a_{\phi}^{t+1})$
    \EndIf
\EndFor
\end{algorithmic}
\label{algo:sgld}
\end{algorithm}


\section{Experiments}

In this section, we aim to provide a comprehensive evaluation of \alg's capabilities for UFR, as well as generation tasks. We consider several ground truth datasets, composed of both Gaussian and non-Gaussian processes, as well as a real-world dataset of earthquake recordings with highly non-Gaussian characteristics.~\footnote{\url{https://github.com/yzshi5/OpFlow}}

\subsection{Universal Functional Regression}

\textbf{Summary of parameters.}
Here, we show UFR experiments using the domain decomposed implementation of \alg, and leave the codomain \alg UFR experiments for the appendix. Table \ref{tab:reg_table} lists the UFR hyperparameters used, as outlined in Algorithm \ref{algo:sgld}. For training \alg, we take $\mathcal{A} \sim \mathcal{GP()}$ with a Matern covariance operator parameterized by length scale $l$ and roughness $\nu$. The mean function for the $\mathcal{GP}$ is set to zero over $\mathcal{D}$. The noise level ($\sigma^2$) of observations in Eq. \ref{eqn:reg3} is assumed as $0.01$ for all regression tasks.

\textbf{Gaussian Processes.}
This experiment aims to duplicate the results of classical GPR, where the posterior is known in closed form, but with \alg instead. We build a training dataset for $\mathcal{U} \sim \mathcal{GP}$, taking $l_{\mathcal{U}} = 0.5, \nu_{\mathcal{U}}=1.5$ and the training dataset contains $30000$ realizations. For the latent space GP, we take $l_{\mathcal{A}} = 0.1, \nu_{\mathcal{A}}=0.5$.The resolution of $\mathcal{D}$ is set to $128$ for the 1D regression experiment. Note that, the \alg is not explicitly notified with the mean and covariance of the data GP, and learns them using samples, while the GP ground truth is based on known parameters. 

\begin{figure*}
\vspace*{-.2cm}
    \centering
    \begin{subfigure}{0.23\textwidth}
        \includegraphics[height=.73\textwidth]{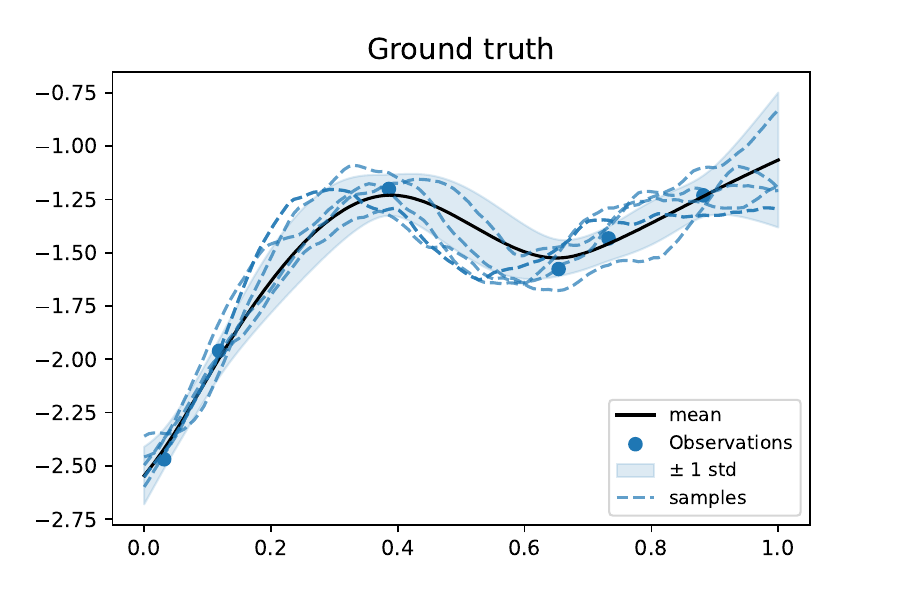}
        \vspace*{-.6cm}
        \caption{}
    \end{subfigure}
    \quad
    \begin{subfigure}{0.23\textwidth}
        \includegraphics[height=.73\textwidth]{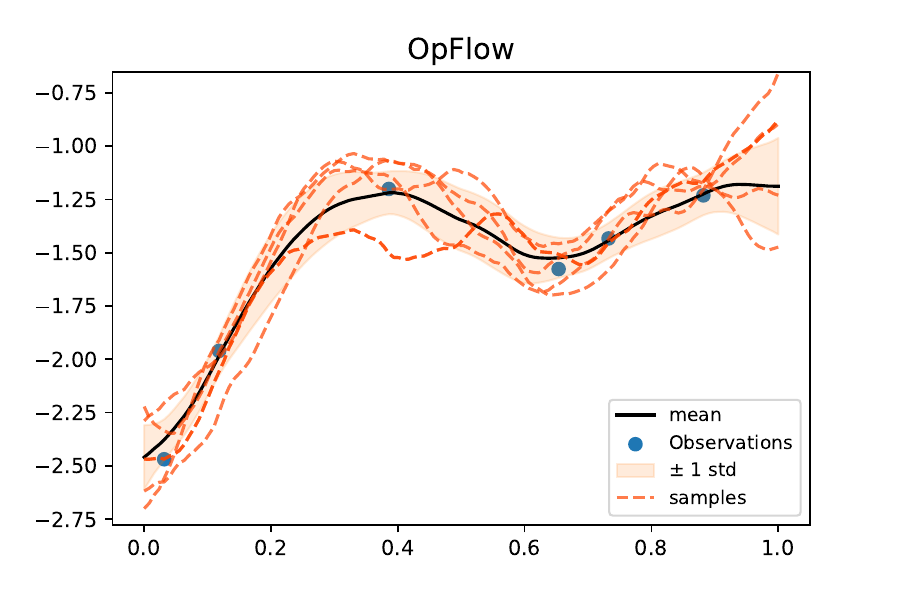}
        \vspace*{-.6cm}
        \caption{}
    \end{subfigure}
    \quad
    \begin{subfigure}{0.23\textwidth}
        \includegraphics[height=.73\textwidth]{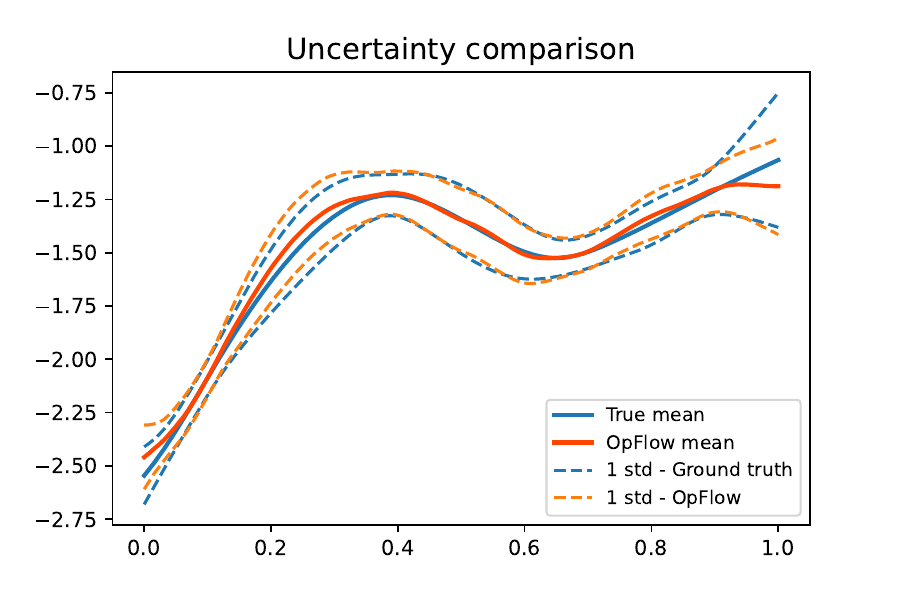}
        \vspace*{-.6cm}
        \caption{}
    \end{subfigure}
    \caption{ \alg regression on GP data. (a) Ground truth GP regression with observed data and predicted samples (b) OpFlow regression with observed data and predicted samples. (c) Uncertainty comparison between true GP and OpFlow predictions.}
    \label{fig:GP_domain_decomposed_opflow_reg}
\end{figure*}

For performing the regression, we generate a single new realization of $\mathcal{U}$, and select observations at $6$ random positions in $\mathcal{D}$. We then infer the posterior for $u$ across the $128$ positions, including noise-free estimations for the observed points. Fig.~\ref{fig:GP_domain_decomposed_opflow_reg} displays the observation points and analytical solution for GP regression for this example, along with the posterior from \alg. In general, the \alg solution matches the analytical solution well; the mean predictions of \alg (equivalent to the MAP estimate for GP data) and its assessed uncertainties align closely with the ground truth. This experiment demonstrates \alg's ability to accurately capture the posterior distribution, for a scenario where the observations are from a Gaussian Process with an exact solution available.

Furthermore, we present the regression results of \alg compared with other Deep GP models in Appendix \ref{Apx:DGP}. In this comparison, \alg exhibits superior performance over conventional Deep GP approach. Last, we explore the minimal size of the training dataset to train an effective prior. In Appendix \ref{Apx:small_set}, we detail our analysis of the 1D GP regression experiment using \alg with reduced training dataset sizes. From this investigation, we conclude that the training dataset containing between 3000 and 6000 samples (10\% to 20\% of the original dataset size) is sufficient to train a robust and effective prior.

\textbf{Truncated Gaussian Processes.} 
Next, we apply \alg to a non-Gaussian functional regression problem. Specifically, we use the truncated Gaussian process (TGP)~\citep{swiler_survey_2020} to construct a 1D training dataset. The TGP is analogous to the truncated Gaussian distribution, which bounds the functions to have amplitudes lying in a specific range. For the dataset, we take $l_{\mathcal{U}} = 0.5$ and $\nu_{\mathcal{U}}=1.5$, and truncate the function amplitudes to the range $[-1.2, 1.2]$. For the latent space GP, we take $l_{\mathcal{A}} = 0.1, \nu_{\mathcal{A}}=0.5$. The resolution is set to 128. We conduct an experiment with only 4 random observation points, infer the function across all 128 positions. Unlike most non-Gaussian processes, where the true posterior is often intractable, the TGP framework, governed by parameters $l_{\mathcal{U}}, \nu_{\mathcal{U}}$ and known bounds, allows for true posterior estimation using rejection sampling. We consider a narrow tolerance value of $\pm 0.10$ around each observation. The tolerance range can be analogous to but conceptually distinct from noise level~\citep{robert_monte_1999,swiler_survey_2020}. For the training dataset, we use 30000 realizations of the TGP. For the posterior evaluation, we generate 1000 TGP posterior samples, and taken the calculated mean and uncertainty from the posterior samples as the ground truths.

\begin{figure*}
\vspace*{-.2cm}
    \centering
    \begin{subfigure}{0.23\textwidth}
        \includegraphics[height=.73\textwidth]{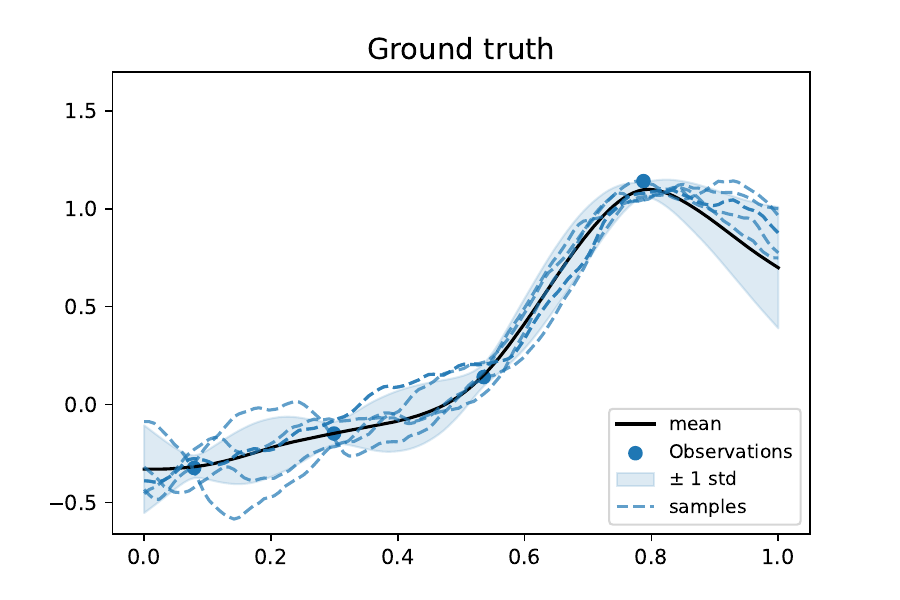}
        \vspace*{-.6cm}
        \caption{}
    \end{subfigure}
    \begin{subfigure}{0.23\textwidth}
        \includegraphics[height=.73\textwidth]{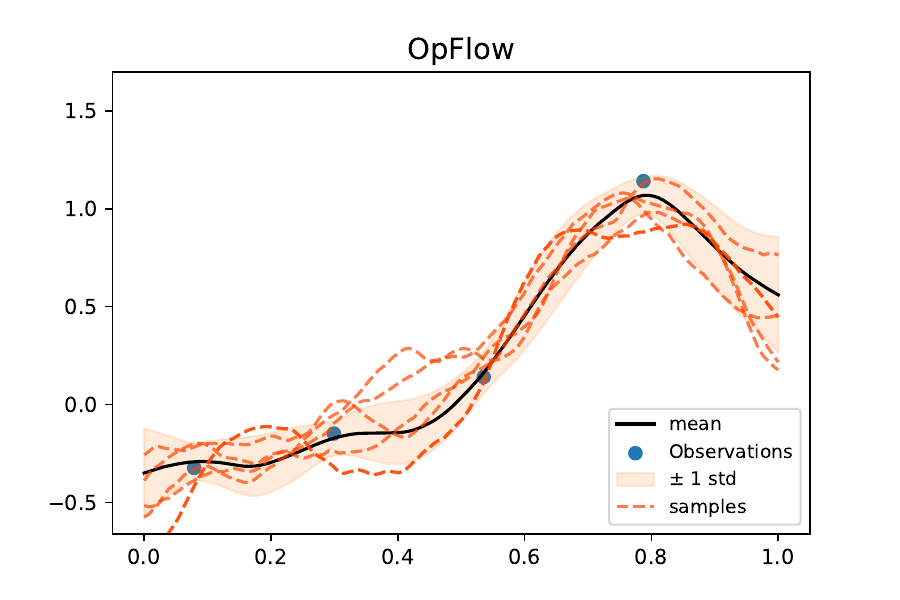}
        \vspace*{-.6cm}
        \caption{}
    \end{subfigure}
    \begin{subfigure}{0.23\textwidth}
        \includegraphics[height=.73\textwidth]{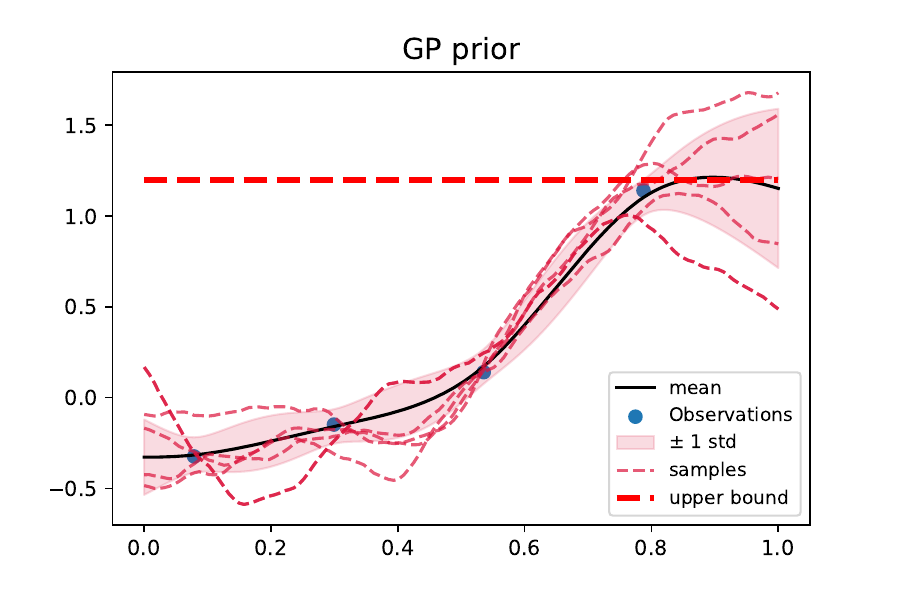}
        \vspace*{-.6cm}
        \caption{}
    \end{subfigure}
    \begin{subfigure}{0.23\textwidth}
        \includegraphics[height=.73\textwidth]{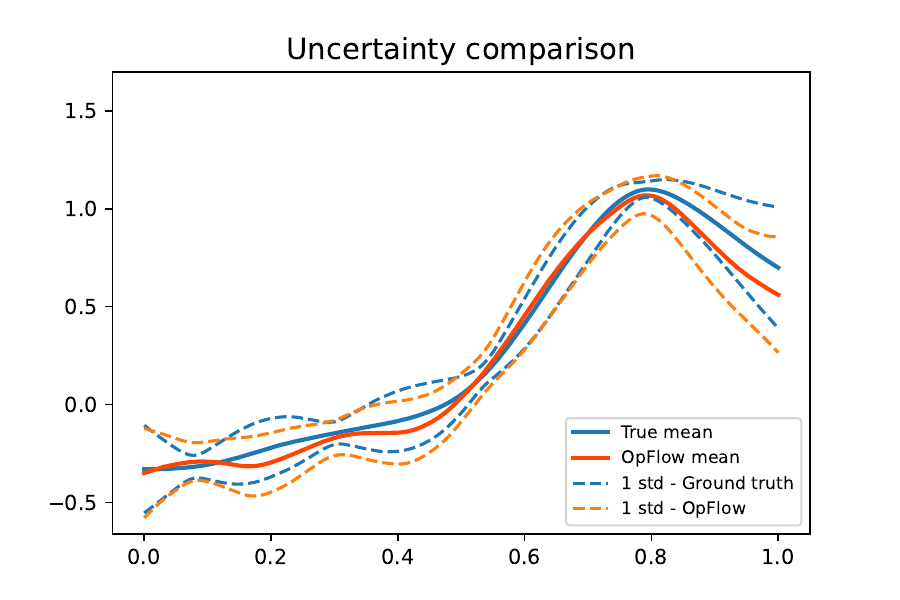}
        \vspace*{-.6cm}
        \caption{}
    \end{subfigure}
    \caption{ \alg regression on TGP data. (a) Ground truth TGP regression with observed data and predicted samples (b) \alg regression with observed data and predicted samples. (c) prior GP regression with observed data and predicted samples (c) Uncertainty comparison between true TGP and \alg predictions. }
    \label{fig:TGP_domain_decomposed_opflow_reg}
\end{figure*}

Fig. \ref{fig:TGP_domain_decomposed_opflow_reg} shows the UFR results for this dataset alongside the true TGP posterior. We see that \alg effectively captures the true non-Gaussian posterior, producing samples that are visually aligned with the ground truth. For comparison, we also show the results of applying GP regression to this dataset. Notice that the GP prior allows for non-zero probability outside of the truncation range, whereas the \alg prior learns these constraints, resulting in a much more accurate posterior. Fig.~\ref{fig:TGP_domain_decomposed_opflow_reg} reveals a slight deviation in \alg's mean prediction starting from $x=0.8$, yet the uncertainty remains accurately represented. This discrepancy may be explained by the challenging nature of this task, which involves using only 4 observations. With more observations, \alg's predictions are expected to improve.

\begin{figure*}
\vspace*{-.2cm}
\hspace*{-.5cm}  
    \centering
    \begin{subfigure}{0.14\textwidth}
        \includegraphics[height=0.8\textwidth,trim={0, 0.2cm, 0, 0},clip]{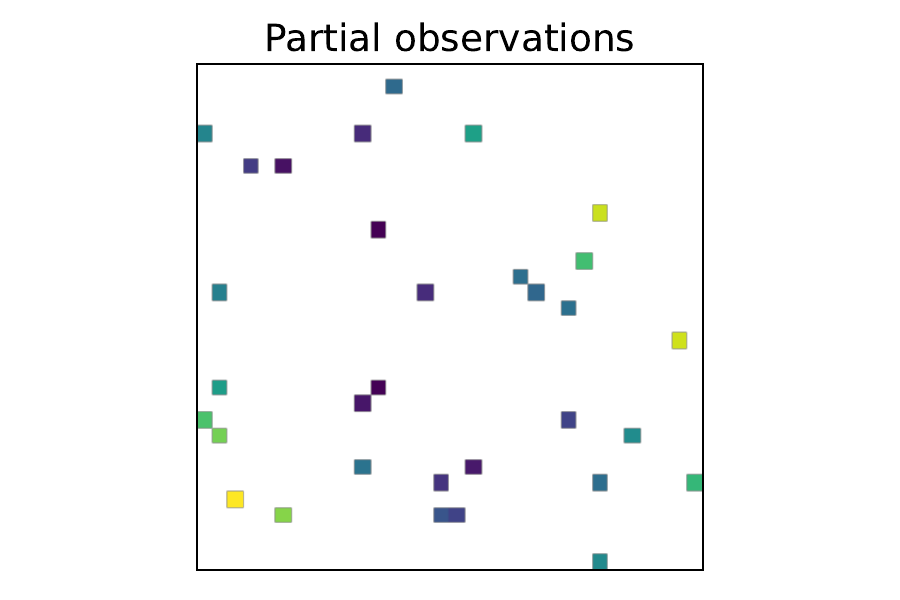}  
        \vspace*{-.6cm}
        \caption{}
    \end{subfigure}
    \begin{subfigure}{0.14\textwidth}
        \centering
        \includegraphics[height=.8\textwidth]{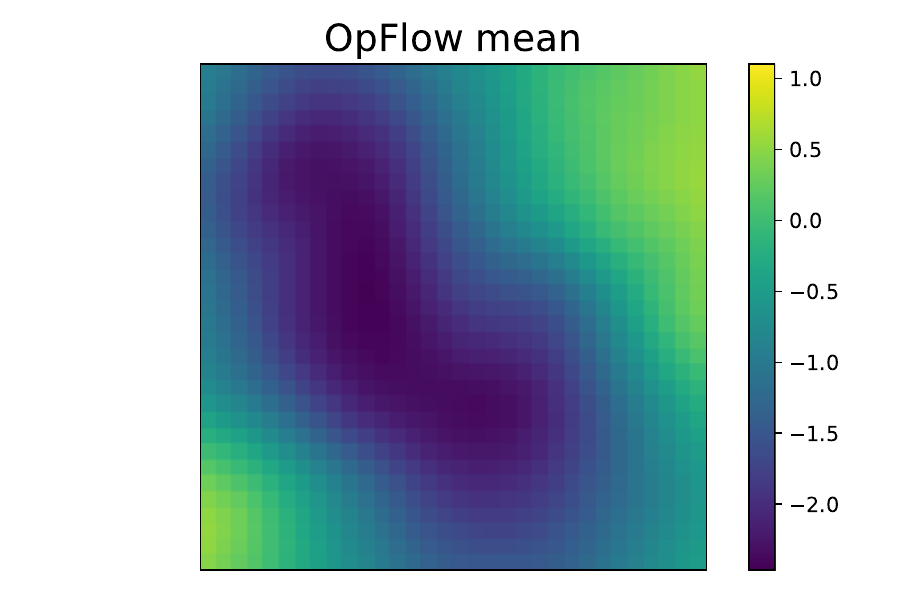}
        \vspace*{-.6cm}
        \caption{}
    \end{subfigure}
    \begin{subfigure}{0.14\textwidth}
        \includegraphics[height=.8\textwidth]{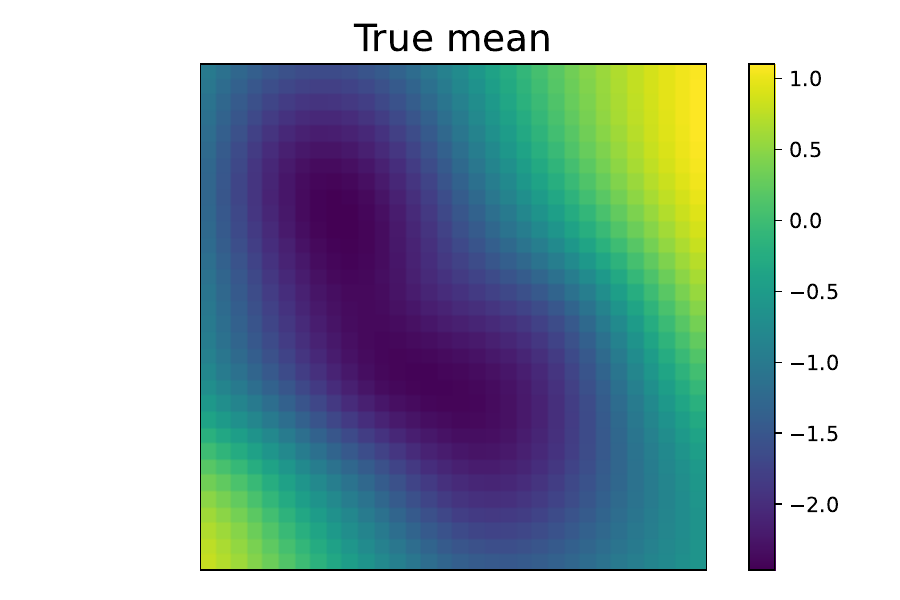}
        \vspace*{-.6cm}
        \caption{}
    \end{subfigure}
    \begin{subfigure}{0.14\textwidth}
        \includegraphics[height=.8\textwidth]{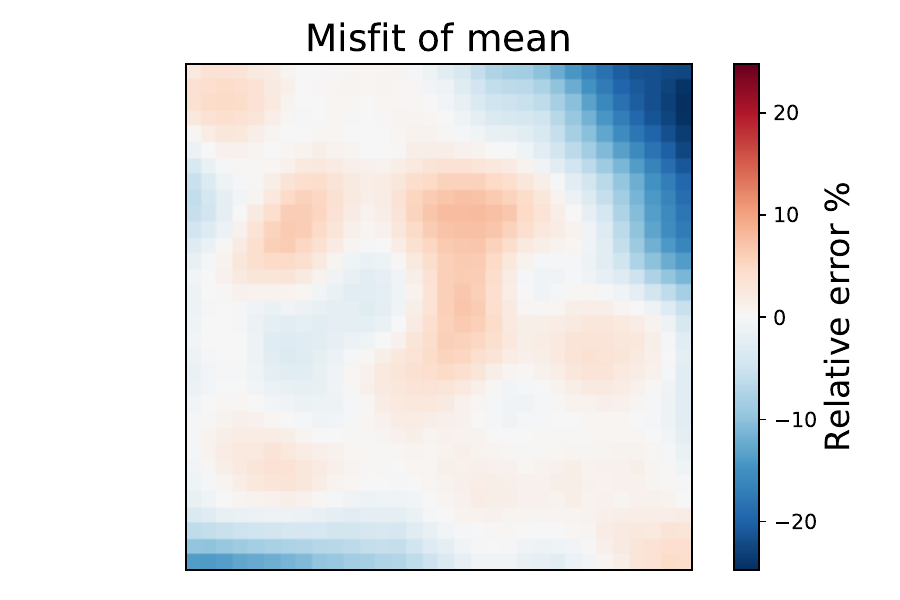}
        \vspace*{-.6cm}
        \caption{}
    \end{subfigure}
    \begin{subfigure}{0.14\textwidth}
        \includegraphics[height=.8\textwidth]{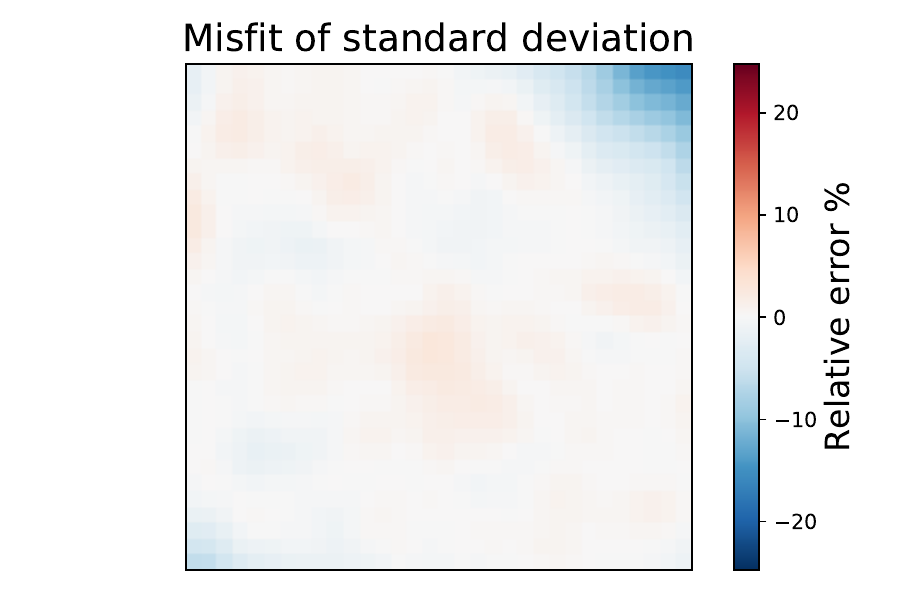}
        \vspace*{-.6cm}
        \caption{}
    \end{subfigure}
    
    \begin{subfigure}{0.35\textwidth}
        \includegraphics[height=.4\textwidth]{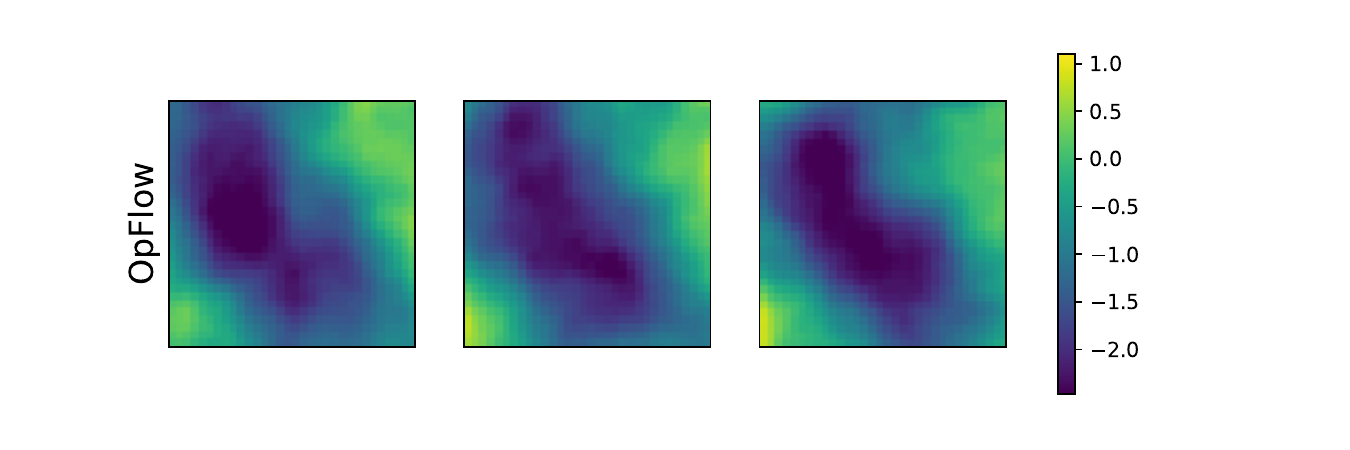}
        \vspace*{-1.cm}
        \caption{}
    \end{subfigure}
    \begin{subfigure}{0.35\textwidth}
        \includegraphics[height=.4\textwidth]{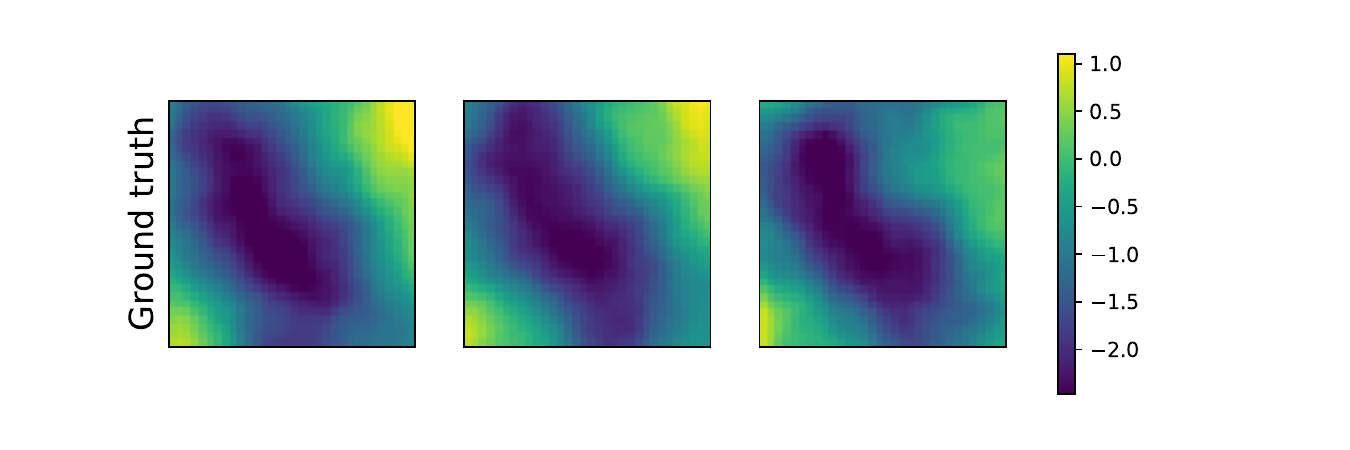}
        \vspace*{-1.cm}
        \caption{}
    \end{subfigure}

    \caption{ \alg regression on 32$\times$32 GRF data. (a) 32 random observations. (b) Predicted mean from \alg. (c) Ground truth mean from GP regression. (d) Misfit of the predicted mean. (e) Misfit of predicted uncertainty. (f) Predicted samples from \alg. (g) Predicted samples from GP regression. }
    \label{fig:domain_decom_GRF_reg2_reg}
\end{figure*}

\textbf{Gaussian random fields.}
We build a 2D dataset of Gaussian random fields having $l_{\mathcal{U}} = 0.5$ and $\nu_{\mathcal{U}}=1.5$. For the latent space, we have a GP with $l_{\mathcal{A}} = 0.1, \nu_{\mathcal{A}}=0.5$. We set the resolution of the physical domain to $32\times32$ for the training data. We generate 20000 samples in total for training. This is another experiment with an analytical expression available for the true posterior. We generate one new realization of the data for functional regression, and choose observations at 32 random positions, as shown in Fig \ref{fig:domain_decom_GRF_reg2_reg}a. We then infer the possible outcomes over the entire $32\times32$ grid. Fig \ref{fig:domain_decom_GRF_reg2_reg} shows that \alg reliably recovers the posterior without assuming beforehand that the true data are from a Gaussian process. The error is non-uniform over the domain, the largest of which occurs in the data-sparse upper right corner. Fig \ref{fig:domain_decom_GRF_reg2_reg} shows several samples of \alg's posterior, mirroring the true posterior variability produced by GPR. We show a second scenario for this same function in the appendix, but where the observations are concentrated along linear patterns (Fig \ref{fig:domain_decom_GRF_reg3_reg}).Similarly, \alg faithfully approximates the true posterior.

\begin{figure*}
\vspace*{-.2cm}
    \centering
    \begin{subfigure}{0.23\textwidth}
        \includegraphics[height=0.70\textwidth]{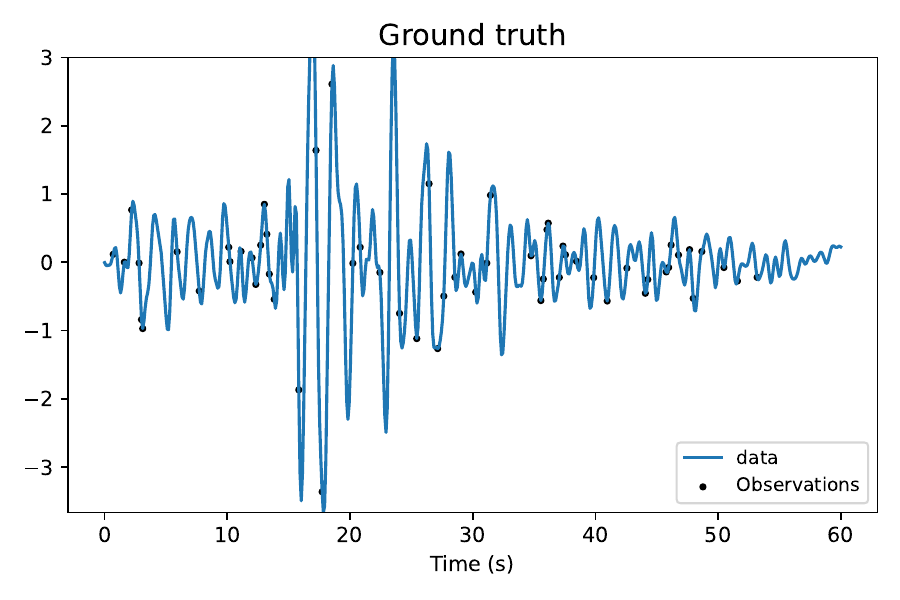}
        \vspace*{-.6cm}
        \caption{}
    \end{subfigure}
    \begin{subfigure}{0.23\textwidth}
        \includegraphics[height=.70\textwidth]{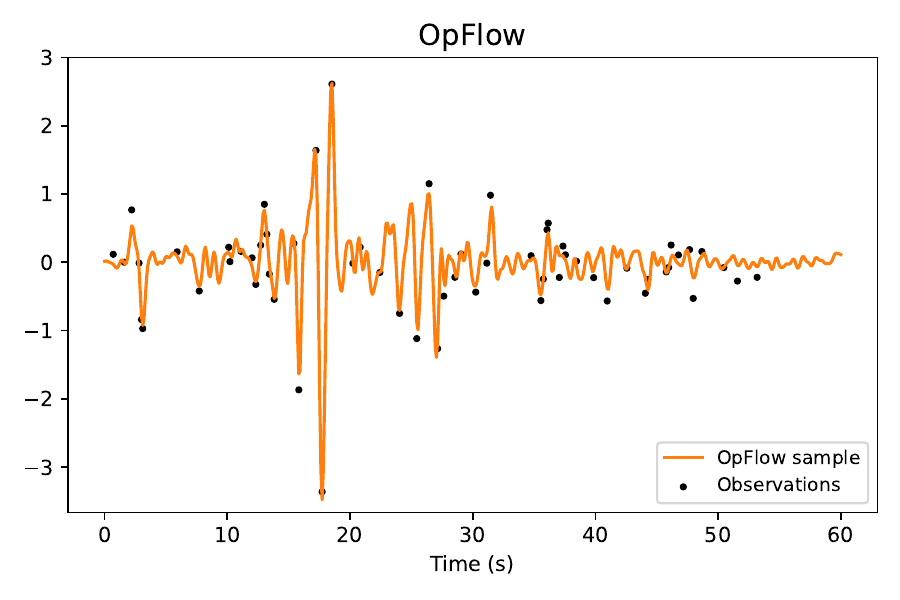}
        \vspace*{-.6cm}
        \caption{}
    \end{subfigure}
    \begin{subfigure}{0.23\textwidth}
        \includegraphics[height=.70\textwidth]{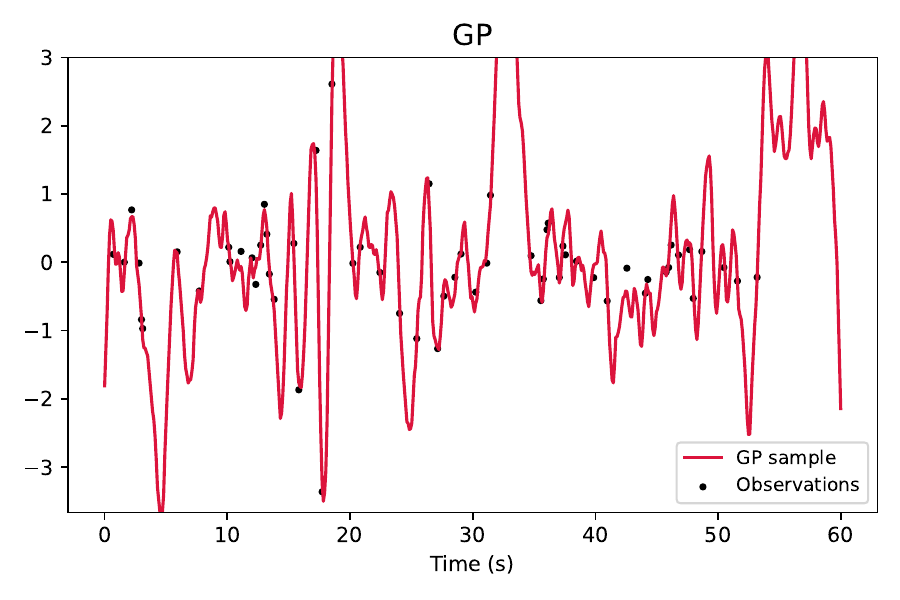}
        \vspace*{-.6cm}
        \caption{}
    \end{subfigure}
    \begin{subfigure}{0.23\textwidth}
        \includegraphics[height=.70\textwidth]{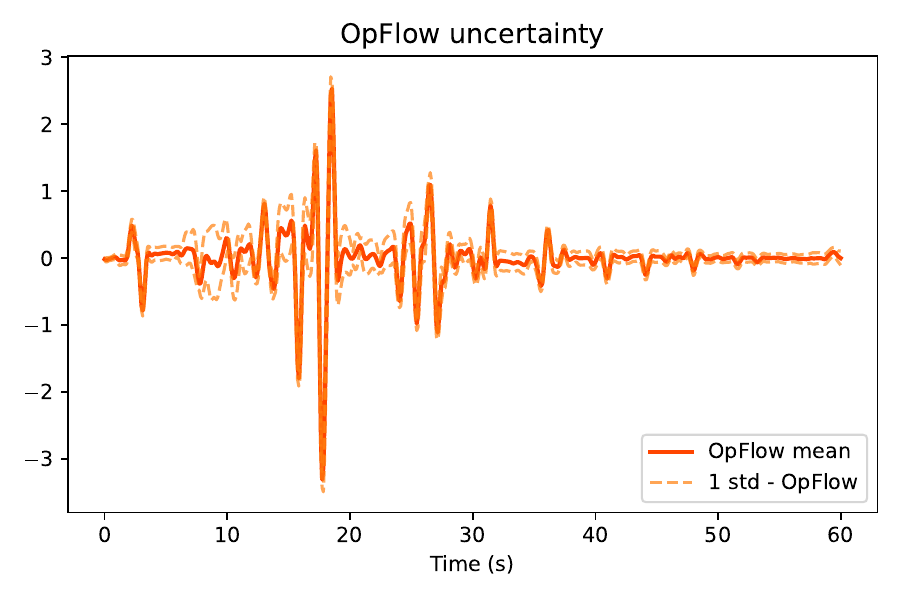}
        \vspace*{-.6cm}
        \caption{}
    \end{subfigure}
    \begin{subfigure}{0.23\textwidth}
        \includegraphics[height=.70\textwidth]{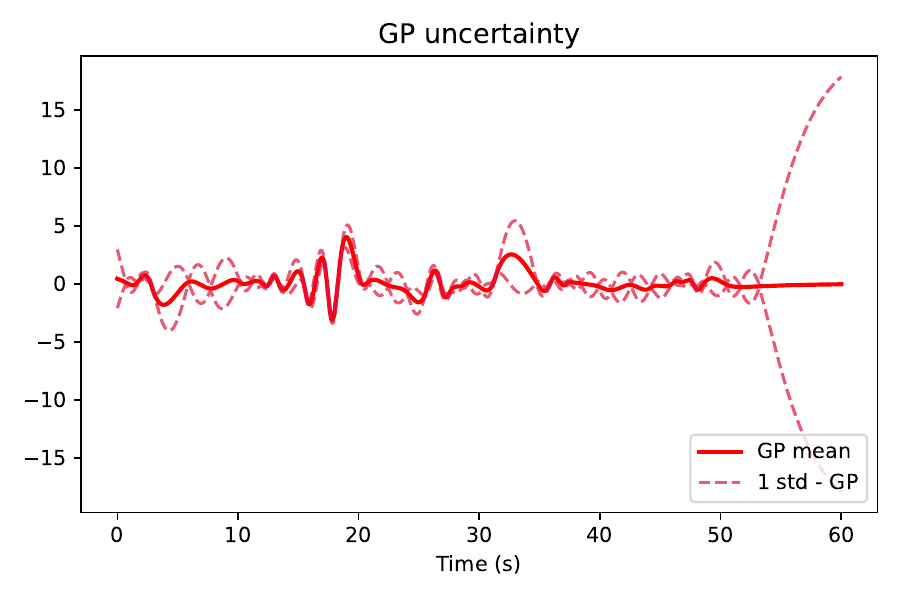}
        \vspace*{-.6cm}
        \caption{}
    \end{subfigure}
    \begin{subfigure}{0.23\textwidth}
        \includegraphics[height=.70\textwidth]{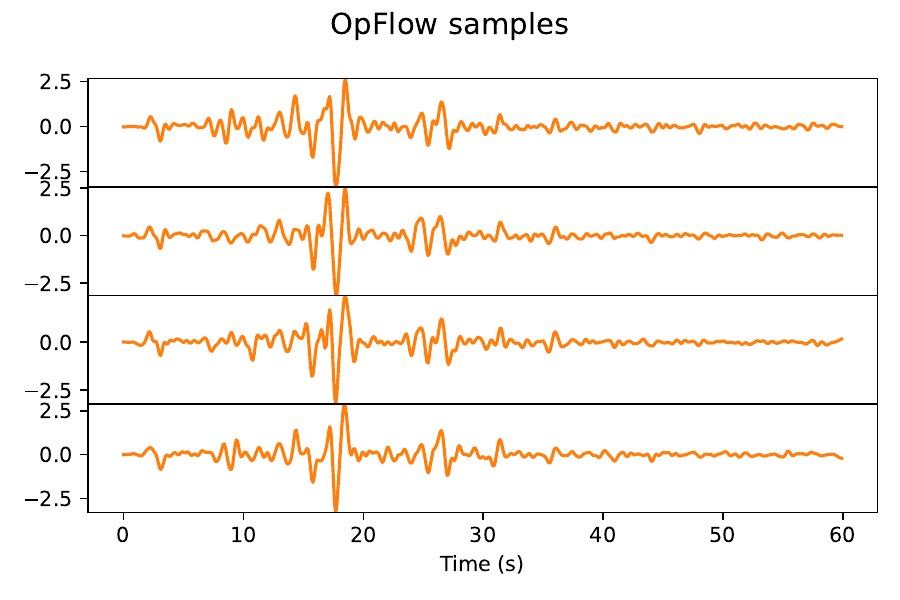}
        \vspace*{-.6cm}
        \caption{}
    \end{subfigure}
    \begin{subfigure}{0.23\textwidth}
        \includegraphics[height=.70\textwidth]{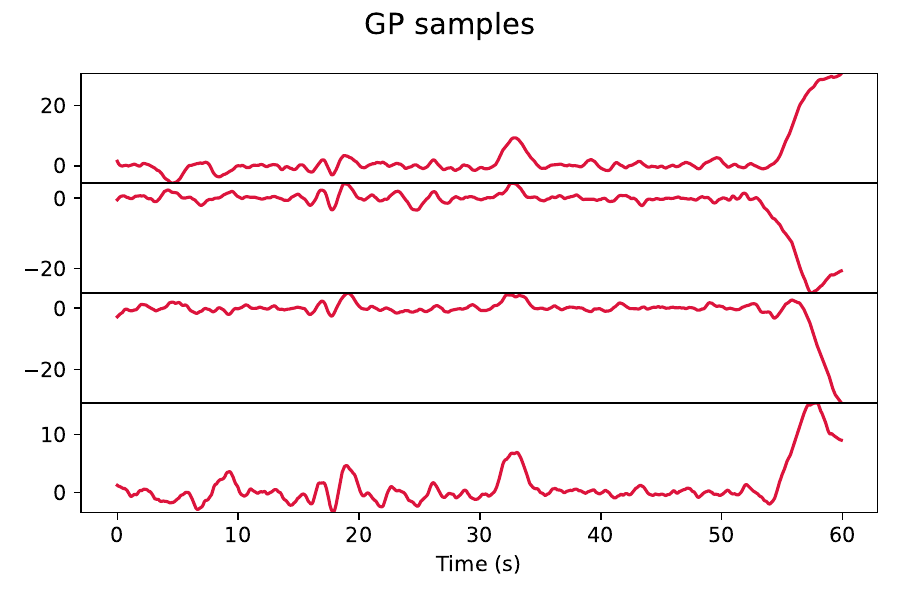}
        \vspace*{-.6cm}
        \caption{}
    \end{subfigure}
    \caption{ \alg regression on seismic waveform data. (a) Ground truth waveform. (b) One predicted sample from \alg. (c) One predicted sample from the best fitted GP. (d) Predicted mean and uncertainty from \alg. (e) Predicted mean and uncertainty from the best fitted GP. (f) Samples from \alg. (g) Samples from the best fitted GP.}
    \label{fig:domain_decom_seismic_reg}
\end{figure*}

\textbf{Seismic waveforms regression.}
In this experiment, \alg regression is applied to a real-world non-Gaussian dataset consisting of earthquake ground motion time series records from Japan Kiban–Kyoshin network (KiK‐net). The raw data 
are provided by the Japanese agency, National Research Institute for Earth Science and Disaster Prevention~\citep{nied_strong-motion_2013}. The KiK-net data encompasses 20,643 time series of ground velocity for earthquakes with magnitudes in the range 4.5 to 5, recorded between 1997 and 2017. All seismic signals were resampled at 10Hz and have a 60 second duration, resulting in a resolution of 600 for our analysis. The time series are aligned such that $t=0$ coincides with the onset of shaking. A second-order Gaussian filter was used to attenuate high-frequency wiggles. We used $l_{\mathcal{A}}=0.05, \nu_{\mathcal{U}}=0.5$ for the latent Gaussian space $\mathcal{A}$. After training the \alg prior, we select a new (unseen) time series randomly for testing (Fig. \ref{fig:domain_decom_seismic_reg}a), and perform UFR over the entire domain with 60 randomly selected observation points, (Fig. \ref{fig:domain_decom_seismic_reg}). For comparison, we show the results for GP regression with parameters optimized via likelihood maximization. Fig \ref{fig:domain_decom_seismic_reg}b shows that a sample from \alg matches the actual data, with \alg accurately capturing key characteristics in the data that include various types of seismic waves arriving at different times in the time series~\citep{shearer_introduction_2019}. Conversely, the standard GP model struggles with the complexity of the seismic data as shown in Fig. \ref{fig:domain_decom_seismic_reg}c. Further, Fig. \ref{fig:domain_decom_seismic_reg}d,e illustrate that \alg provides a reasonable estimation of the mean and uncertainty, whereas the GP model’s uncertainty significantly increases after 50 seconds due to a lack of observations. Additional samples from \alg and GP (Fig. \ref{fig:domain_decom_seismic_reg}f,g) further highlight the superior performance of \alg for functional regression of seismic waveforms. 

\subsection{Generation Tasks}
While our emphasis in this study is on functional regression, there is still value in evaluating the generative capabilities of \alg. Here, we take GANO as a baseline model as it is a generative neural operator with discretization invariance. The following study demonstrates that domain decomposed \alg provides comparable generation capabilities similar to prior works, such as GANO. The implementation of GANO used herein follows that of the paper~\citep{rahman_generative_2022}. In the following experiments, the implementations of the GP for the Gaussian space $\mathcal{A}$ of GANO  is adapted to optimize GANO's performance for each specific case. Unless otherwise specified, the default GP implementation utilize the Gaussian process package available in Scikit-learn~\citep{pedregosa_scikit-learn_2011}. More generation experiments of domain decomposed \alg and codomain \alg are provided in the appendix.

\begin{figure*}[h]
\vspace*{-.2cm}
    \centering
    \hspace{-.6cm}
    \begin{subfigure}{0.14\textwidth}
        \includegraphics[height=.72\textwidth,trim={0 .4cm 0 0},clip]{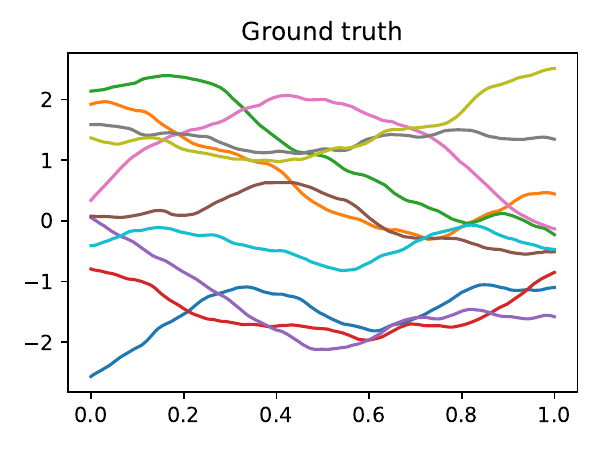}
        \vspace*{-.4cm}
        \caption{}
    \end{subfigure}
    \begin{subfigure}{0.14\textwidth}
        \includegraphics[height=.72\textwidth,trim={0 .4cm 0 0},clip]{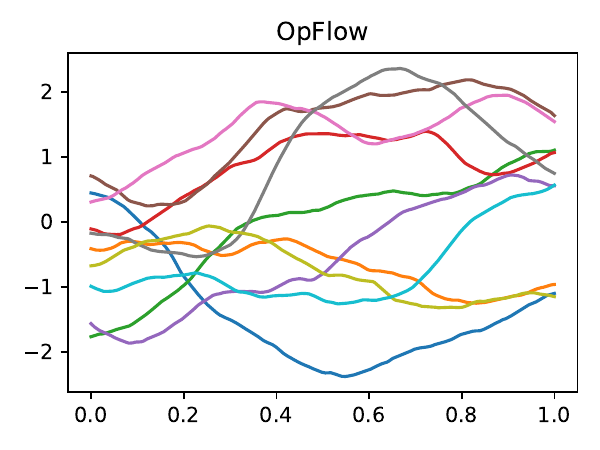}
        \vspace*{-.4cm}
        \caption{}
    \end{subfigure}
    \begin{subfigure}{0.14\textwidth}
        \includegraphics[height=.72\textwidth,trim={0 .4cm 0 0},clip]{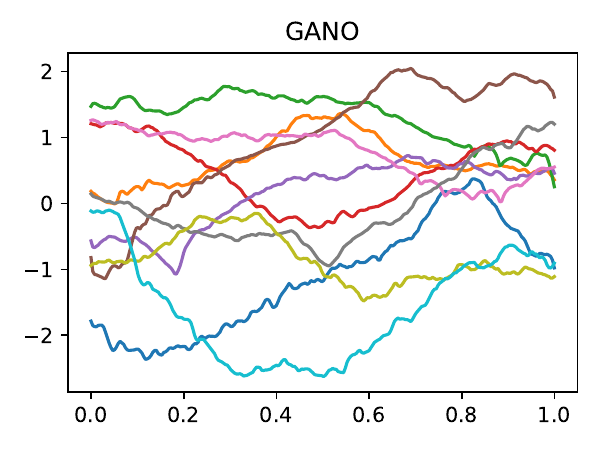}
        \vspace*{-.4cm}
        \caption{}
    \end{subfigure}
    \begin{subfigure}{0.14\textwidth}
        \includegraphics[height=.81\textwidth]{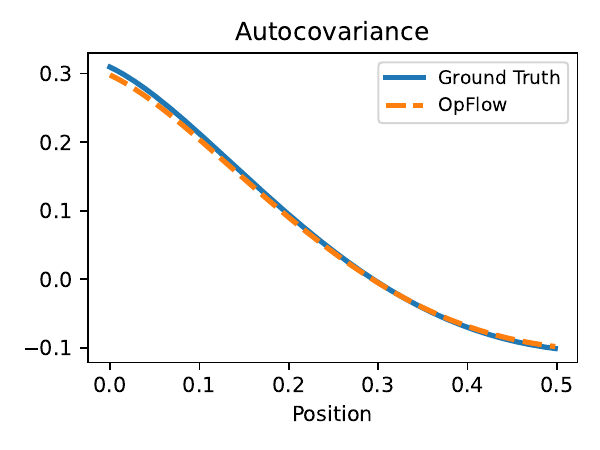}
        \vspace*{-.6cm}
        \caption{}
    \end{subfigure}
    \begin{subfigure}{0.14\textwidth}
        \includegraphics[height=.81\textwidth]{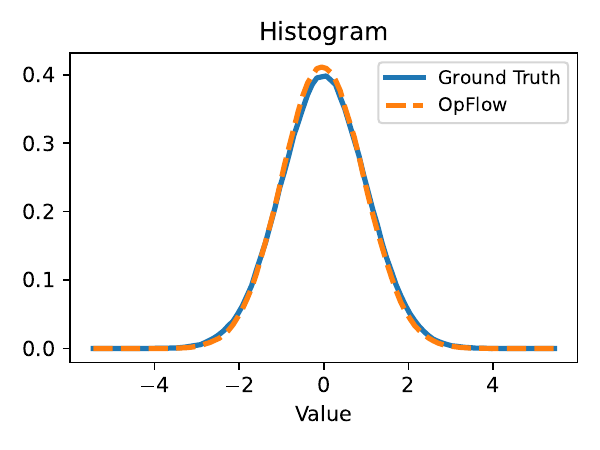}
        \vspace*{-.6cm}
        \caption{}
    \end{subfigure}
    \begin{subfigure}{0.14\textwidth}
        \includegraphics[height=.81\textwidth]{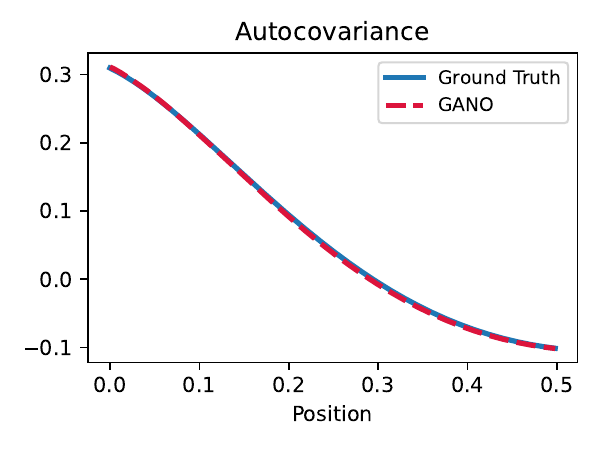}
        \vspace*{-.6cm}
        \caption{}
    \end{subfigure}
    \begin{subfigure}{0.14\textwidth}
        \includegraphics[height=.81\textwidth]{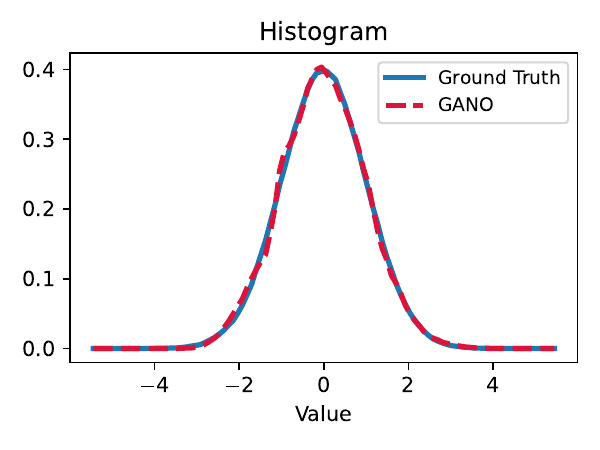}
        \vspace*{-.6cm}
        \caption{}
    \end{subfigure}
    \caption{\alg generation performance on GP data. (a) Samples from the ground truth. (b) Samples from \alg. (c) Samples from GANO. (d) Autocovariance of samples from \alg. (e) Histogram of samples from \alg. (f) Autocovariance of samples from GANO. (g) Histogram of samples from GANO.}
    \label{fig:domain_decom_GP_gen}
\end{figure*}

\textbf{GP data.}
For both \alg and GANO, we set Matern kernel parameters with $l_{\mathcal{U}} = 0.5, \nu_{\mathcal{U}}=1.5$ for the observed data space. And $l_{\mathcal{A}} = 0.05, \nu_{\mathcal{A}}=0.5$ for the latent Gaussian space of \alg, $\alpha=1.5, \tau=1.0$ for the latent Gaussian space of GANO, where $\alpha, \tau$ are the coefficient parameter and inverse length scale parameter of the specific implementation of Matérn based Gaussian
process used in GANO paper~\citep{rahman_generative_2022, nelsen_random_2021}. Data resolution used here is 256. Fig. \ref{fig:domain_decom_GP_gen} shows that \alg can generate realistic samples, virtually consistent with the true data. In contrast, GANO's samples exhibit small high frequency artifacts. Both \alg and GANO successfully recover the statistics with respect to the autovariance function and histogram of the point-wise value (amplitude distribution). We note that there is a significant difference in model complexity here, as \alg has 15M parameters whereas GANO has only 1.4M. The appendix shows a super-resolution experiment for this dataset.

\begin{figure*}[h]
\vspace*{-.2cm}
    \centering
    \hspace{-.6cm}
    \begin{subfigure}{0.14\textwidth}
        \includegraphics[height=.72\textwidth,,trim={0 .4cm 0 0},clip]{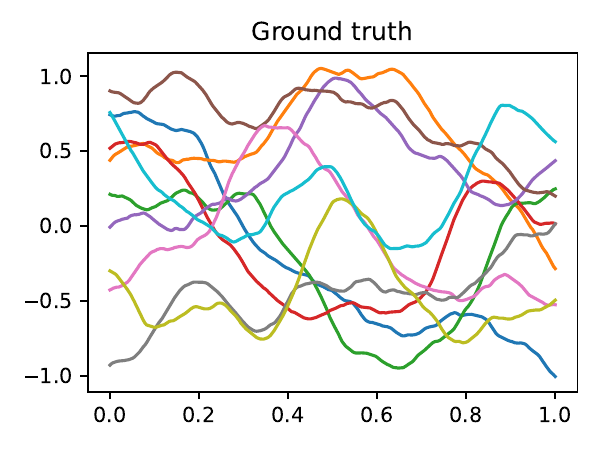}
        \vspace*{-.4cm}
        \caption{}
    \end{subfigure}
    \begin{subfigure}{0.14\textwidth}
        \includegraphics[height=.72\textwidth,trim={0 .4cm 0 0},clip]{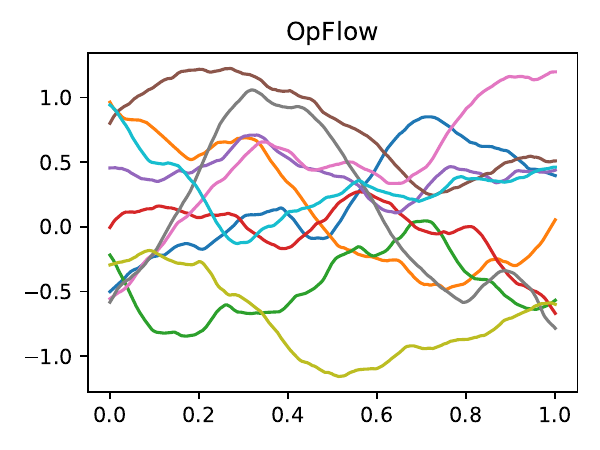}
        \vspace*{-.4cm}
        \caption{}
    \end{subfigure}
    \begin{subfigure}{0.14\textwidth}
        \includegraphics[height=.72\textwidth,trim={0 .4cm 0 0},clip]{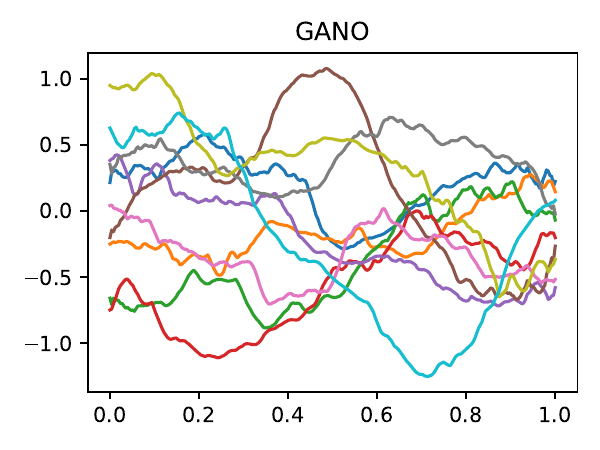}
        \vspace*{-.4cm}
        \caption{}
    \end{subfigure}
    \begin{subfigure}{0.14\textwidth}
        \includegraphics[height=.81\textwidth]{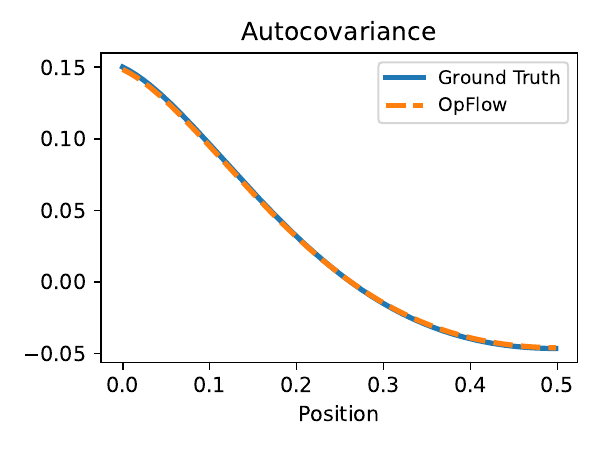}
        \vspace*{-.6cm}
        \caption{}
    \end{subfigure}
    \begin{subfigure}{0.14\textwidth}
        \includegraphics[height=.81\textwidth]{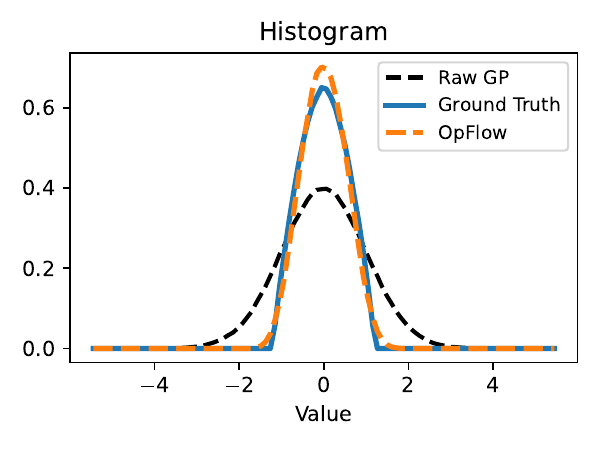}
        \vspace*{-.6cm}
        \caption{}
    \end{subfigure}
    \begin{subfigure}{0.14\textwidth}
        \includegraphics[height=.81\textwidth]{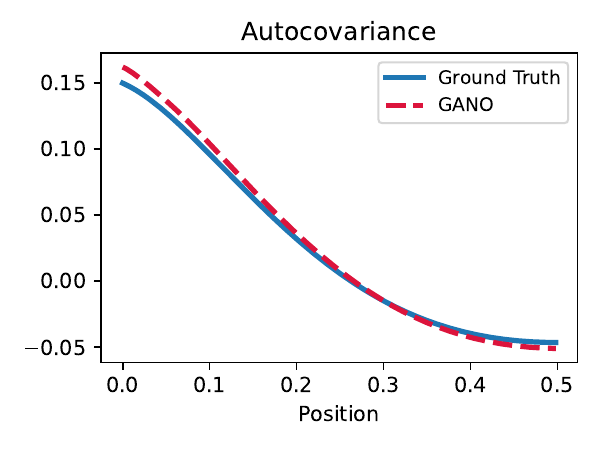}
        \vspace*{-.6cm}
        \caption{}
    \end{subfigure}
    \begin{subfigure}{0.14\textwidth}
        \includegraphics[height=.81\textwidth]{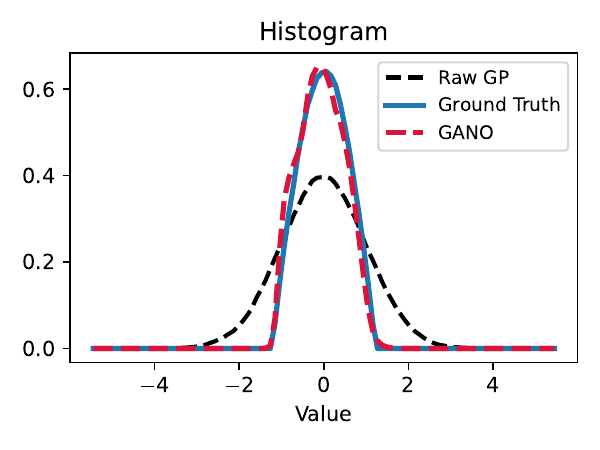}
        \vspace*{-.6cm}
        \caption{}
    \end{subfigure}
    \quad

    \caption{\alg generation performance on TGP data. (a) Samples from the ground truth. (b) Samples from \alg. (c) Samples from GANO. (d) Autocovariance of samples from \alg. (e) Histogram of samples from \alg. (f) Autocovariance of samples from GANO. (g) Histogram of samples from GANO.}
    \label{fig:domain_decom_TGP_gen}
\end{figure*}

\textbf{TGP data.}
We maintain the same parameter settings as used in the TGP functional regression experiment, with bounds set between -1.2 and 1.2, and Matern kernel parameters $l_{\mathcal{U}} = 0.5, \nu_{\mathcal{U}}=1.5$ for the observed space and $l_{\mathcal{A}} = 0.1, \nu_{\mathcal{A}}=0.5$ for the latent space of \alg, $\alpha=1.5, \tau=1.0$ for the latent Gaussian space of GANO. The only difference is that we increase the resolution to 256 from 128. Fig. \ref{fig:domain_decom_TGP_gen} demonstrates that \alg is capable of generating realistic samples that closely mimic the ground truth samples, while GANO tends to produce samples that appear rougher in texture. Details on the super-resolution experiment for this task are available in the appendix.

\begin{figure*}[h]
\vspace*{-.2cm}
    \centering
    \hspace*{-.8cm} 
    \begin{subfigure}{0.18\textwidth}
        \includegraphics[height=1.\textwidth]{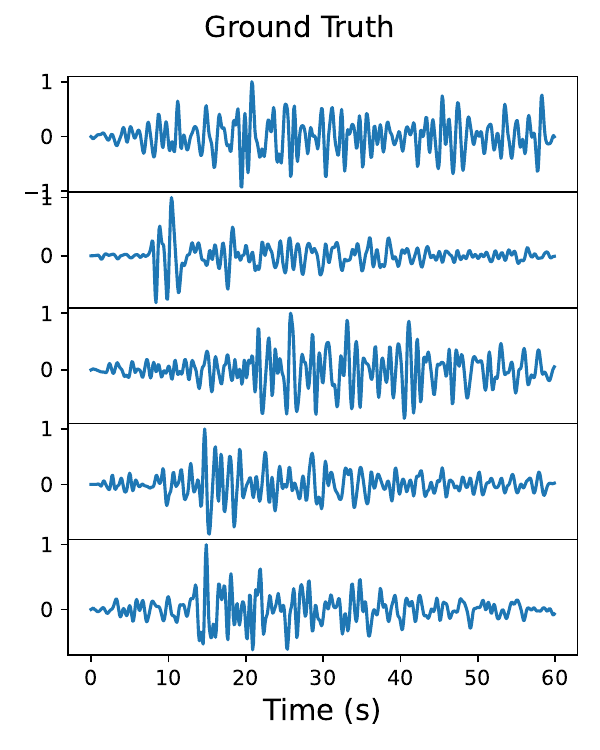}
        \vspace*{-.3cm}        
        \caption{}
    \end{subfigure}
    \begin{subfigure}{0.18\textwidth}
        \includegraphics[height=1.\textwidth]{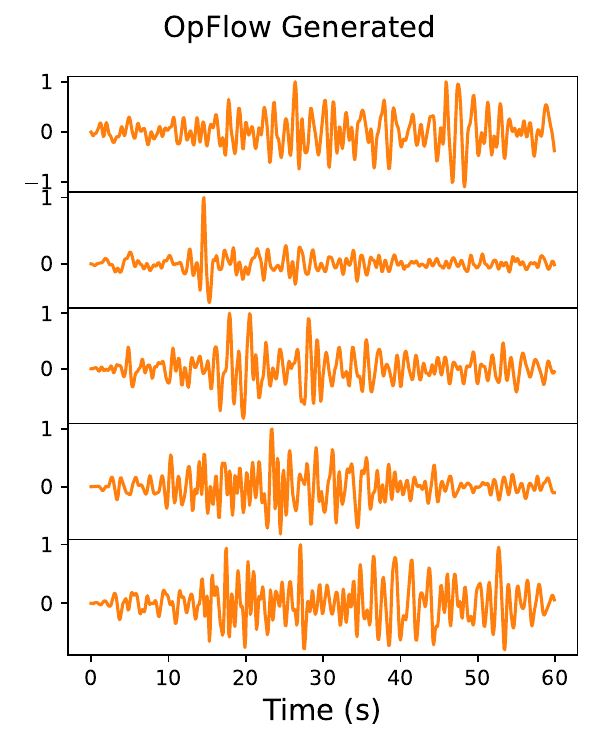}
        \vspace*{-.3cm}
        \caption{}
    \end{subfigure}
    \begin{subfigure}{0.18\textwidth}
        \includegraphics[height=1.\textwidth]{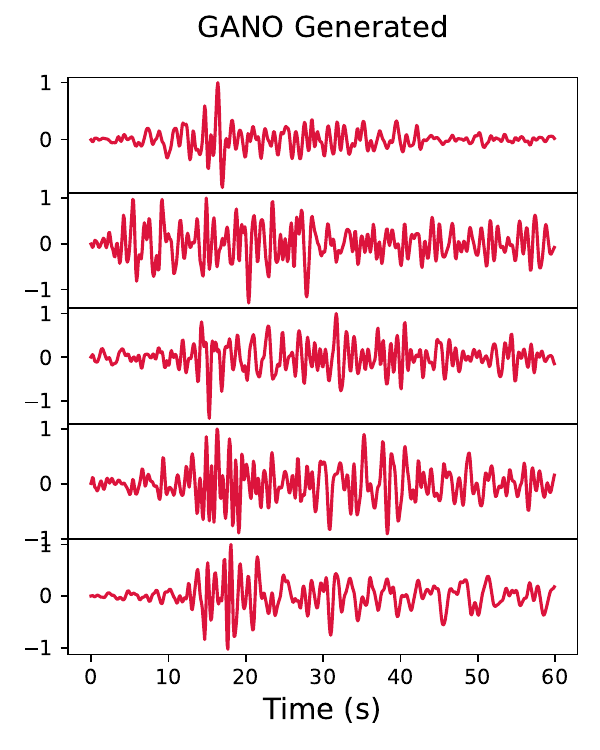}
        \vspace*{-.3cm}
        \caption{}
    \end{subfigure}
    \hspace{-.4cm}
    \begin{subfigure}{0.2\textwidth}
        \includegraphics[height=0.81\textwidth,trim={0 1cm 0 0},clip]{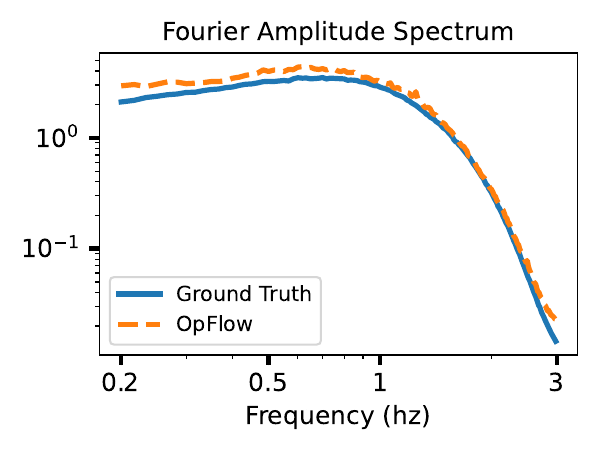}
        \caption{}
    \end{subfigure}
    \quad
    \begin{subfigure}{0.2\textwidth}
        \includegraphics[height=0.81\textwidth,trim={0 1cm 0 0},clip]{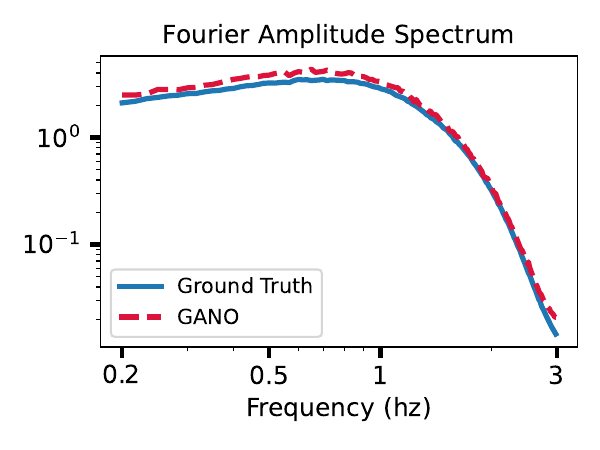}
        \caption{}
    \end{subfigure}
    
    \caption{\alg generation performance on seismic velocity data from kik-net. (a) Samples from the ground truth. (b) Samples from \alg. (c) Samples from GANO. (d) Fourier amplitude spectrum of samples from \alg. (e) Fourier amplitude spectrum of samples from GANO}
    \label{fig:domain_decom_seismic_gen}
\end{figure*}

\textbf{Seismic Data.}
The dataset is the same as used in the functional regression section. However for this test, we no longer use a Gaussian filter to simplify the training dataset, which makes it more challenging to learn. The results are shown in Fig. \ref{fig:domain_decom_seismic_gen}. These figures demonstrate that both \alg and GANO can generate visually realistic seismic waveforms. For validation purposes, we examine the Fourier Amplitude Spectrum, a critical engineering metric~\citep{shi_broadband_2024}, in Fig. \ref{fig:domain_decom_seismic_gen}. The analysis shows that both models accurately learn the frequency content of the seismic velocity dataset, underscoring their potential utility in seismic hazard analysis and engineering design~\citep{shi_broadband_2024}. 

\section{Conclusion}

In this paper, we have introduced Neural Operator Flows (\alg), an invertible infinite-dimensional generative model that allows for universal functional regression. \alg offers a learning-based solution for end-to-end  universal functional regression, which has the potential to outperform traditional functional regression with Gaussian processes. This is in addition to \alg comparing favorably with other algorithms for generation tasks on function spaces. However, we acknowledge that our research primarily focuses on 1D and 2D datasets. Learning operators for functions defined on high-dimensional domains remains challenging and is an area that has not been thoroughly developed. Looking ahead, \alg may allow for new avenues in analysis of functional data by learning directly from the data, rather than assuming the type of stochastic process beforehand.

\section*{Acknowledgements}
The authors would like to express their gratitude to the TMLR reviewers and the action editor for their constructive comments. We also extend our thanks to Chenxi Kong for providing the data for the Rayleigh-Bénard convection problem. Z. Ross was supported by a fellowship from the David and Lucile Packard Foundation.


\bibliography{OpFlow}
\bibliographystyle{tmlr}

\newpage

\appendix
\section{Appendix}
\subsection{Forward and Inverse processes of \alg}
\begin{algorithm}
\caption{Forward and Inverse processes of \alg}
\begin{algorithmic}[1]
\State 
\textbf{Input and Parameters:} Input function $(v')^i$ to the $i$th layer of \alg; $i$th actnorm layer parameterized by $s_{\theta}^i, b_{\theta}^i$; $i$th affine coupling layer $\mathcal{T}_{\theta}^i$; Discretized domain $D_{\theta}$ for $s_{\theta}^i, b_{\theta}^i$ and $D_{s}$ for the output functions of the affine coupling layer
\State \textbf{Forward process:} 
\State \hspace{\algorithmicindent} Input function: $(v')^i$
\State \hspace{\algorithmicindent} Actnorm layer: $v^i = s_{\theta}^i \odot (v')^i + b_{\theta}^i $
\State \hspace{\algorithmicindent} Affine coupling layer: $h_1^i, h_2^i = \mathbf{split}(v^i)$; $ \log(s^i), b^i = \mathcal{T}^i_{\theta}(h_1^i)$; $(h^{\prime})^i_2 = s^i \odot h_2^i + b^i$
\State \hspace{\algorithmicindent} Output function:
$(v^{\prime})^{i+1} = \mathbf{concat}(h_1^i, (h^{\prime})_2^i)$

\State \textbf{Inverse process:} 
\State \hspace{\algorithmicindent} Input function: $(v')^{i+1}$
\State \hspace{\algorithmicindent} Partition: $h_1^i, (h^{\prime})_2^i = \mathbf{split}((v')^{i+1})$

\State \hspace{\algorithmicindent} Inverse of the affine coupling layer: $ \log(s^i), b^i = \mathcal{T}^i_{\theta}(h_1^i)$;
$h_2^i = ((h^{\prime})^i_2 - b^i)/s^i$; $v^i = \mathbf{concat}(h^i_1, h^i_2)$

\State \hspace{\algorithmicindent} Inverse of the actnorm layer: $(v')^i = (v^i - b_{\theta}^i)/s_{\theta}^i $

\State \hspace{\algorithmicindent} Output function:$(v')^i$

\State \textbf{Log-determinant:}
\State \hspace{\algorithmicindent} Actnorm layer: $\textbf{sum}(\log(s_{\theta}^i|_{D_{\theta}}))$
\State \hspace{\algorithmicindent} Affine coupling layer:
$\textbf{sum}(\log(s^i|_{D_s}))$

\end{algorithmic}
\label{algo:forward_inverse}
\end{algorithm}
\subsection{Discretization agnostic property of \alg}
In this part, we first expand the definition of neural operators and elaborate on their discretization-invariant properties, then explain why \alg is also discretization invariant.

Given two separable Banach spaces $\mathcal{A}$ and $\mathcal{U}$, a neural operator (denoted as $\mathcal{G}_{\theta}$) learns mappings between two function spaces. $\mathcal{G}_{\theta} : \mathcal{A} \rightarrow \mathcal{U}$, involving input and output functions on bounded domain in real-valued spaces. Thus for an input function $a \in \mathcal{A}$, we have $\mathcal{G}_{\theta}(a) = u$, where $u \in \mathcal{U}$. The architecture of neural operators is usually composed of three parts~\citep{kovachki_neural_2023} : (1) \textbf{Lifting:} Converts input functions \( a: \mathcal{D} \rightarrow \mathbb{R}^{d_a} \) into a higher-dimensional space \( \mathbb{R}^{d_{v_0}} \) using a local operator, with \( d_{v_0} > d_a \). (2) \textbf{Kernel Integration:} For \( i = 0 \) to \( T-1 \), map hidden state \( v_i \) to the next \( v_{i+1} \) by a combination of local linear, non-local integral operator and a bias function. (3) \textbf{Projection:} The final hidden state \( v_T \) is projected to the output space \( \mathbb{R}^{d_u} \) through a local operator, where \( d_{v_T} > d_u \). 

The lift and projection parts in neural operators are usually shallow neural networks, such as Multilayer Perceptrons. For $i$th kernel integration layer, we have the integral kernel operator $\mathcal{K}_i$ with corresponding kernel function $k_i \in C(\mathcal{D}_{i+1}\times\mathcal{D}_i;   \mathbb{R}^{d_{v_{i+1}}\times d_{v_i}}) $, where $v_{i+1}(y), v_{i}(x)$ are defined on domains $\mathcal{D}_{i+1}, \mathcal{D}_i$, respectively. Denote $\mu_i$ as the measure on $\mathcal{D}_i$, $W_i$ as the local linear opeartor, $b_i$ as the bias function. Then for the $i$th kernel integration layer, we have 
\begin{equation}
v_{i+1}(y) = \int_{D_i} \kappa_i(x, y)v_i(x)d\mu_i(x) + W_i v_i(y) + b_i(y), \quad \forall y \in \mathcal{D}_{i+1} 
\label{kernel_int}
\end{equation}
Fourier Neural Operator~\citep{li_fourier_2021} simplifies Eq \ref{kernel_int} using Fourier transform and achieves quasi-linear time complexity, which can be expressed as
\begin{equation}
    v_{i+1} = \mathcal{F}^{-1}(R_{i}(\mathcal{F}v_i))  + W_i v_i + b_i
\label{kernel_FNO}
\end{equation}
where $R_i$ are learnable parameters defined in Fourier space associated with user-defined Fourier modes, $\mathcal{F}$ and $\mathcal{F}^{-1}$ are Fourier transformation and inverse Fourier transformation respectively. The calculation of Eq \ref{kernel_int},\ref{kernel_FNO} for data with arbitrary discretization involves Riemann summation. For more details of calculation of the kernel integration, please refer to~\cite{kovachki_neural_2023, li_fourier_2021, rahman_generative_2022}.

Furthermore, the formulation of \alg is for function spaces, all components and operations invovled in \alg are directly defined on function space. The likelihood computation in Eq. \ref{loss:basic} can take place for any discretization, and moreover, the 2-Wasserstein distance in Eq. \ref{eq:w2dist} is derived for function spaces, which we show how to compute on given discretizations in Eq. \ref{eq:gp2dist_matrix}. Therefore \alg is discretization agnostic. We further use a series of zero-shot super-resolution experiments in Appendix A.7 to demonstrate the discretization-invariant feature of \alg.

\subsection{Parameters used in regression tasks}

Table \ref{tab:reg_table} shows the regression parameters (explained in Algorithm \ref{algo:sgld}) for all regression experiments used in this paper. The learning rate is decayed exponentially, which starts from $\{\eta_t, t=0 \}$, ends with $\{\eta_t, t=N\}$
\begin{table}[h]
\centering
\caption{Datasets and regression parameters}
\begin{tabular}{lp{1.5cm}p{2cm}p{2cm}p{2cm}p{2.1cm}p{2.5cm}}
\hline
Datasets & total iterations ($N$) & burn-in iterations ($b$) & sample iterations ($t_N$) & temperature of noise ($T$) & initial learning rate ($\eta_t, t=0$) & end learning rate ($\eta_t, t=N$) \\ \hline
GP & 4e4 & 2e3 & 10 & 1 & 5e-3 & 4e-3\\
TGP & 4e4 & 2e3 & 10 & 1 & 5e-3 & 4e-3\\
GRF & 2e4 & 2e3 & 10 & 1 & 5e-3 & 4e-3\\
Seismic & 4e4 & 2e3 & 10 & 1 &1e-3 & 8e-4\\
Codomain GP & 4e4 & 2e3 & 10 & 1 &5e-5 &4e-5 \\\hline
\end{tabular}
\label{tab:reg_table}
\end{table}

\subsection{Comparison against Deep GPs on Gaussian Process example}\label{Apx:DGP}

In this experiment, we evaluate the regression performance of \alg against that of variational Deep GP~\citep{salimbeni_doubly_2017} and Deep Sigma Point Processes (DSSP)~\citep{jankowiak_deep_2020}. We maintain the same dataset and hyperparameter settings for \alg as those used in the Gaussian process regression example discussed in the main text. The only difference in this experiment is that it includes 10 observations. For the baseline models, we use official implementations from GPyTorch~\citep{gardner_gpytorch_2021} and present the best available results. 

\begin{figure*}[h]
\vspace*{-.2cm}
    \centering
    \begin{subfigure}{0.23\textwidth}
        \includegraphics[height=.73\textwidth]{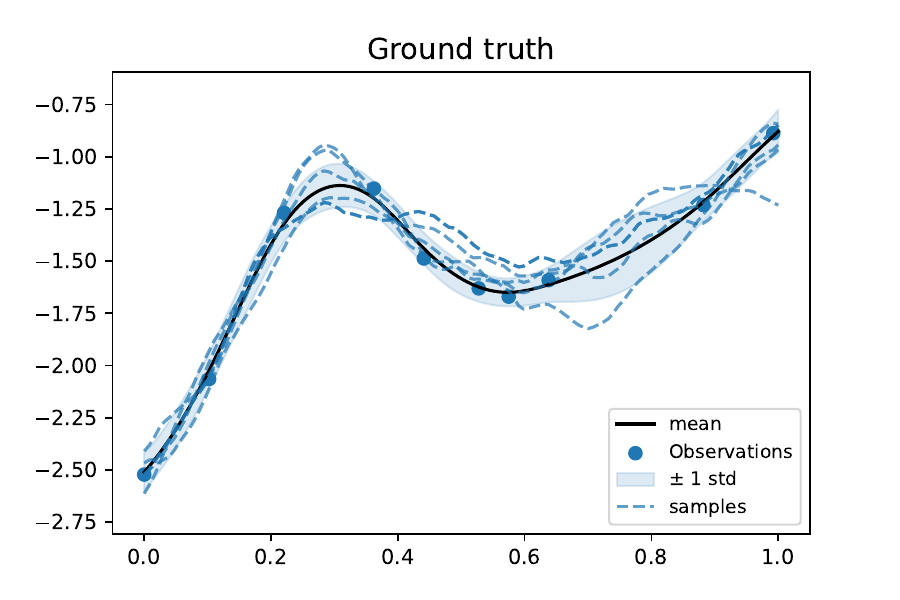}
        \vspace*{-.6cm}
        \caption{}
    \end{subfigure}
    \begin{subfigure}{0.23\textwidth}
        \includegraphics[height=.73\textwidth]{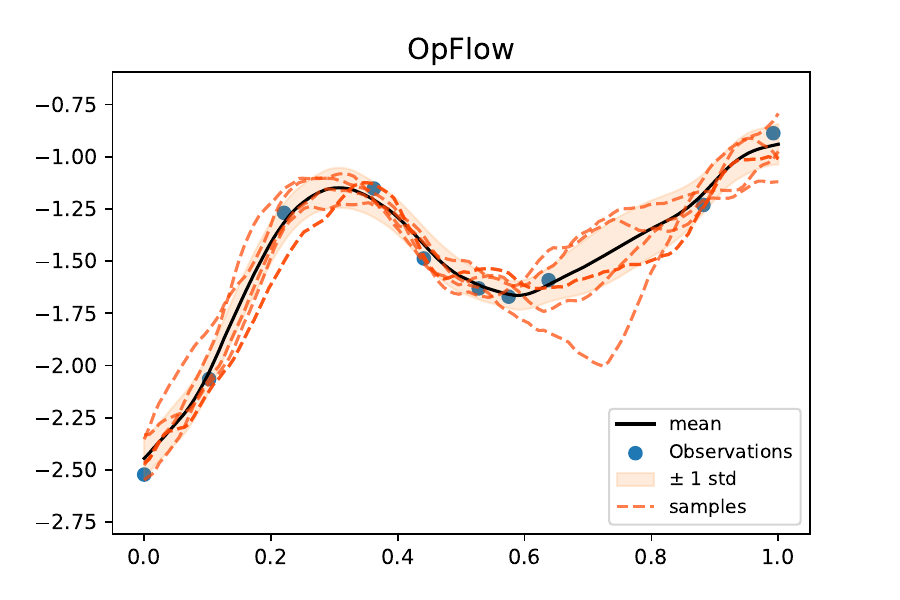}
        \vspace*{-.6cm}
        \caption{}
    \end{subfigure}
    \begin{subfigure}{0.23\textwidth}
        \includegraphics[height=.73\textwidth]{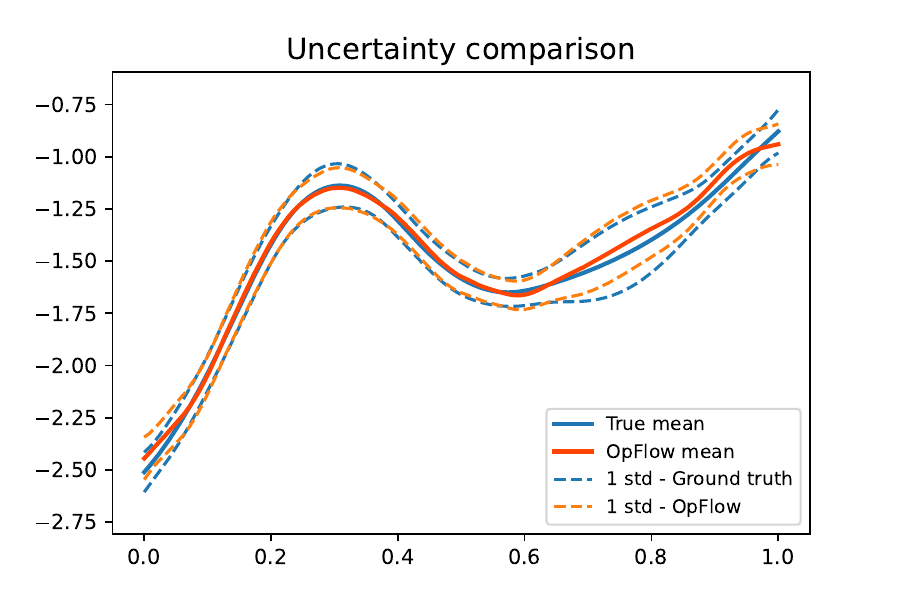}
        \vspace*{-.6cm}
        \caption{}
    \end{subfigure}

    \begin{subfigure}{0.23\textwidth}
        \includegraphics[height=.73\textwidth]{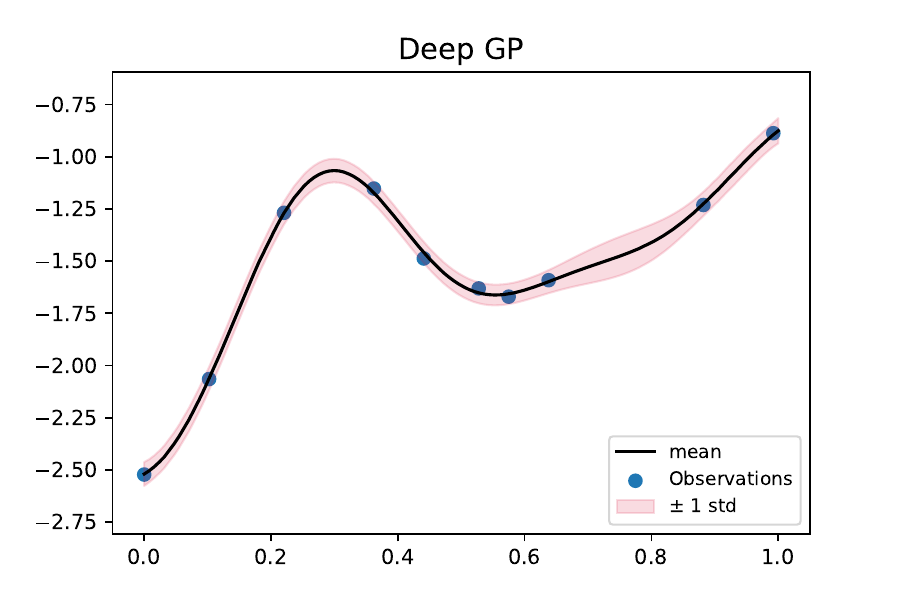}
        \vspace*{-.6cm}
        \caption{}
    \end{subfigure}
    \begin{subfigure}{0.23\textwidth}
        \includegraphics[height=.73\textwidth]{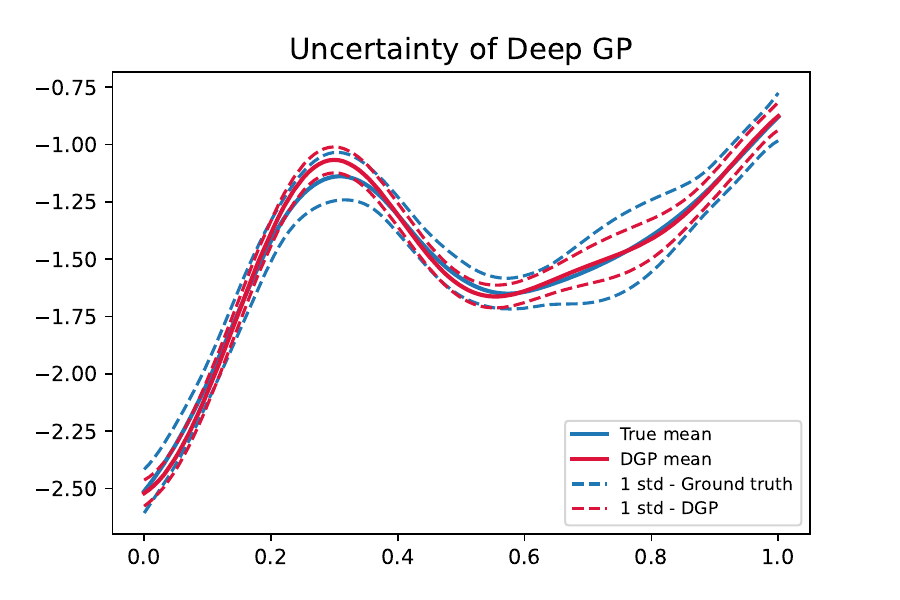}
        \vspace*{-.6cm}
        \caption{}
    \end{subfigure}
    \begin{subfigure}{0.23\textwidth}
        \includegraphics[height=.73\textwidth]{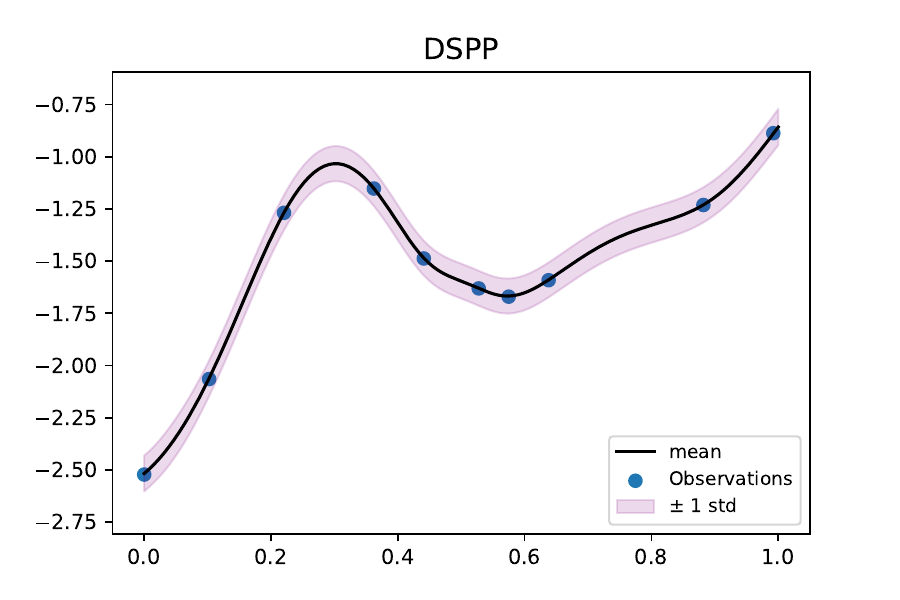}
        \vspace*{-.6cm}
        \caption{}
    \end{subfigure}
    \begin{subfigure}{0.23\textwidth}
        \includegraphics[height=.73\textwidth]{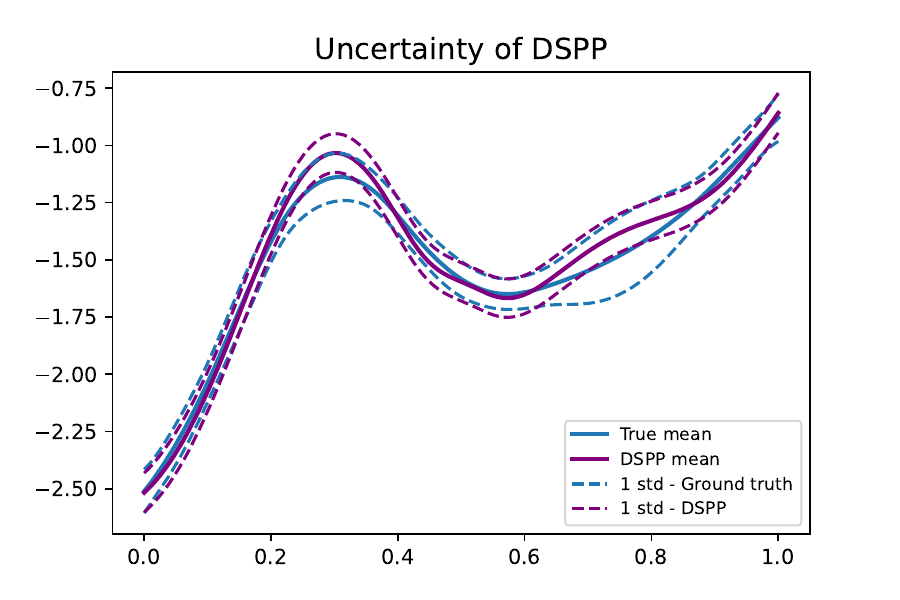}
        \vspace*{-.6cm}
        \caption{}
    \end{subfigure}
    \caption{ \alg regression on GP data. (a) Ground truth GP regression with observed data and predicted samples (b) \alg regression with observed data and predicted samples. (e) Uncertainty comparison between true GP and \alg predictions. (d) Deep GP regression with observed data and predicted samples. (e) Uncertainty comparison between true GP and Deep GP predictions. (f) DSSP regression with observed data and predicted samples. (g) Uncertainty comparison between true GP and DSSP predictions.  }
    \label{fig:DGP_reg}
\end{figure*}

For this comparison, we employ the following metrics: (1) Standardized Mean Squared Error (SMSE), which normalizes the mean squared error by the variance of the ground truth; and (2) Mean Standardized Log Loss (MSLL), first proposed in Equation 2.34 of ~\cite{rasmussen_gaussian_2006}, defined as:
\begin{equation}
    - \log p(\hat{y}|D, \hat{x}) = \frac{1}{2}\log(2\pi\hat{\sigma}^2) + \frac{(y - \hat{y})^2}{2\hat{\sigma}^2}
\end{equation}
where $D$ represents the observation pairs $(x_{obs}, y_{obs})$, and $\hat{x}, \hat{y}, y, \hat{\sigma}^2$ denote the inquired new positions, test data, predicted mean and predicted variance, respectively. We collect a test data contains 3000 true posterior samples for averaging out SMSE the MSLL. As depicted in Fig.\ref{fig:DGP_reg} and Table \ref{tab:DGP_reg_table}, \alg generates realistic posterior samples and achieves lower SMSE and MSLL, demonstrating superior performance over the current state-of-the-art Deep GPs.

\begin{table}[h]
\centering
\caption{Comparison with benchmark GPs}
\begin{tabular}{lp{3cm}p{3cm}p{3cm}}
\hline
Models & SMSE  & MSLL \\ \hline
\alg  & \textbf{0.083} & \textbf{-0.879}\\
Deep GP & 0.084 &  -0.442\\
DSPP & 0.094 & -0.673\\

\end{tabular}
\label{tab:DGP_reg_table}
\end{table}

\subsection{Valid stochastic processes induced by \alg via Kolmogorov extension theorem}\label{Apx:Valid}

In finite dimensional spaces, normalizing flow is provided to have universal representation for any data distribution under some mild conditions~\citep{kong_expressive_2020, papamakarios_normalizing_2021}. Following the fact that universal approximation of normalizing flow, we expect such universal approximation also be generalizable to \alg, similar to its finite-dimensional case. However, providing a rigorous proof of universal approximation of \alg remains beyond the scope of this paper and we leave it for future work. In this paper, we focus on demonstrating that \alg can induce valid stochastic processes and able to learn more general stochastic processes compared to the marginal flow described in Transformed GP. The marginal flow is based on pointwise transformation of GP values, ergo, a pointwise operator. It induces valid stochastic processes and since it is limited to only pointwise transformation of GP values, it  produces transformations with diagonal Jacobians~\citep{maronas_transforming_2021}. In contrast, \alg acts as a more general bijective operator that results in transformations with triangular Jacobian matrix (non-diagonal) and jointly transforms the GP values. We next formally define the problem setting and provide a proof of our statements.

Consider a stochastic process $F_0 : \mathcal{X}_0 \rightarrow \mathcal{Y}_0$. For a finite sequence $x_0^1:x_0^n = (x_0^1, x_0^2, \cdots, x_0^n)$ with $x_0^i \in \mathcal{X}_0$,  we define the marginal joint distribution over the function values $y_0^1:y_0^n = (F_0(x_0^1), \cdots, F_0(x_0^n))$. Within the framework of \alg, $F_0$ is a Gaussian Process  and the joint distribution $p(y_0^1:y_0^n) = p_{x_0^1:x_0^n}(y_0^1:y_0^n)$ is a multivariate Gaussian. With a bijective operator $\mathcal{G}_{\theta}$, we have $G_{\theta}(y_0^1:y_0^n) = y^1:y^n$ and $G_{\theta}^{-1}(y^1:y^n) = (y_0^1:y_0^n)$, where $y^1:y^n$ is defined over $x^1:x^n$. The point alignment between $x^1:x^n$ and $x_0^1:x_0^n$ is determined through an underlying homeomorphic map. According to the Kolmogorov Extension Theorem~\citep{kolmogorov_foundations_2018}, to establish that a valid stochastic process $F$ has $p(y^1:y^n)$ as its finite dimensional distributions, it is essential to demonstrate that such a joint distribution satisfies the following two consistency properties:

\textbf{Permutation invariance.}
Given a permutation $\pi$ of $\{1, \cdots, n\}$, the joint distribution should remain invariant when elements of $\{x^1, \cdots, x^n\}$ are permuted, such that 
\begin{equation}
   p_{x^1:x^n}(y^1:y^n) = p_{x^{\pi(1)}:x^{\pi(n)}}(y^{\pi(1)}:y^{\pi(n)}) 
\end{equation}

\textbf{Consistency in marginalization.} This property states that if a part of the sequence is marginalized out, the marginal distribution remains consistent with the distribution defined on the original sequence, such that for $m \geq 1$
\begin{equation}
    p(y^1:y^n) = \int p(y^1:y^{n+m}) dy^{n+1:n+m}
    \label{KEH:marginal}
\end{equation}

Let's first check the permutation invariance condition. The joint distribution can be expanded as follows:
\begin{equation}
    p(y^1:y^n) = p(y_0^1:y_0^n)|\det(\frac{\partial (y_0^1:y_0^n)}{\partial(y^1:y^n)})|
    \label{KEH:base}    
\end{equation}

Given that $\mathcal{G}_{\theta}$ consists of $t$ invertible layers, we have $\mathcal{G}_{\theta} = \mathcal{G}_{\theta}^t\circ   \cdots \mathcal{G}_{\theta}^2 \circ 
 \mathcal{G}_{\theta}^1$ that gradually transforms $(y_0^1:y_0^n)$ to $(y_1^1:y_1^n), (y_2^1:y_2^n)\cdots (y^1:y^n)$. Since \alg generalize RealNVP structure, the Jacobian matrix for transformations between any two adjacent layers is triangular, which implies the determination of the Jacobian matrix is the product of its diagonal components (Eq. 8. of~\citep{kingma_glow_2018}), then $|\det(\frac{\partial (y_i^1:y_i^n)}{\partial(y_{i+1}^1:y_{i+1}^n)})| = \Pi_{j=1}^{n}|\frac{d y_i^j}{d y_{i+1}^j}|$ for any $i \in \{0,\cdots, t-1\}$, and $|\det(\frac{\partial (y_0^1:y_0^n)}{\partial(y^1:y^n)})| = \Pi_{j=1}^n |\frac{d y_0^j}{d y^j}|$. Thus Eq.  \ref{KEH:base} can be simplified as:
 \begin{equation}
     p(y^1:y^n) = p(y_0^1:y_0^n)\Pi_{j=1}^n |\frac{d y_0^j}{d y^j}|
    \label{KEH:simplified}     
 \end{equation}

 For this analysis, we adopt the term "volume" to refer to the  determinant of Jacobian matrix discussed in in Eq. \ref{KEH:base} or \ref{KEH:simplified}, align with the terminology used in RealNVP~\citep{dinh_density_2017}.  From Eq.\ref{KEH:simplified}, we have the joint distribution under permutation $\pi$ as $p(y^{\pi(1)}:y^{\pi(n)}) = p(y_0^{\pi(1)}:y_0^{\pi(k)})\Pi_{j=1}^n |\frac{d y_0^{\pi(j)}}{d y^{\pi(j)}}|$. Given that $p(y_0^1:y_0^k)$ is a multivariate Gaussian, which is inherently permutation invariant, the volume $\Pi_{j=1}^n|\frac{d y_0^{\pi(j)}}{d y^{\pi(j)}}|$ is independent of the order of variables, which is also permutation invariant. Thus, we have:
 \begin{equation}
  p(y^{\pi(1)}:y^{\pi(n)}) = p(y_0^{\pi(1)}:y_0^{\pi(k)})\Pi_{j=1}^n |\frac{d y_0^{\pi(j)}}{d y^{\pi(j)}}| = p(y_0^1:y_0^n)\Pi_{j=1}^n |\frac{d y_0^j}{d y^j}| = p(y^1:y^n) 
 \end{equation}
 In this way, we verify the permutation invariant property. Intuitively, this can be explained by the fact that the joint distribution $p(y^1:y^n)$ is fully determined by a multivariate Gaussian and the volume. Under permutation of variables, the both the multivariate Gaussian and volume remain unchanged.

Now, let's verify the consistency in the marginalization. According to Eq.\ref{KEH:simplified}, the product $\Pi_{j=1}^{n+m}|\frac{d y_0^j}{d y^j}|$ is separable and since $p(y_0^1:y_0^{m+n})$ is a multivariate Gaussian determined from a Gaussian Process, the marginalization as specified in Eq.\ref{KEH:marginal} naturally holds, which verifies valid stochastic process can be induced from \alg. Last, the Jacobian matrix involved in \alg is a triangular matrix, which allows \alg to model more general stochastic processes compared to the marginal flow used in~\citep{maronas_transforming_2021},where the Jacobian of transformation is diagonal matrix.

\subsection{\alg regression on Gaussian random fields example}
\begin{figure*}[h]
\vspace*{-.2cm}
\hspace*{-.5cm}  
    \centering
    \begin{subfigure}{0.14\textwidth}
        \includegraphics[height=0.8\textwidth,trim={0, 0.4cm, 0, 0},clip]{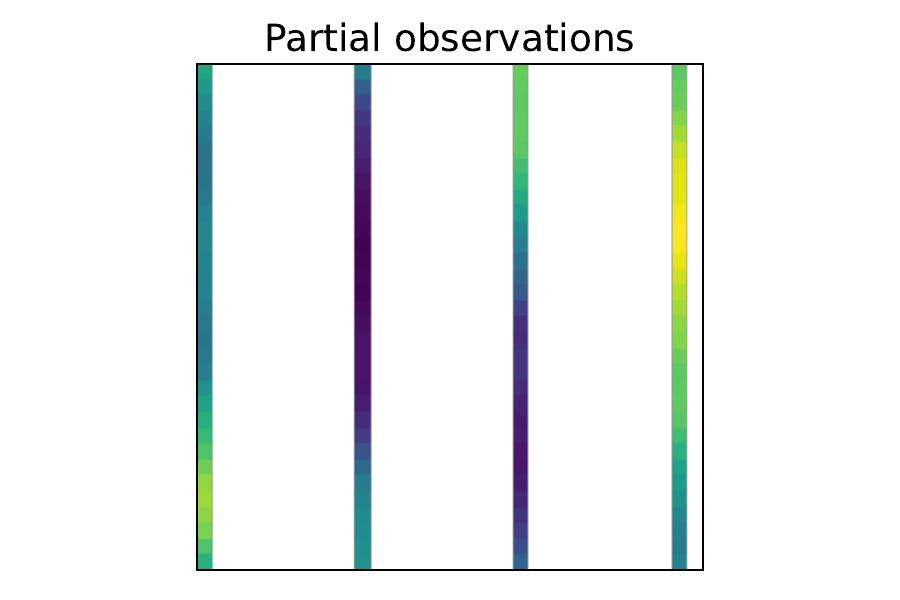}    
        \caption{}
    \end{subfigure}
    \begin{subfigure}{0.14\textwidth}
        \centering
        \includegraphics[height=.8\textwidth]{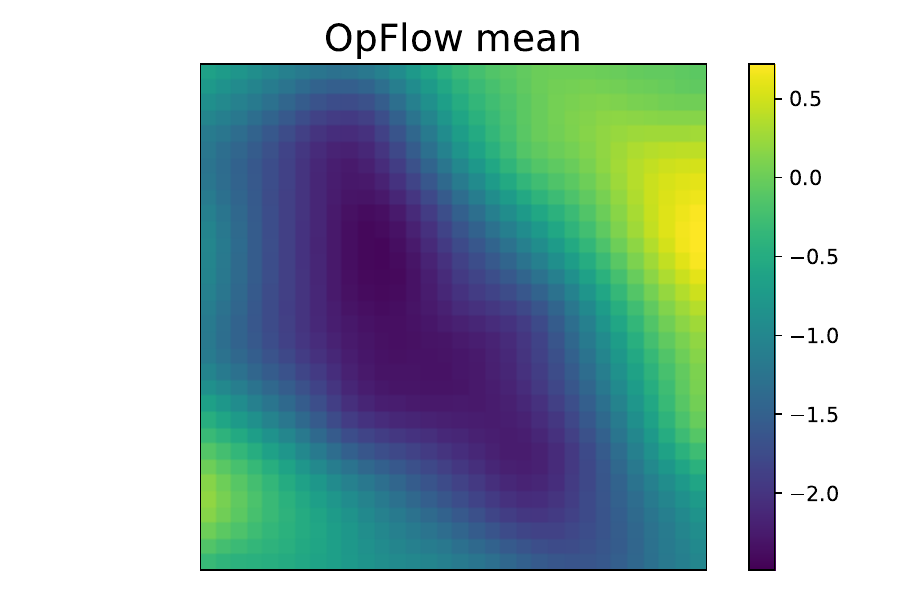}
        \vspace*{-.5cm}
        \caption{}
    \end{subfigure}
    \begin{subfigure}{0.14\textwidth}
        \includegraphics[height=.8\textwidth]{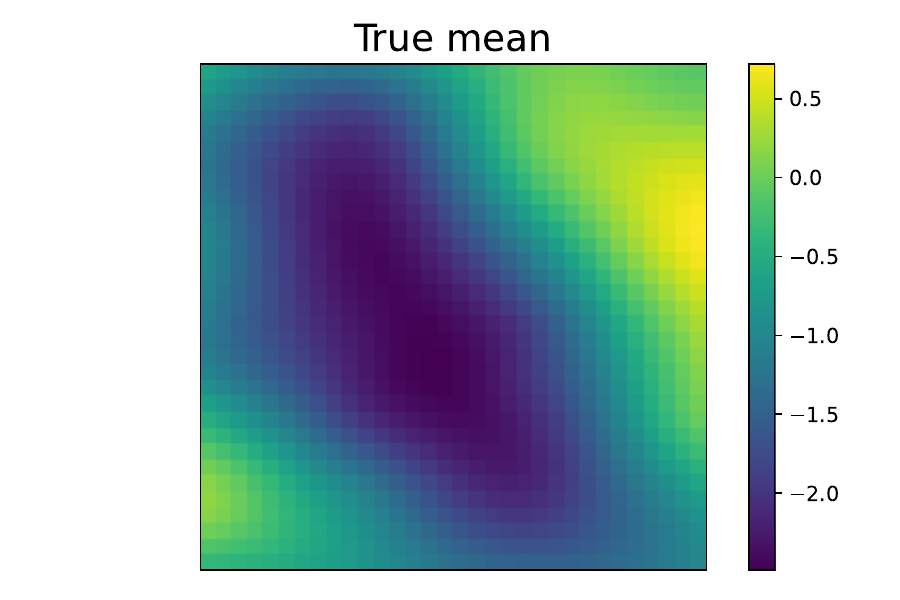}
        \vspace*{-.5cm}
        \caption{}
    \end{subfigure}
    \begin{subfigure}{0.14\textwidth}
        \includegraphics[height=.8\textwidth]{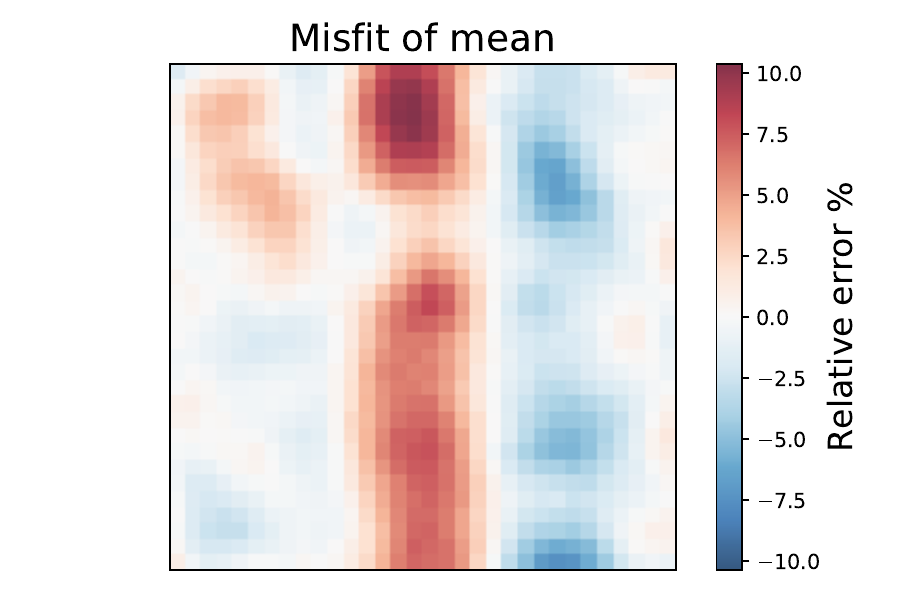}
        \vspace*{-.5cm}
        \caption{}
    \end{subfigure}
    \begin{subfigure}{0.14\textwidth}
        \includegraphics[height=.8\textwidth]{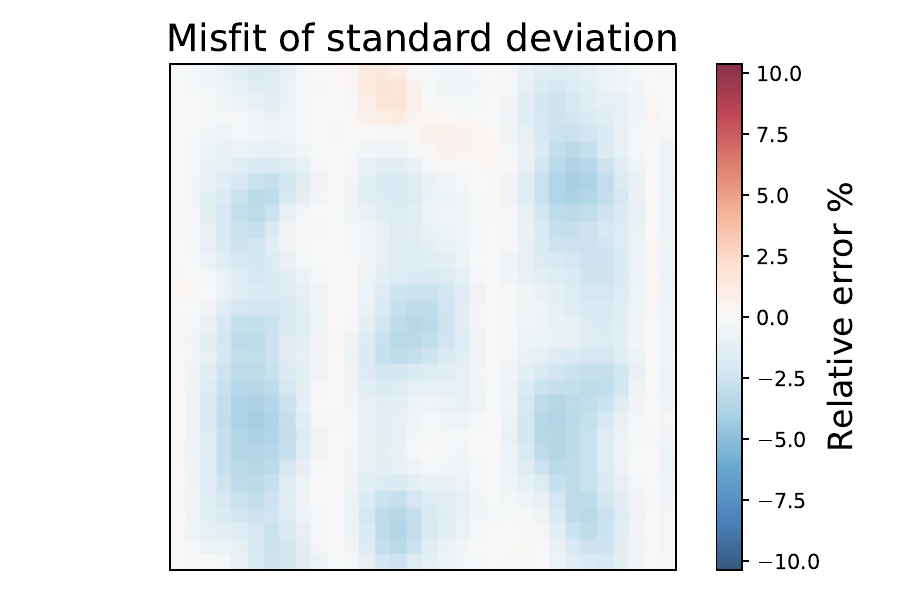}
        \vspace*{-.5cm}
        \caption{}
    \end{subfigure}
    
    \begin{subfigure}{0.35\textwidth}
        \includegraphics[height=.4\textwidth]{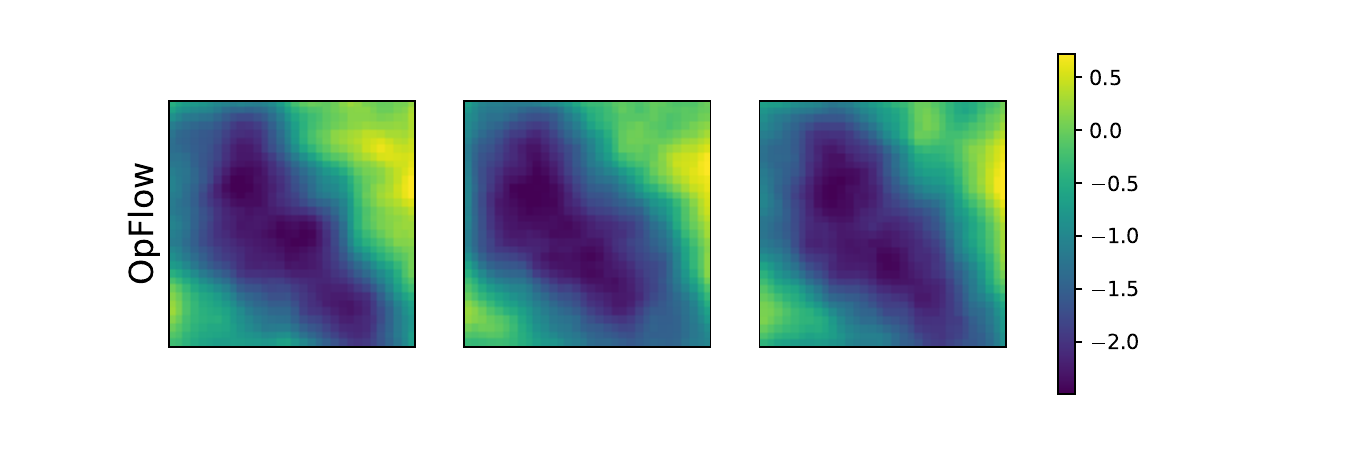}
        \vspace*{-.8cm}
        \caption{}
    \end{subfigure}
    \begin{subfigure}{0.35\textwidth}
        \includegraphics[height=.4\textwidth]{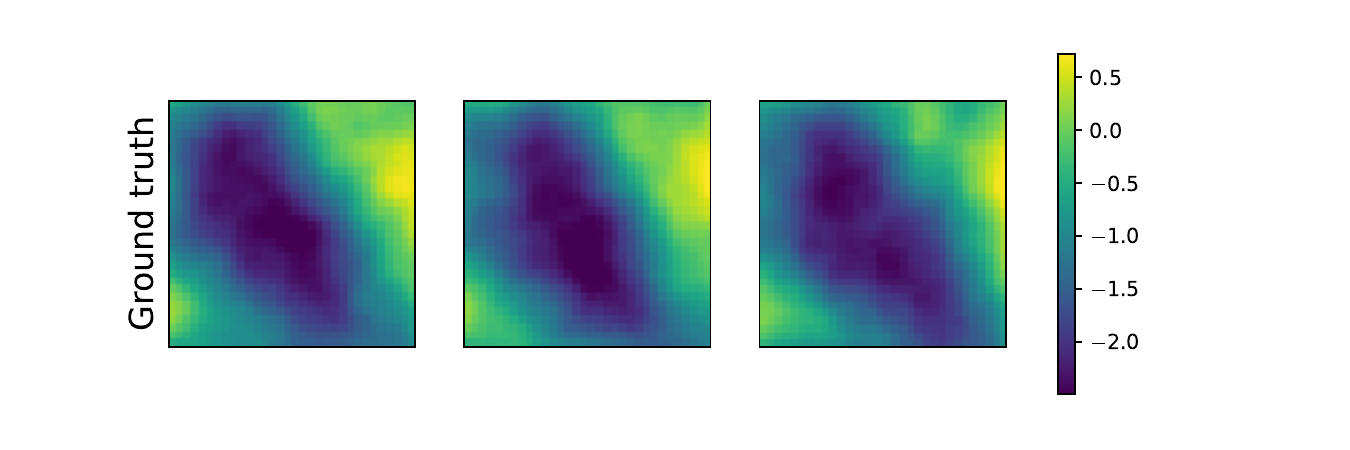}
        \vspace*{-.8cm}
        \caption{}
    \end{subfigure}

    \caption{ \alg regression on 32$\times$32 GRF data. (a) 4 strips observations. (b) Predicted mean from \alg. (c) Ground truth mean from GP regression. (d) Misfit of the predicted mean. (e) Misfit of predicted uncertainty. (f) Predicted samples from \alg. (g) Predicted samples from GP regression. }
    \label{fig:domain_decom_GRF_reg3_reg}
\end{figure*}

This is the second Gaussian random field regression example mentioned in Section 5.1. As depicted in Fig. \ref{fig:domain_decom_GRF_reg3_reg}, \alg faithfully approximates the true posterior given the observations are concentrated along linear patterns. The relative error is the error normalized by the absolute max value of ground truth mean from GP regression. 

\subsection{More generation experiments with domain decomposed \alg}

\begin{figure*}[h]
\vspace*{-.2cm}
    \centering
    \hspace{-.4cm}
    \begin{subfigure}{0.33\textwidth}
        \includegraphics[height=.25\textwidth,trim={4cm .5cm 4cm .5cm},clip]{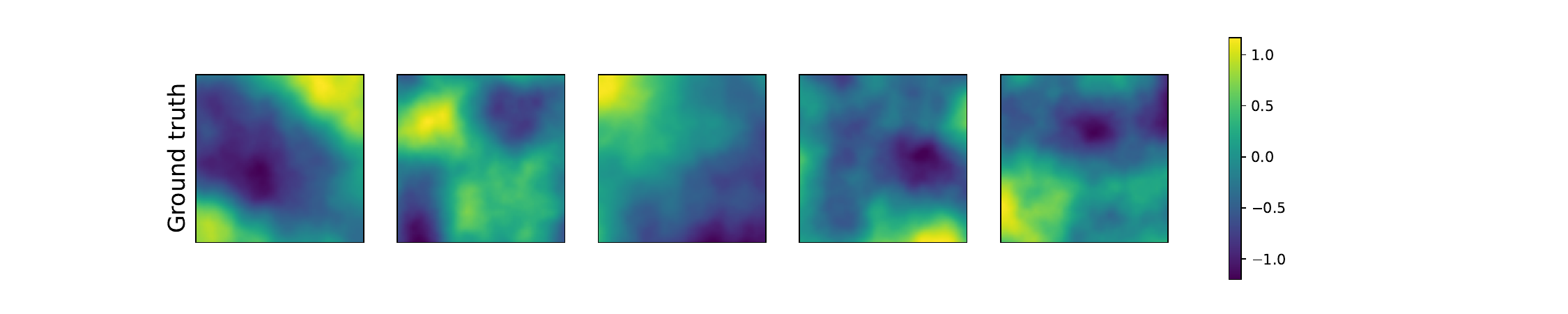}
        \vspace*{-.6cm}
        \caption{}
    \end{subfigure}
    \hspace{-.3cm}
    \begin{subfigure}{0.33\textwidth}
    \includegraphics[height=.25\textwidth,trim={4cm .5cm 4cm .5cm},clip]{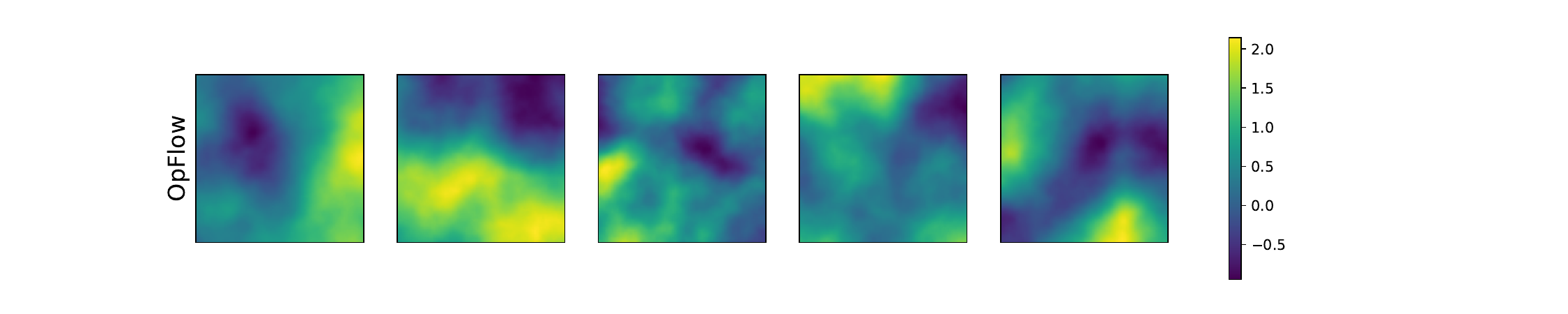}
        \vspace*{-.6cm}
        \caption{}
    \end{subfigure}
    \hspace{-.3cm}
    \begin{subfigure}{0.33\textwidth}
        \includegraphics[height=.25\textwidth,trim={4cm .5cm 4cm .5cm},clip]{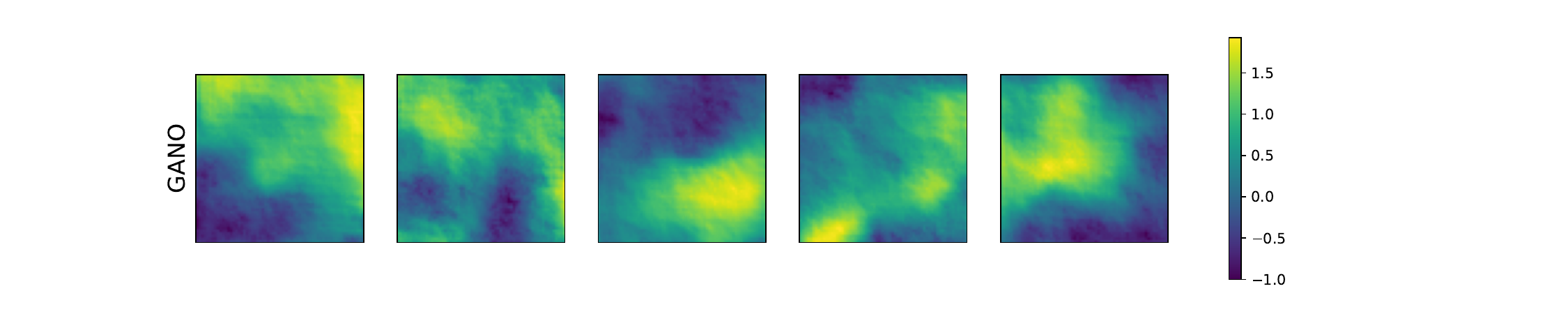}
        \vspace*{-.6cm}
        \caption{}
    \end{subfigure}

    \hspace{-1.2cm}
    \begin{subfigure}{0.22\textwidth}
        \includegraphics[height=.80\textwidth]{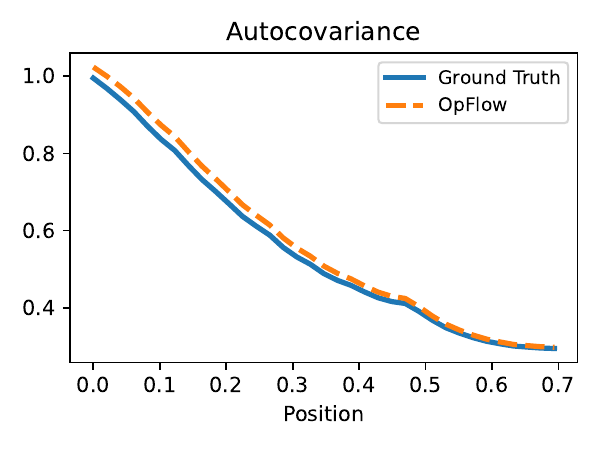}
        \vspace*{-.6cm}
        \caption{}
    \end{subfigure}
    \begin{subfigure}{0.22\textwidth}
        \includegraphics[height=.80\textwidth]{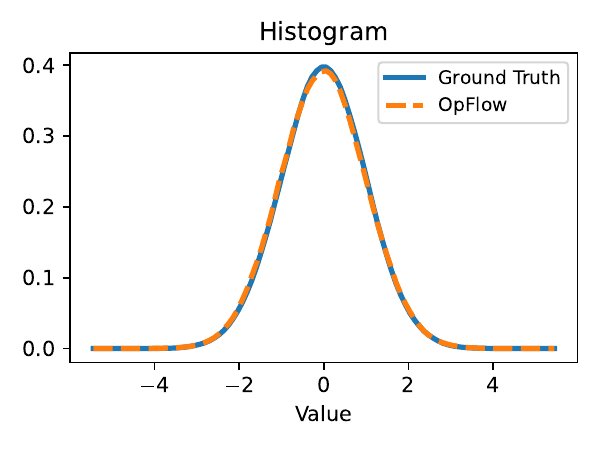}
        \vspace*{-.6cm}
        \caption{}
    \end{subfigure}
    \begin{subfigure}{0.22\textwidth}
        \includegraphics[height=.80\textwidth]{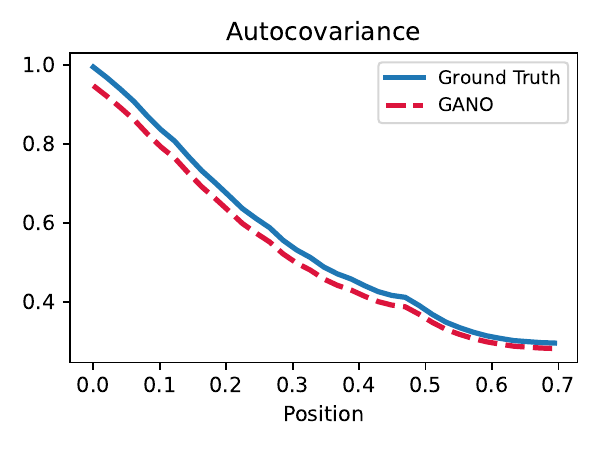}
        \vspace*{-.6cm}
        \caption{}
    \end{subfigure}
    \begin{subfigure}{0.22\textwidth}
        \includegraphics[height=.80\textwidth]{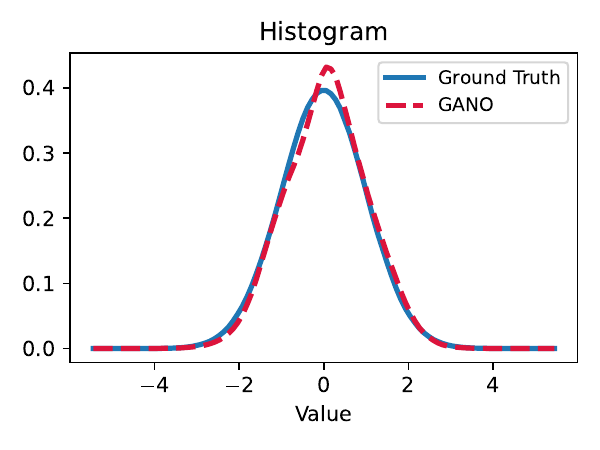}
        \vspace*{-.6cm}
        \caption{}
    \end{subfigure}
    \caption{\alg generation performance on GRF data. (a) Samples from the ground truth. (b) Samples from \alg. (c) Samples from GANO. (d) Autocovariance of samples from \alg. (e) Histogram of samples from \alg. (f) Autocovariance of samples from GANO. (g) Histogram of samples from GANO.}
    \label{fig:domain_decom_GRF_gen}
\end{figure*}

\textbf{GRF data.} For both \alg and GANO, we take the hyperparameters to be $l_{\mathcal{U}} = 0.5, \nu_{\mathcal{U}}=1.5$ for the observed space and $l_{\mathcal{A}} = 0.1, \nu_{\mathcal{A}}=0.5$ for the GP defined on latent Gaussian space $\mathcal{A}$. A resolution of $64 \times 64$ is chosen. Fig. \ref{fig:domain_decom_GRF_gen} illustrates that \alg successfully generates realistic samples, closely resembling the ground truth in both autocovariance and amplitude distribution.

\begin{figure*}[h]
\vspace*{-.2cm}
    \centering
    \hspace{-.4cm}
    \begin{subfigure}{0.33\textwidth}
        \includegraphics[height=.25\textwidth,trim={4cm .5cm 4cm .5cm},clip]{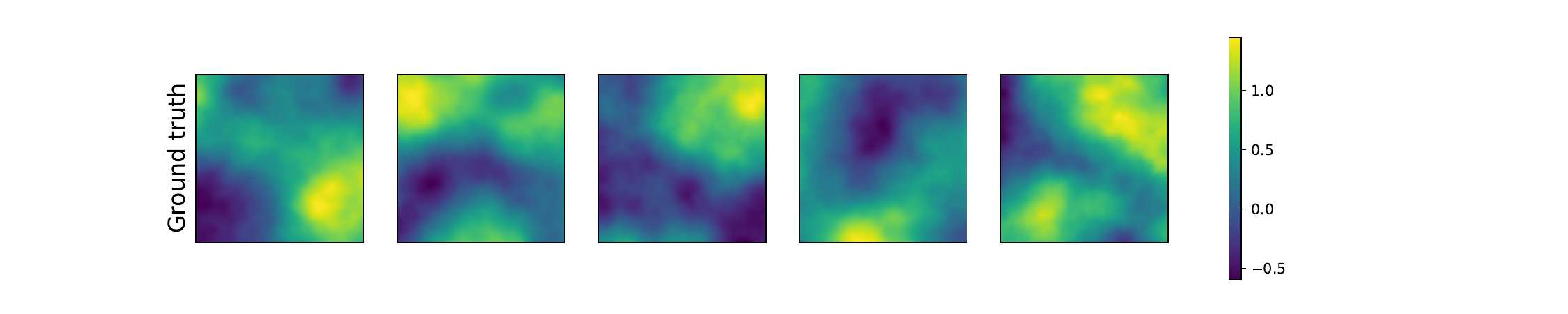}
        \vspace*{-.6cm}
        \caption{}
    \end{subfigure}
    \hspace{-.3cm}
    \begin{subfigure}{0.33\textwidth}
        \includegraphics[height=.25\textwidth,trim={4cm .5cm 4cm .5cm},clip]{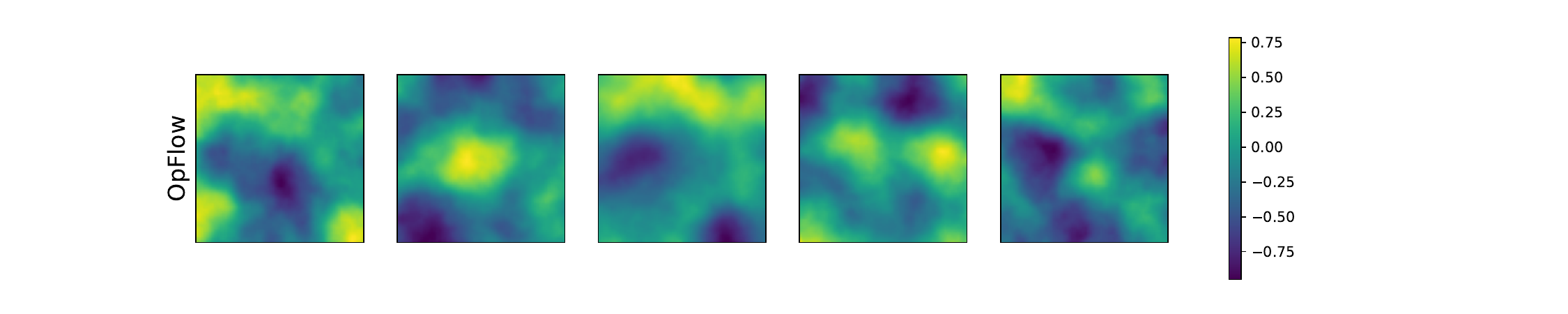}
        \vspace*{-.6cm}
        \caption{}
    \end{subfigure}
    \hspace{-.3cm}
    \begin{subfigure}{0.33\textwidth}
        \centering
        \includegraphics[height=.25\textwidth,trim={4cm .5cm 4cm .5cm},clip]{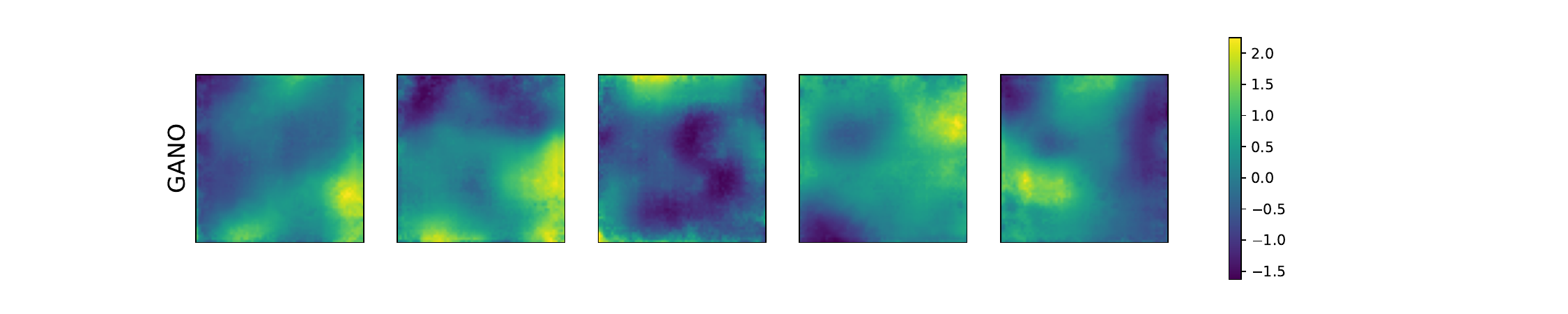}
        \vspace*{-.6cm}
        \caption{}
    \end{subfigure}

    \hspace{-1.2cm}
    \begin{subfigure}{0.22\textwidth}
        \includegraphics[height=.80\textwidth]{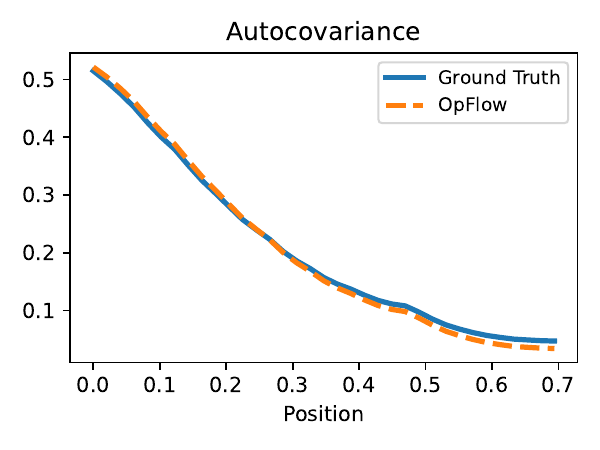}
        \vspace*{-.6cm}
        \caption{}
    \end{subfigure}
    \begin{subfigure}{0.22\textwidth}
        \includegraphics[height=.80\textwidth]{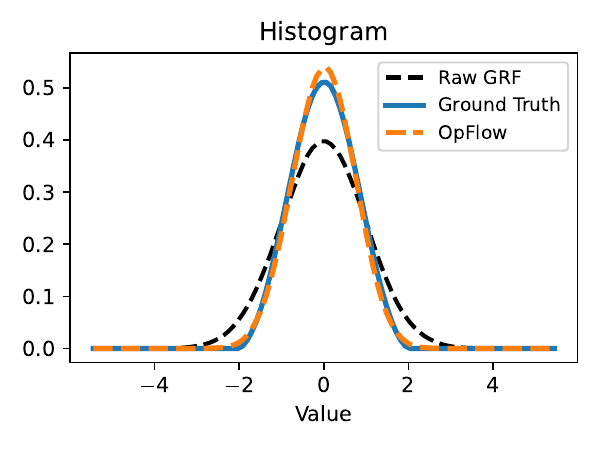}
        \vspace*{-.6cm}
        \caption{}
    \end{subfigure}
    \begin{subfigure}{0.22\textwidth}
        \includegraphics[height=.80\textwidth]{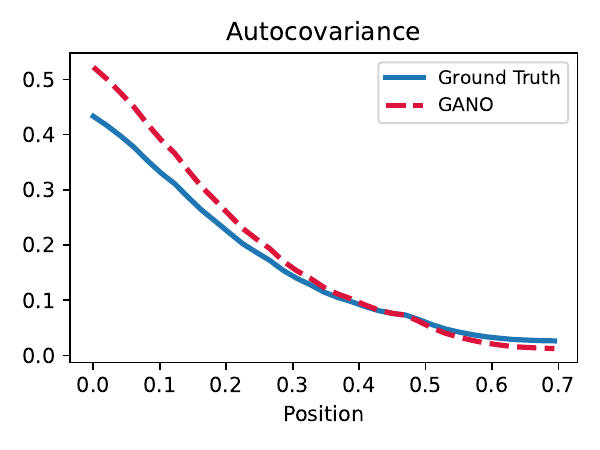}
        \vspace*{-.6cm}
        \caption{}
    \end{subfigure}
    \begin{subfigure}{0.22\textwidth}
        \includegraphics[height=.80\textwidth]{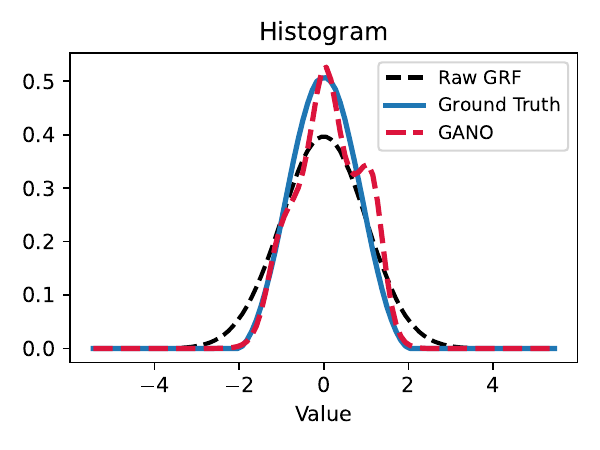}
        \vspace*{-.6cm}
        \caption{}
    \end{subfigure}
    \caption{\alg generation performance on TGRF data. (a) Samples from the ground truth. (b) Samples from \alg. (c) Samples from GANO. (d) Autocovariance of samples from \alg. (e) Histogram of samples from \alg. (f) Autocovariance of samples from GANO. (g) Histogram of samples from GANO.}
    \label{fig:domain_decom_TGRF_gen}
\end{figure*}

\textbf{TGRF data.}
We take the hyperparameters $l_{\mathcal{U}} = 0.5, \nu_{\mathcal{U}}=1.5; l_{\mathcal{A}} = 0.1, \nu_{\mathcal{A}}=0.5$ for generating truncated 2D Gaussian random fields. The distribution is truncated between $[-2, 2]$, with a resolution of $64 \times 64$. From Figure \ref{fig:domain_decom_TGRF_gen}, both \alg and GANO generate high quality samples. In Fig. \ref{fig:domain_decom_TGRF_gen}, we also include the histogram of the GP training data before truncation, which shows that the TGRF data strongly deviates from Gaussian. While GANO exhibits some artifacts in the generated samples, it is worth noting the covariance function and amplitude histogram are learned reasonable well, indicating that the artifacts are only present at very high frequency.
\subsection{Generation with codomain \alg}
For the codomain implementation of \alg, we will focus on super-resolution tasks, which do not suffer from the masking artifacts that occur with the domain decomposed setting. It is often the case that the function of interest is multivariate, i.e. the codomain spans more than one dimension. For instance, in the study of fluid dynamics, velocity fields are often linked with pressure fields. In these cases, there exists a natural partitioning among the co-domain for the affine coupling layers. However, in order to make codomain \alg applicable to univariate functions, 
we first randomly shuffle the training dataset $\{u_j |_D \}_{j=1}^{m}$, and partition the data into two equal parts $ \{u_i|_{D} \}_{i=1}^{m_1} $ and $ \{u'_i|_{D} \}_{i=1}^{m_1} $. We then concatenate these parts along the channel dimension to create a paired bivariate dataset for training, \( \{(u_i, u'_i)|_{D} \}_{i=1}^{m_1} \).

\begin{figure*}[h]
\vspace*{-.2cm}
    \centering
    \hspace{.6cm}
    \begin{subfigure}{0.22\textwidth}
        \includegraphics[height=.7\textwidth]{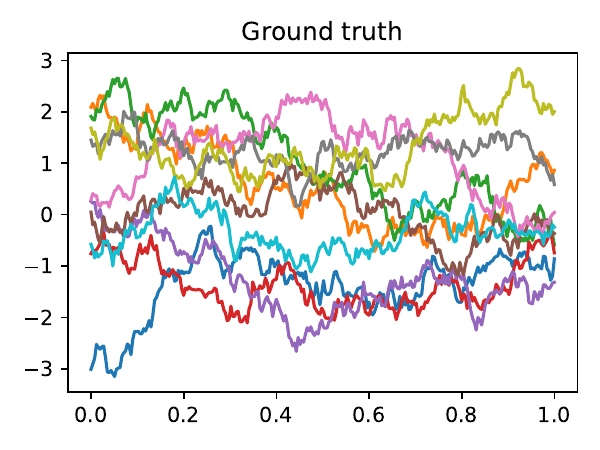}
        \vspace*{-.3cm}
        \caption{}
    \end{subfigure}
    \quad
    \begin{subfigure}{0.22\textwidth}
        \includegraphics[height=.7\textwidth]{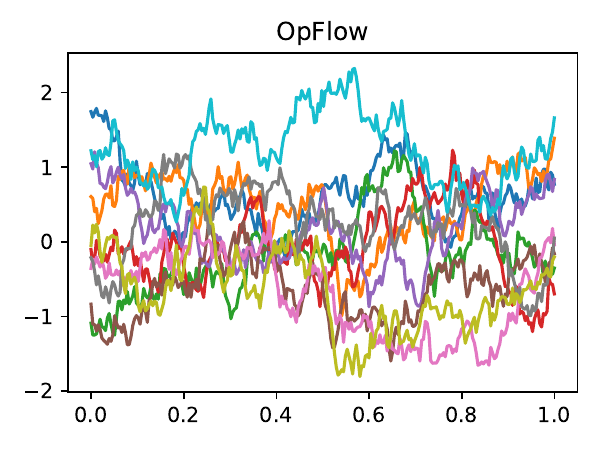}
        \vspace*{-.3cm}
        \caption{}
    \end{subfigure}
    \quad
    \begin{subfigure}{0.22\textwidth}
        \includegraphics[height=.7\textwidth]{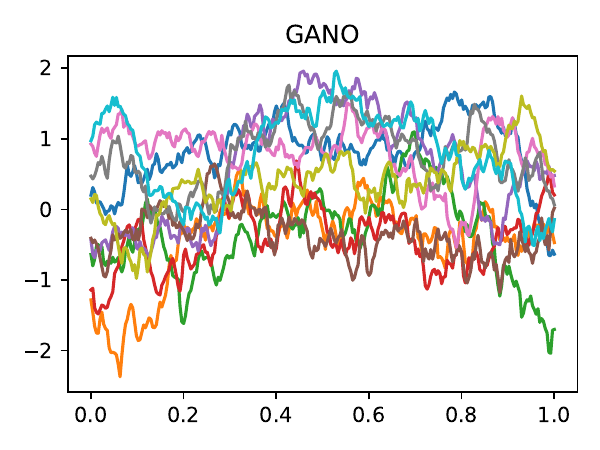}
        \vspace*{-.3cm}
        \caption{}
    \end{subfigure}

    \begin{subfigure}{0.22\textwidth}
        \includegraphics[height=.85\textwidth]{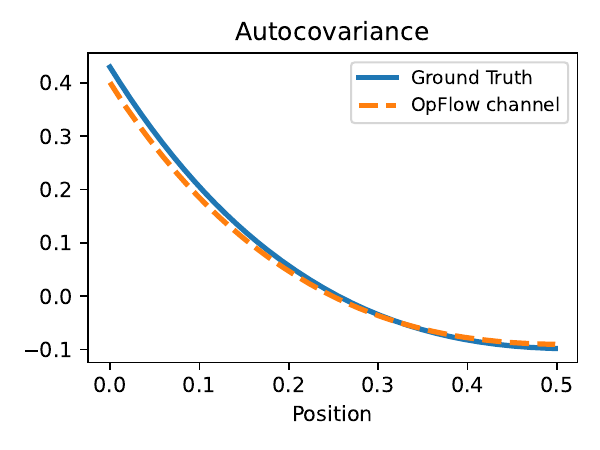}
        \vspace*{-.6cm}
        \caption{}
    \end{subfigure}
    \quad
    \begin{subfigure}{0.22\textwidth}
        \includegraphics[height=.85\textwidth]{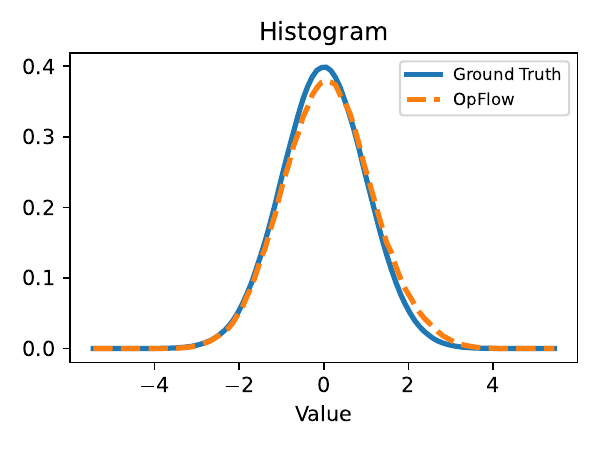}
        \vspace*{-.6cm}
        \caption{}
    \end{subfigure}
    \quad
    \begin{subfigure}{0.22\textwidth}
        \includegraphics[height=.85\textwidth]{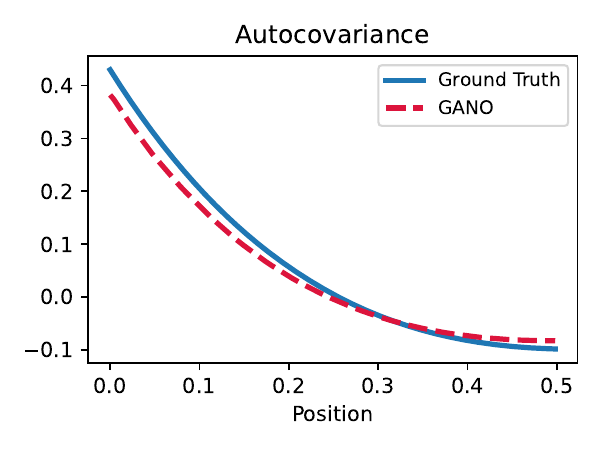}
        \vspace*{-.6cm}
        \caption{}
    \end{subfigure}
    \quad
    \begin{subfigure}{0.22\textwidth}
        \includegraphics[height=.85\textwidth]{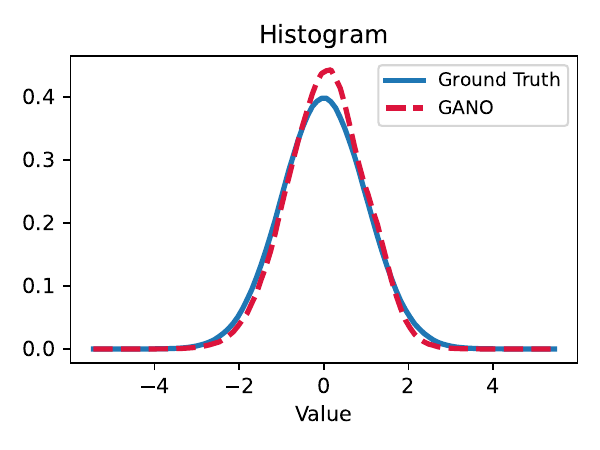}
        \vspace*{-.6cm}
        \caption{}
    \end{subfigure}
    \quad

    \caption{Codomain \alg generation performance on GP data. (a) Samples from the ground truth. (b) Samples from \alg. (c) Samples from GANO. (d) Autocovariance of samples from \alg. (e) Histogram of samples from \alg. (f) Autocovariance of samples from GANO. (g) Histogram of samples from GANO.}
    \label{fig:codomain_GP_gen}
\end{figure*}

\textbf{Codomain GP data.}
We found that the codomain \alg requires different hyperparameters from the domain decomposed setting. Here we set $l_{\mathcal{U}} = 0.5, \nu_{\mathcal{U}}=0.5; l_{\mathcal{A}} = 0.1, \nu_{\mathcal{A}}=0.5$, with a resolution of 256. In particular, these values suggest that codomain \alg has less expressivity than domain decomposed \alg, which we attribute to the need for codomain \alg to learn the joint probability measures, which is hard by nature. Given that both channels convey identical information, we present results from only the first channel for samples generated by codomain \alg. Fig. \ref{fig:codomain_GP_gen} demonstrates that both codomain \alg and GANO effectively produce realistic samples, capturing both the autocovariance function and the histogram distribution accurately. Contrastingly, in the appendix, codomain \alg shows great performance in super-resolution task, potentially due to the channel-split operation preserving the physical domain's integrity. To the best of our knowledge, codomain \alg is the inaugural model within the normalizing flow framework to achieve true resolution invariance. 

\begin{figure*}[h]
\vspace*{-.2cm}
    \centering
    \hspace{-.4cm}
    \begin{subfigure}{0.33\textwidth}
        \includegraphics[height=.25\textwidth,trim={4cm .5cm 4cm .5cm},clip]{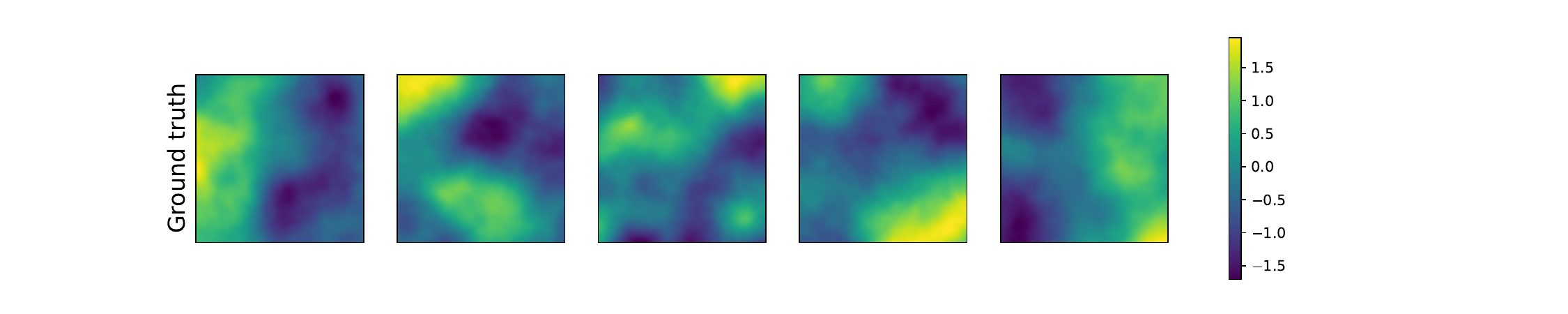}
        \vspace*{-.6cm}
        \caption{}
    \end{subfigure}
    \hspace{-.3cm}
    \begin{subfigure}{0.33\textwidth}
        \includegraphics[height=.25\textwidth,trim={4cm .5cm 4cm .5cm},clip]{Figures_all/GRF2TruncatedGRF_FNO_codomain_64x64_set1_OpFLow_samples_channel1.pdf}
        \vspace*{-.6cm}
        \caption{}
    \end{subfigure}
    \hspace{-.3cm}
    \begin{subfigure}{0.33\textwidth}
        \includegraphics[height=.25\textwidth,trim={4cm .5cm 4cm .5cm},clip]{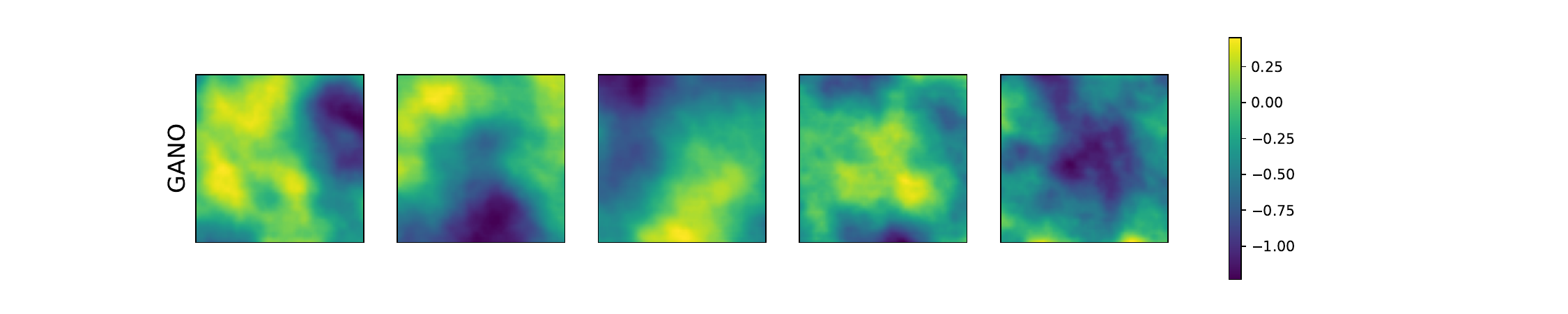}
        \vspace*{-.6cm}
        \caption{}
    \end{subfigure}    
    \hspace{-1.2cm}
    \begin{subfigure}{0.22\textwidth}
        \includegraphics[height=.8\textwidth]{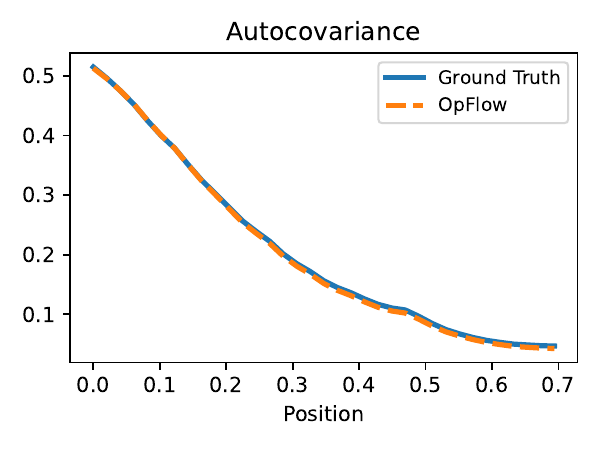}
        \vspace*{-.6cm}
        \caption{}
    \end{subfigure}
    \begin{subfigure}{0.22\textwidth}
        \includegraphics[height=.8\textwidth]{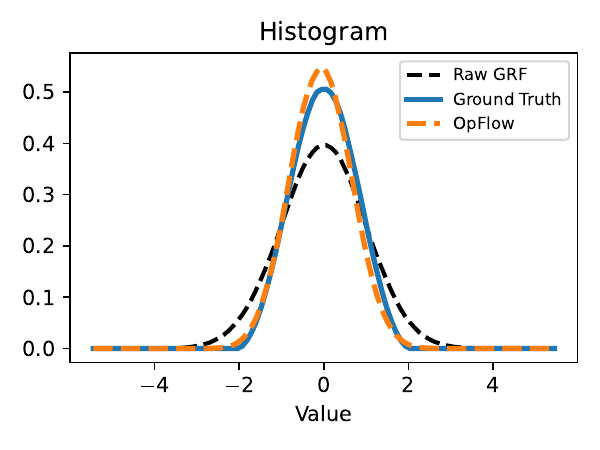}
        \vspace*{-.6cm}
        \caption{}
    \end{subfigure}
    \begin{subfigure}{0.22\textwidth}
        \includegraphics[height=.8\textwidth]{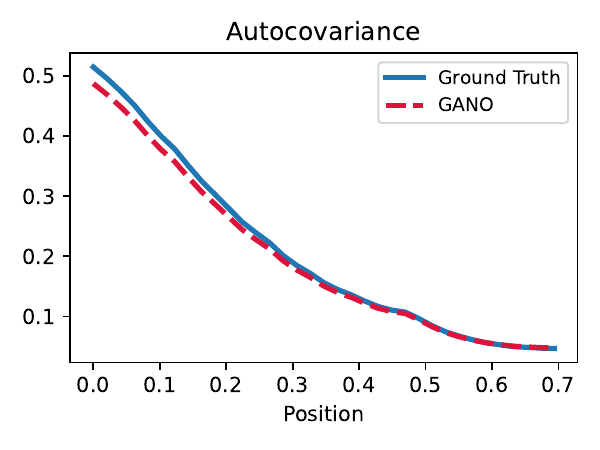}
        \vspace*{-.6cm}
        \caption{}
    \end{subfigure}
    \begin{subfigure}{0.22\textwidth}
        \includegraphics[height=.8\textwidth]{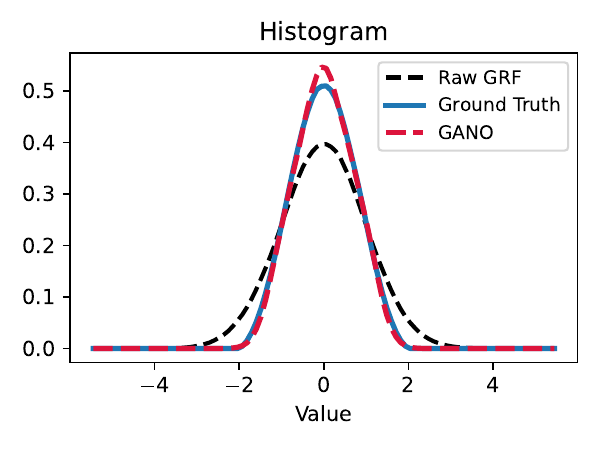}
        \vspace*{-.6cm}
        \caption{}
    \end{subfigure}
    \caption{Codomain \alg generation performance on TGRF data. (a) Samples from the ground truth. (b) Samples from \alg. (c) Samples from GANO. (d) Autocovariance of samples from \alg. (e) Histogram of samples from \alg. (f) Autocovariance of samples from GANO. (g) Histogram of samples from GANO}
    \label{fig:codomain_TGRF_gen}
\end{figure*}

\textbf{Codmain TGRF data.}
We set the hyperparameters to $l_{\mathcal{U}} = 0.5, \nu_{\mathcal{U}}=1.5$ and $l_{\mathcal{A}} = 0.2, \nu_{\mathcal{A}}=1.5$, applying bounds of -2 and 2 to the data, with a resolution of $64 \times 64$. Fig. \ref{fig:codomain_TGRF_gen} demonstrates that, under these parameters, both \alg and GANO accurately capture the autocovariance function and histogram distribution. This experiment underscores codomain \alg's capability in effectively handling non-Gaussian 2D data. Additionally, super-resolution results included in the appendix reveal that codomain \alg is free from artifacts during super-resolution tasks, showcasing its robustness. 

\subsection{Zero-shot super-resolution generation experiments of \alg}

\begin{figure*}[h]
    \centering
    \hspace{-.4cm}
    \quad
    \begin{subfigure}{0.22\textwidth}
        \includegraphics[height=.82\textwidth]{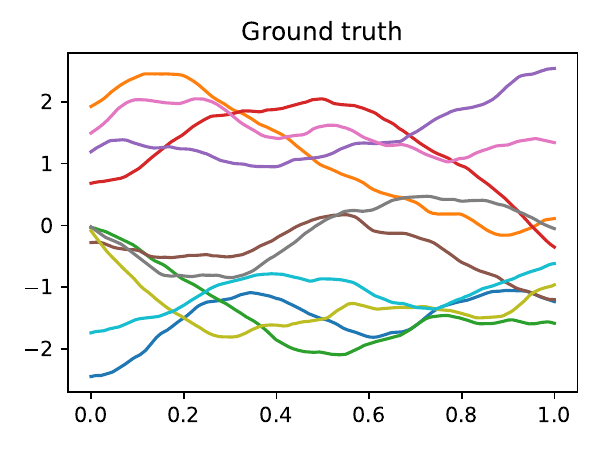}
        \vspace*{-.6cm}
        \caption{}
    \end{subfigure}
    \quad
    \begin{subfigure}{0.22\textwidth}
        \includegraphics[height=.82\textwidth]{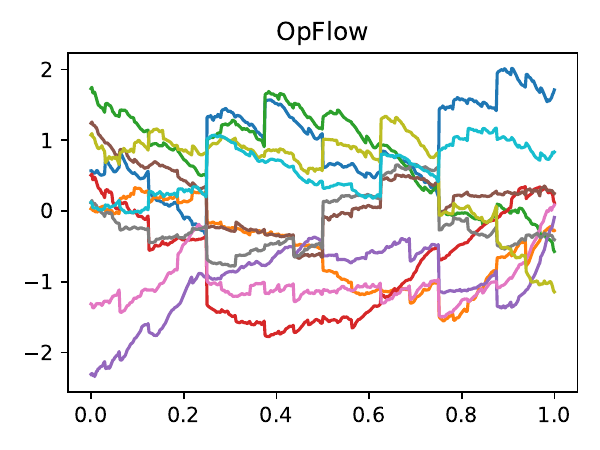}
        \vspace*{-.6cm}
        \caption{}
    \end{subfigure}
    \quad
    \begin{subfigure}{0.22\textwidth}
        \includegraphics[height=.82\textwidth,trim={0 .4cm 0 0},clip]{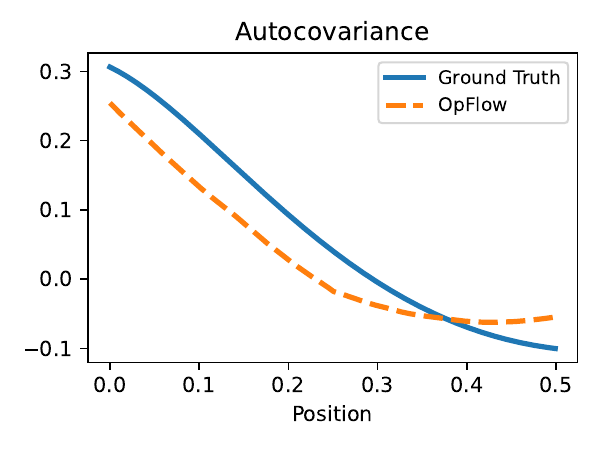}
        \vspace*{-.6cm}
        \caption{}
    \end{subfigure}
    \quad
    \begin{subfigure}{0.22\textwidth}
        \includegraphics[height=.82\textwidth,trim={0 .4cm 0 0},clip]{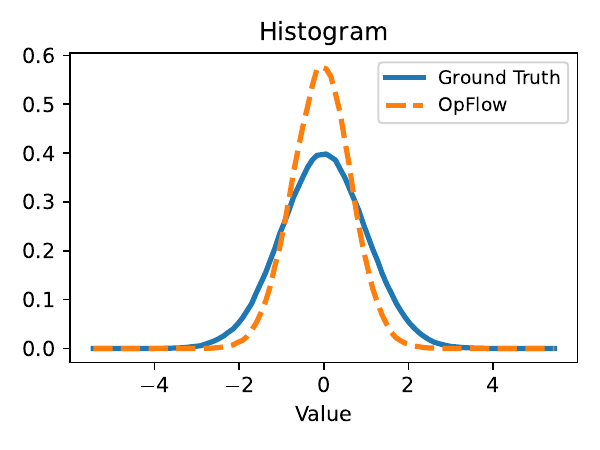}
        \vspace*{-.6cm}
        \caption{}
    \end{subfigure}
    \quad

    \caption{Domain decomposed \alg super-resolution experiment on generating GP data with resolution of 512. (a) Samples from the ground truth. (b) Samples from \alg. (c) Autocovariance of samples from \alg. (d) Histogram of samples from \alg }
    
    \label{fig:domain_decom_GP_gen_sup}
\end{figure*}

Here we present zero-shot super-resolution results with domain-decomposed \alg and codomain \alg, corresponding to the experiments detailed in the \textit{Generation Tasks} section. All statistics, such as the autocovariance function and the histogram of point-wise values, were calculated from 5,000 realizations. Fig. \ref{fig:domain_decom_GP_gen_sup} depicts the domain-decomposed \alg, which was trained at a resolution of 256 and then evaluated at a resolution of 512. In Fig. \ref{fig:domain_decom_GP_gen_sup}(a) and (b), the super-resolution samples from domain-decomposed \alg display a jagged pattern and show high discontinuities, Additionally, Fig. \ref{fig:domain_decom_GP_gen_sup}(c) and (d) show that both the autocovariance function and the histogram function evaluated from the domain-decomposed \alg diverge from the ground truths. We attribute these artifacts to the domain decomposition operation, which splits the domain into smaller patches and can result in a loss of derivative continuity at the edges of these patches. This artifact is analogous to the jagged artifacts in vision transformers (ViTs), caused by the discontinuous patch-wise tokenization process~\citep{dosovitskiy_image_2021, qian_blending_2021}. 

\begin{figure*}[h]
    \centering
    \hspace{-.4cm}
    \quad
    \begin{subfigure}{0.22\textwidth}
        \includegraphics[height=.82\textwidth]{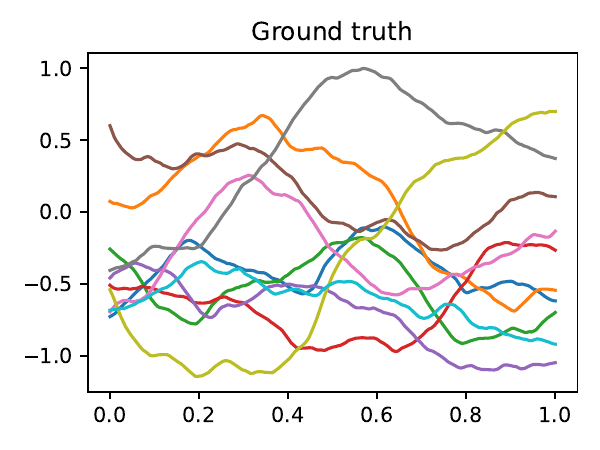}
        \vspace*{-.6cm}
        \caption{}
    \end{subfigure}
    \quad
    \begin{subfigure}{0.22\textwidth}
        \includegraphics[height=.82\textwidth]{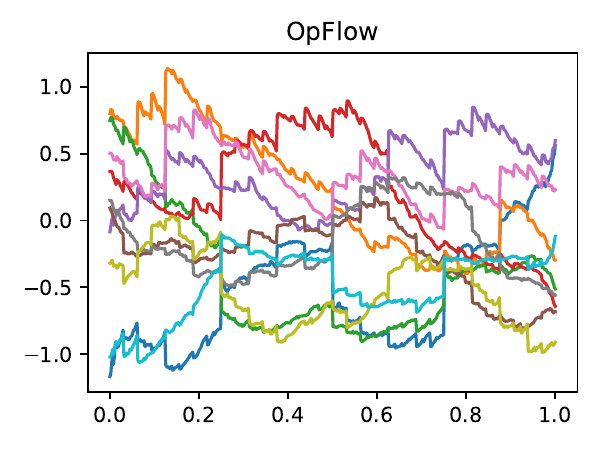}
        \vspace*{-.6cm}
        \caption{}
    \end{subfigure}
    \quad
    \begin{subfigure}{0.22\textwidth}
        \includegraphics[height=.82\textwidth,trim={0 .4cm 0 0},clip]{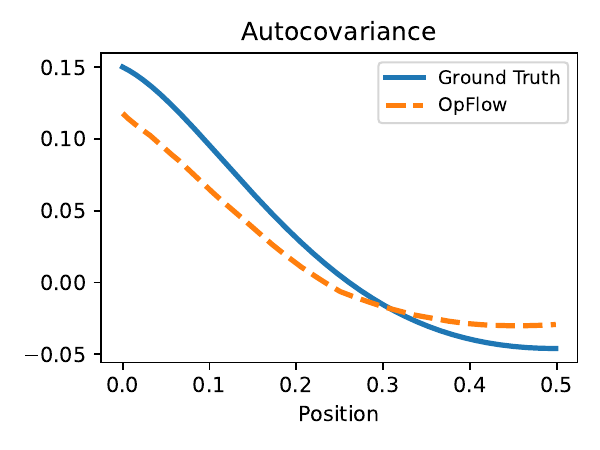}
        \vspace*{-.6cm}
        \caption{}
    \end{subfigure}
    \quad
    \begin{subfigure}{0.22\textwidth}
        \includegraphics[height=.82\textwidth,trim={0 .4cm 0 0},clip]{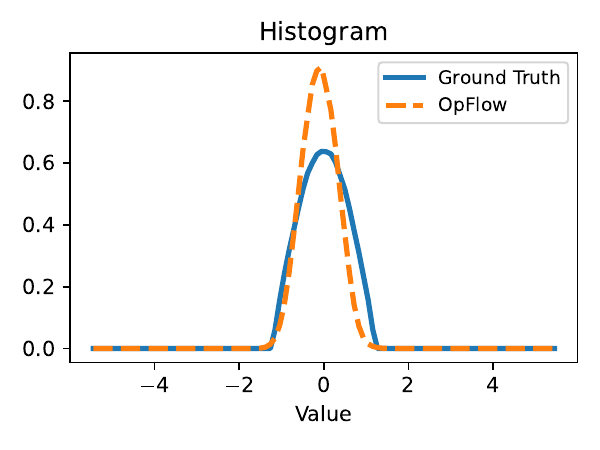}
        \vspace*{-.6cm}
        \caption{}
    \end{subfigure}
    \quad

    \caption{Domain decomposed \alg super-resolution experiment on generating TGP data with resolution of 512. (a) Samples from the ground truth. (b) Samples from \alg. (c) Autocovariance of samples from \alg. (d) Histogram of samples from \alg}
    
    \label{fig:domain_decom_TGP_gen_sup}
\end{figure*}

\begin{figure*}[h]
    \centering
    \begin{subfigure}{0.48\textwidth}
        \includegraphics[height=.25\textwidth]{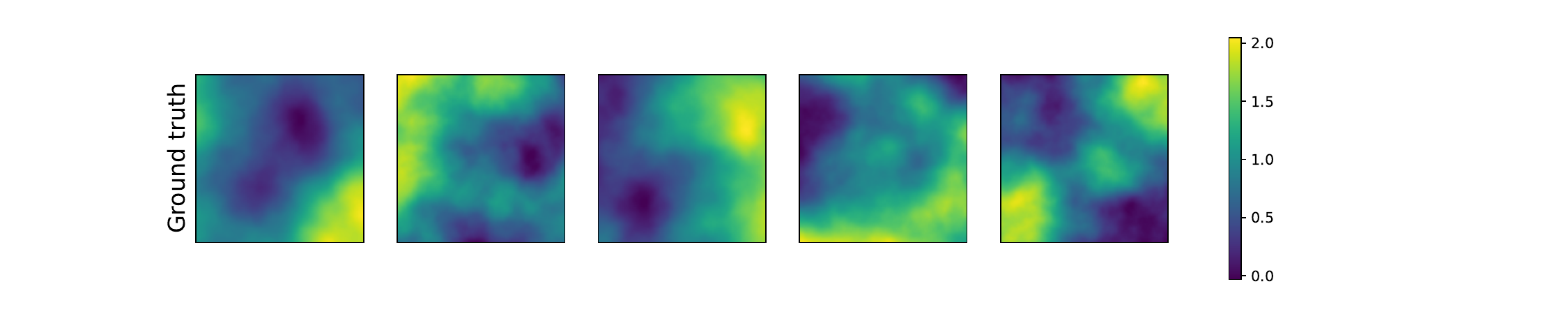}
        \vspace*{-.8cm}
        \caption{}
    \end{subfigure}
    \quad
    \begin{subfigure}{0.48\textwidth}
        \includegraphics[height=.25\textwidth]{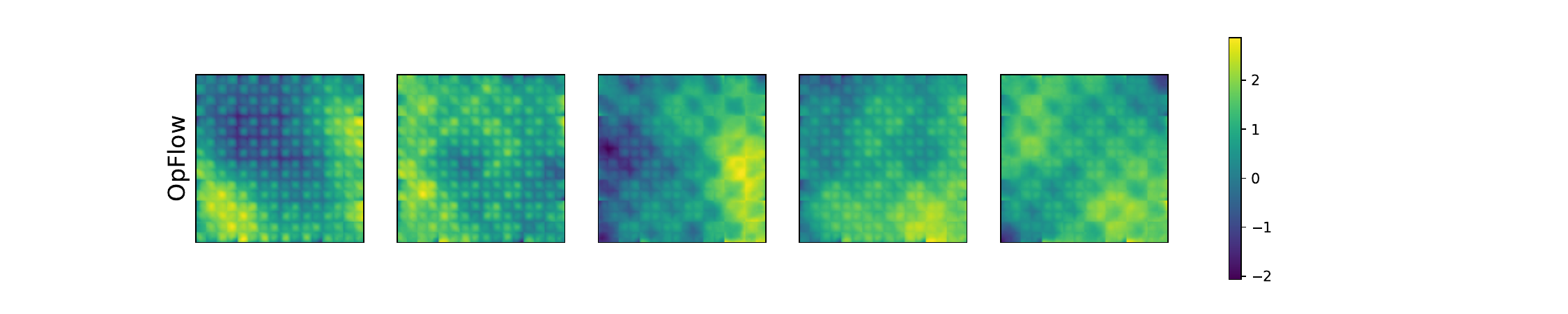}
        \vspace*{-.8cm}
        \caption{}
    \end{subfigure}

    \begin{subfigure}{0.22\textwidth}
        \includegraphics[height=.85\textwidth]{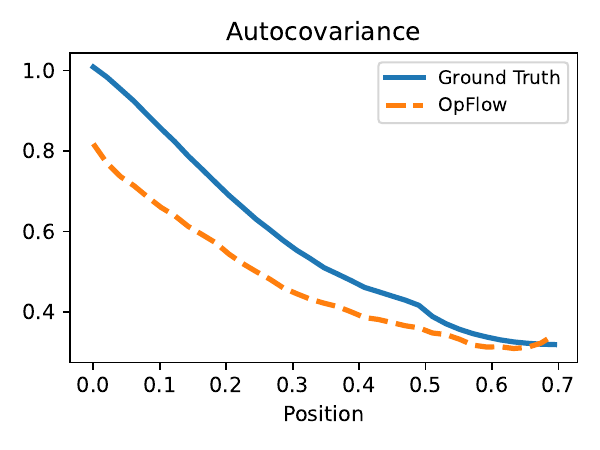}
        \vspace*{-.8cm}
        \caption{}
    \end{subfigure}
    \quad
    \begin{subfigure}{0.22\textwidth}
        \includegraphics[height=.85\textwidth]{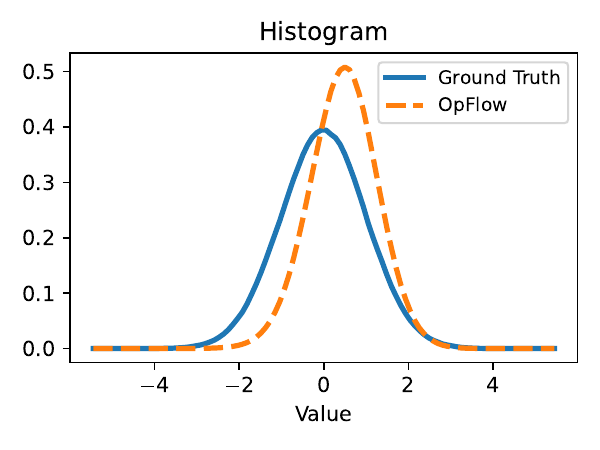}
        \vspace*{-.8cm}
        \caption{}
    \end{subfigure}
    \quad

    \caption{Domain decomposed \alg super-resolution experiment on generating GRF data with resolution of 128$\times$128. (b) Samples from groud truth. (c) Samples from \alg. (d) Autocovariance of samples from \alg. (e) Histogram of samples from \alg}
    \label{fig:domain_decom_GRF_gen_sup}
\end{figure*}

The super-resolution study was extended to non-Gaussian data, with results illustrated in Fig. \ref{fig:domain_decom_TGP_gen_sup}. Analysis of Fig. \ref{fig:domain_decom_TGP_gen_sup} leads us to a similar conclusion that the domain-decomposition operation introduces artifacts into the super-resolution outputs of \alg. This artifact is also evident in the GRF super-resolution experiment shown in Fig. \ref{fig:domain_decom_GRF_gen_sup}. Here, the domain-decomposed \alg was trained at a resolution of \(64\times64\) and evaluated at \(128\times128\), resulting in observable checkerboard patterns in the 2D super-resolution outcomes. Furthermore, the statistical analysis of the samples generated by the domain-decomposed \alg, as depicted in Fig. \ref{fig:domain_decom_GRF_gen_sup}, reveals a significant divergence from the ground truth.

\begin{figure*}[h]
    \centering
    \hspace{-.4cm}
    \quad
    \begin{subfigure}{0.22\textwidth}
        \includegraphics[height=.82\textwidth]{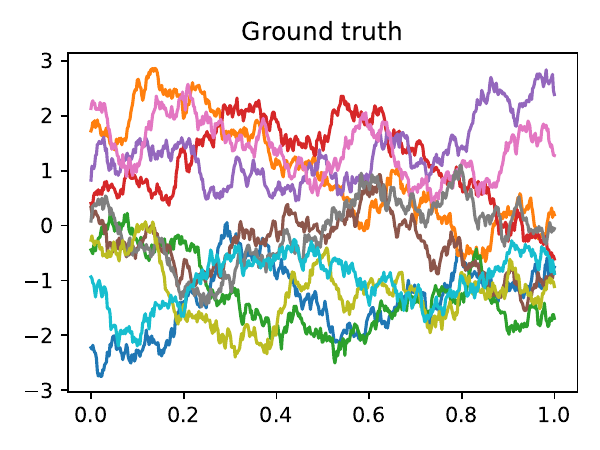}
        \vspace*{-.6cm}
        \caption{}
    \end{subfigure}
    \quad
    \begin{subfigure}{0.22\textwidth}
        \includegraphics[height=.82\textwidth]{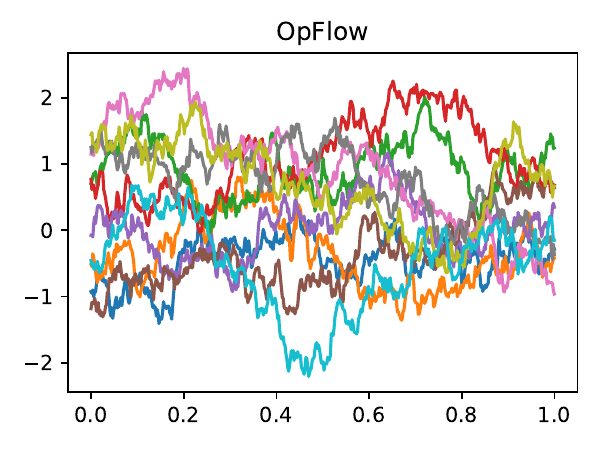}
        \vspace*{-.6cm}
        \caption{}
    \end{subfigure}
    \quad
    \begin{subfigure}{0.22\textwidth}
        \includegraphics[height=.82\textwidth,trim={0 .4cm 0 0},clip]{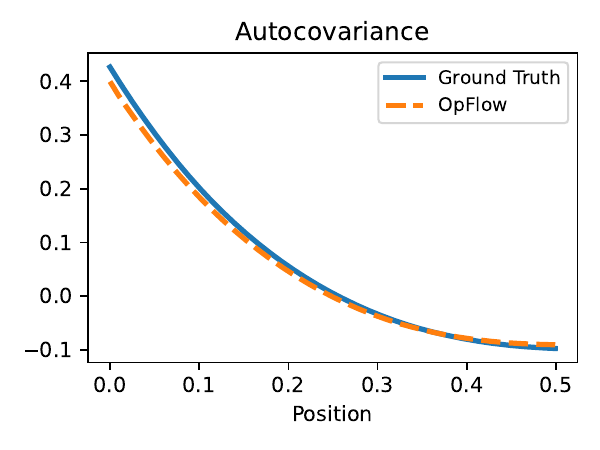}
        \vspace*{-.6cm}
        \caption{}
    \end{subfigure}
    \quad
    \begin{subfigure}{0.22\textwidth}
        \includegraphics[height=.82\textwidth,trim={0 .4cm 0 0},clip]{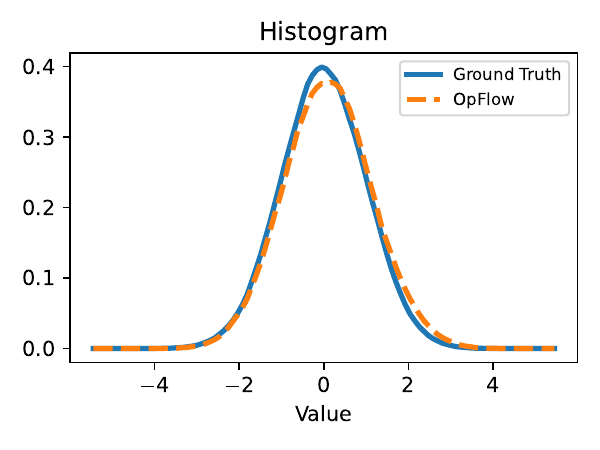}
        \vspace*{-.6cm}
        \caption{}
    \end{subfigure}
    \quad

    \caption{Codomain \alg super-resolution experiment on generating GP data with resolution of 512. (a) Samples from the ground truth. (b) Samples from \alg. (c) Autocovariance of samples from \alg. (d) Histogram of samples from \alg }
    
    \label{fig:codomain_decom_GP_gen_sup}
\end{figure*}

\begin{figure*}[h]
\vspace*{-.2cm}
    \centering
    \begin{subfigure}{0.48\textwidth}
        \includegraphics[height=.25\textwidth]{Figures_all/GRF2TruncatedGRF_FNO_codomain_64x64_set1_sample_ground_truth.pdf}
        \vspace*{-.6cm}
        \caption{}
    \end{subfigure}
    \quad
    \begin{subfigure}{0.48\textwidth}
        \includegraphics[height=.25\textwidth]{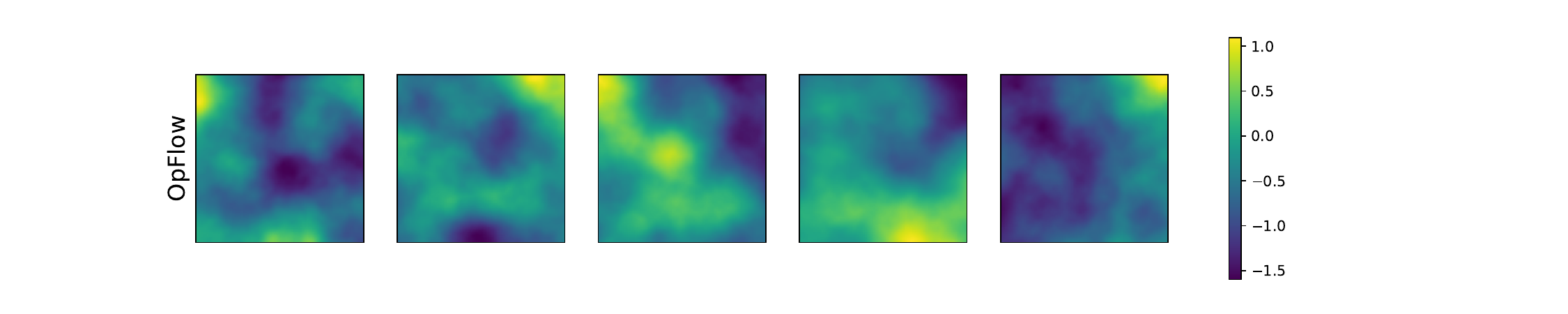}
        \vspace*{-.6cm}
        \caption{}
    \end{subfigure}

    \begin{subfigure}{0.22\textwidth}
        \includegraphics[height=.85\textwidth]{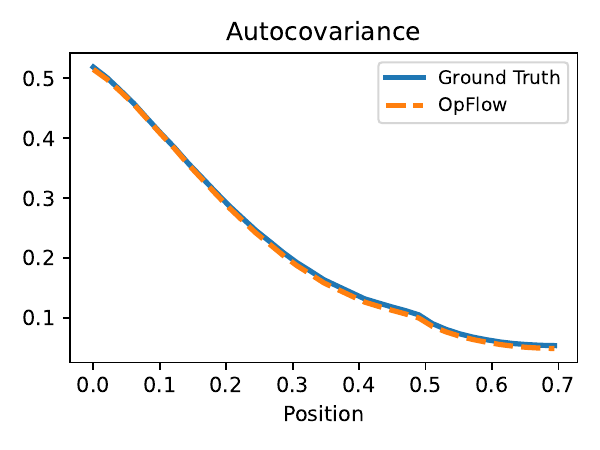}
        \vspace*{-.6cm}
        \caption{}
    \end{subfigure}
    \quad
    \begin{subfigure}{0.22\textwidth}
        \includegraphics[height=.85\textwidth]{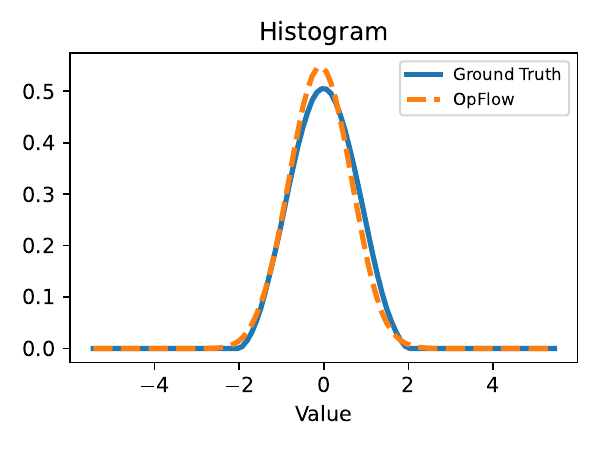}
        \vspace*{-.6cm}
        \caption{}
    \end{subfigure}
    \quad

    \caption{Codomain \alg super-resolution experiment on generating TGRF data with resolution of 128$\times$128. (b) Samples from groud truth. (c) Samples from \alg. (d) Autocovariance of samples from \alg. (e) Histogram of samples from \alg}
    \label{fig:codomain_decom_GRF_gen_sup}
\end{figure*}

However, in the super-resolution tasks using codomain \alg for both GP and non-GP experiments, as shown in Fig. \ref{fig:codomain_decom_GP_gen_sup} and \ref{fig:codomain_decom_GRF_gen_sup}, the jagged or checkerboard artifacts of zero-shot super-resolution performance in \alg are gone. This improvement is attributed to codomain \alg maintaining the integrity of the entire input domain. Yet, this advantage does come with its own set of challenges, notably an increased difficulty in the learning process, which requires more intensive tuning efforts and may yield results that are less expressive when compared to domain-decomposed \alg. Consequently, there exists a trade-off between stronger learning capability and achieving true resolution invariance when we use different partitioning methods of \alg.

\subsection{Ablation study}\label{apx:ablation}
\begin{figure*}[h]
\vspace*{-.2cm}
    \centering
    \begin{subfigure}{0.48\textwidth}
        \includegraphics[height=.25\textwidth]{Figures_all/GRF2GRF_FNO_domain_decomposed_64x64_ground_truth.pdf}
        \vspace*{-.8cm}
        \caption{}
    \end{subfigure}
    \quad
    \begin{subfigure}{0.48\textwidth}
        \includegraphics[height=.25\textwidth]{Figures_all/GRF2GRF_FNO_domain_decomposed_64x64_OpFLow_samples.pdf}
        \vspace*{-.8cm}
        \caption{}
    \end{subfigure}
    \quad
    \begin{subfigure}{0.48\textwidth}
        \includegraphics[height=.25\textwidth]{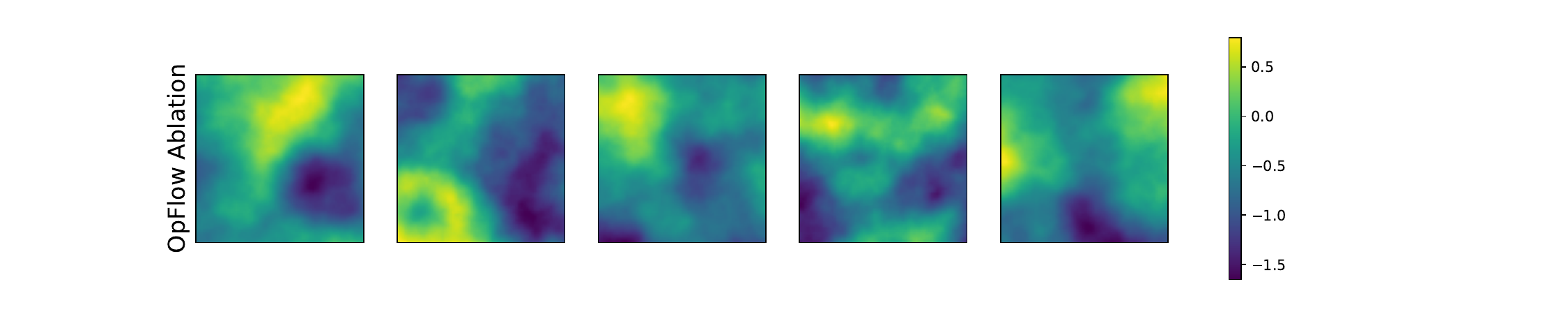}
        \vspace*{-.8cm}
        \caption{}
    \end{subfigure}

    \begin{subfigure}{0.22\textwidth}
        \includegraphics[height=.82\textwidth]{Figures_all/GRF2GRF_FNO_domain_decomposed_64x64_autocov.pdf}
        \vspace*{-.8cm}
        \caption{}
    \end{subfigure}
    \quad
    \begin{subfigure}{0.22\textwidth}
        \includegraphics[height=.82\textwidth]{Figures_all/GRF2GRF_FNO_domain_decomposed_64x64_statistic.pdf}
        \vspace*{-.8cm}
        \caption{}
    \end{subfigure}
    \quad
    \begin{subfigure}{0.22\textwidth}
        \includegraphics[height=.82\textwidth]{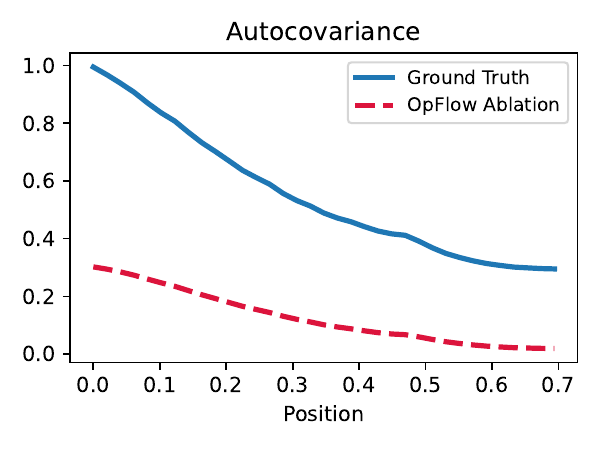}
        \vspace*{-.8cm}
        \caption{}
    \end{subfigure}
    \quad
    \begin{subfigure}{0.22\textwidth}
        \includegraphics[height=.82\textwidth]{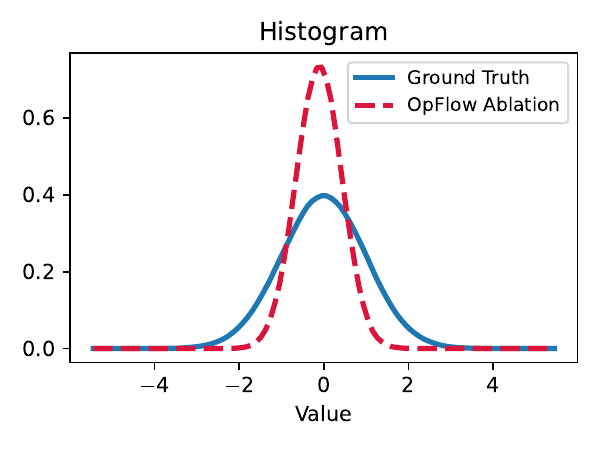}
        \vspace*{-.8cm}
        \caption{}
    \end{subfigure}
    \quad

    \caption{Ablation study of \alg generation performance on GRF data (a) samples from the ground truth. (b) Samples from \alg. (c) Samples from \alg-ablation (\alg trained without regularization). (d) Autocovariance of samples from \alg. (e) Histogram of samples from \alg. (f) Autocovariance of samples from \alg-ablation. (g) Histogram of samples from \alg-ablation}
    \label{fig:ablation}
\end{figure*}

In this ablation study, we investigate the role of 2-Wasserstein ($W_2$) regularization on the performance of the \alg, as detailed in Eq. \ref{eq:firststep}. We set the hyperparameters to $l_{\mathcal{U}} = 0.5, \nu_{\mathcal{U}}=1.5$ for the observed space and $l_{\mathcal{A}} = 0.1, \nu_{\mathcal{A}}=0.5$ for the Gaussian Process (GP) defined in the latent space $\mathcal{A}$, with a chosen resolution of $64 \times 64$. The domain-decomposed \alg model was trained under two conditions: with and without the incorporation of $W_2$ regularization during the warmup phase. According to the results depicted in Fig. \ref{fig:ablation}, incorporating $W_2$ regularization significantly enhanced \alg's ability to accurately learn the statistical properties of the input function data. In contrast, the model trained without $W_2$ regularization demonstrated a notable deficiency in capturing the same statistics, underscoring the critical role of $W_2$ regularization in the model's performance
\subsection{Size of training dataset for learning a prior}\label{Apx:small_set}

In this section, we revisit the 1D GP regression experiment with \alg to assess the minimal dataset size required to train an effective prior. We reduced the size of the training datasets to be substantially fewer than the original count of 30000 samples. Specifically, we conducted tests using datasets of 3000, 6000, and 9000 samples to examine their impact on the regression performance.

\begin{figure*}
\vspace*{-.2cm}
    \centering
    \begin{subfigure}{0.23\textwidth}
        \includegraphics[height=.73\textwidth]{Figures_all/GP2GP_domain_decomposed_gt.pdf}
        \vspace*{-.6cm}
        \caption{}
    \end{subfigure}
    \begin{subfigure}{0.23\textwidth}
        \includegraphics[height=.73\textwidth]{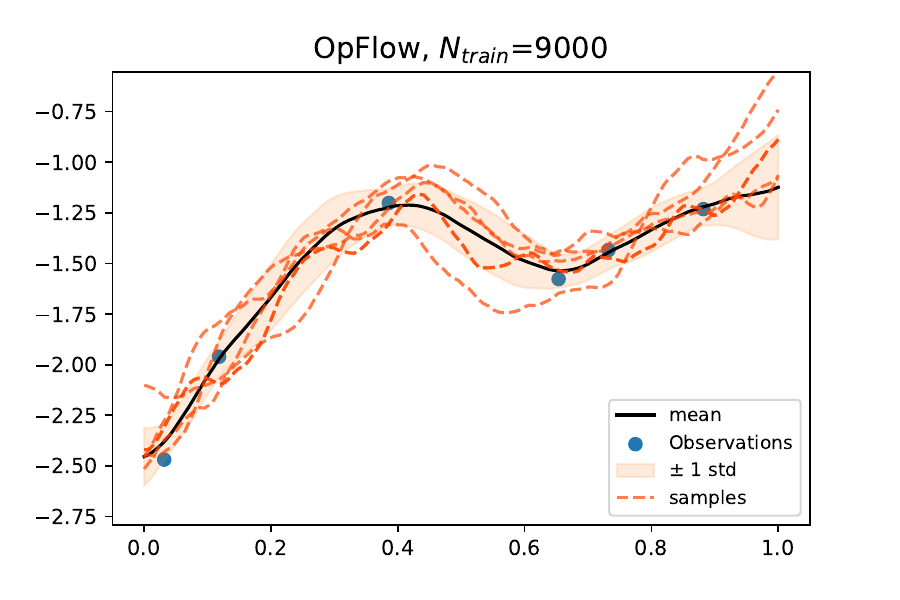}
        \vspace*{-.6cm}
        \caption{}
    \end{subfigure}
    \begin{subfigure}{0.23\textwidth}
        \includegraphics[height=.73\textwidth]{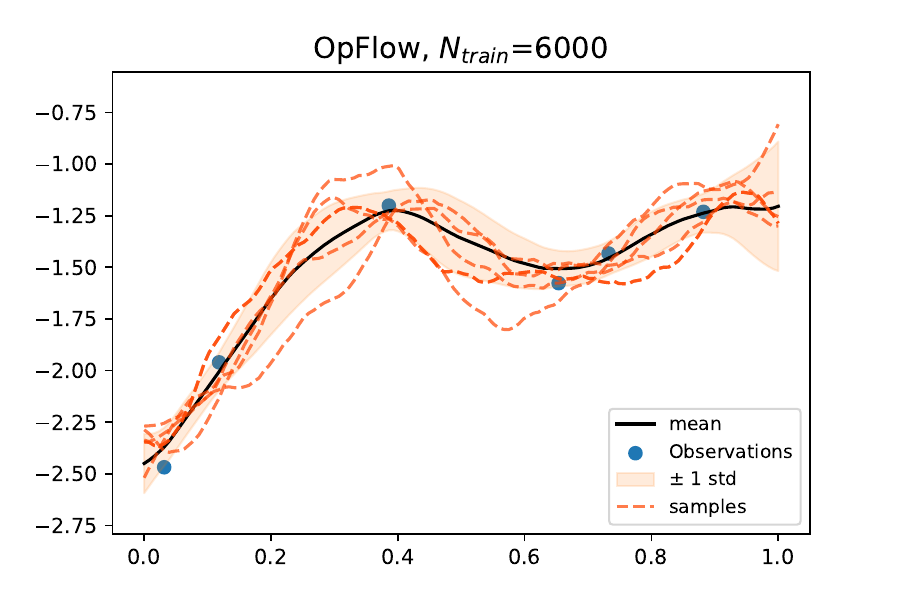}
        \vspace*{-.6cm}
        \caption{}
    \end{subfigure}
    \begin{subfigure}{0.23\textwidth}
        \includegraphics[height=.73\textwidth]{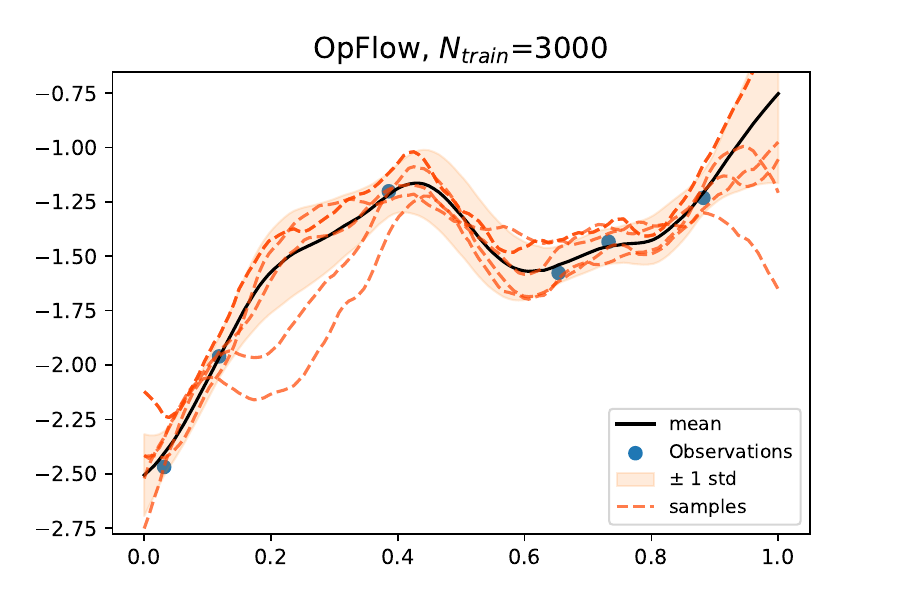}
        \vspace*{-.6cm}
        \caption{}
    \end{subfigure}
    \begin{subfigure}{0.23\textwidth}
        \includegraphics[height=.73\textwidth]{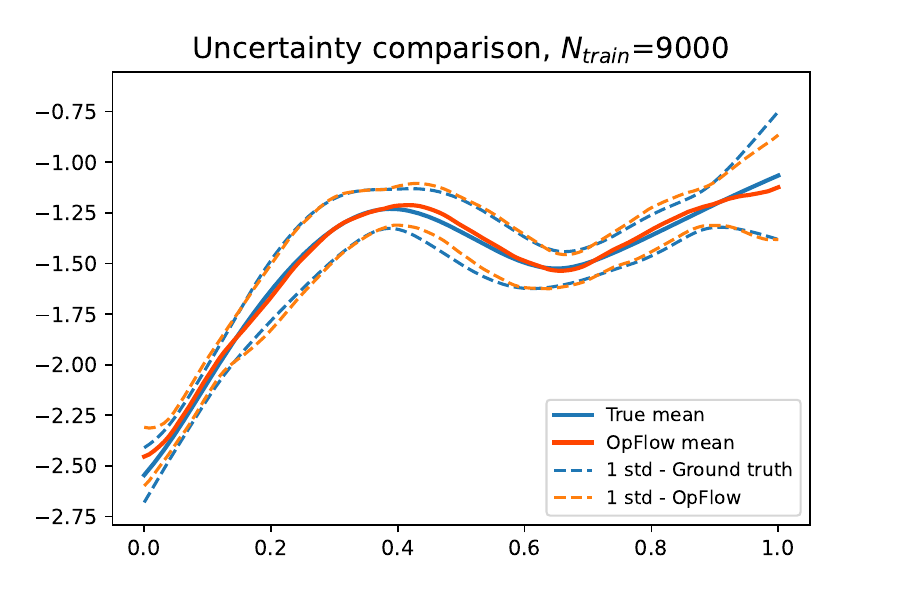}
        \vspace*{-.6cm}
        \caption{}
    \end{subfigure}
    \begin{subfigure}{0.23\textwidth}
        \includegraphics[height=.73\textwidth]{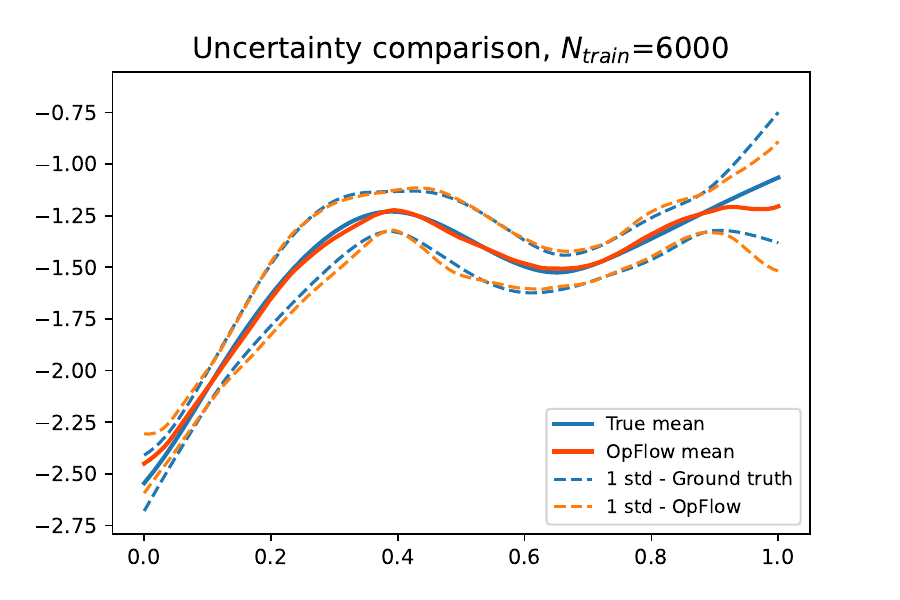}
        \vspace*{-.6cm}
        \caption{}
    \end{subfigure}
    \begin{subfigure}{0.23\textwidth}
        \includegraphics[height=.73\textwidth]{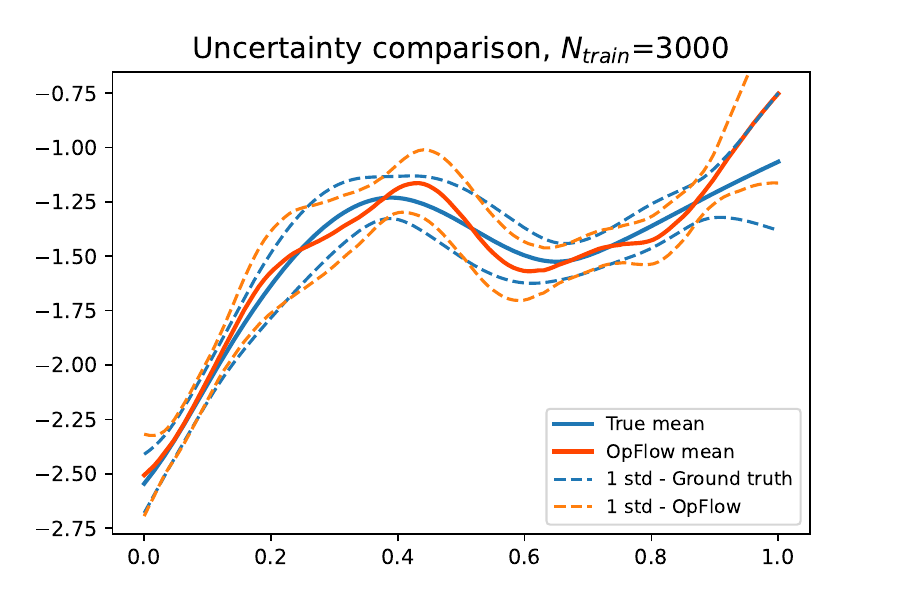}
        \vspace*{-.6cm}
        \caption{}
    \end{subfigure}
    \caption{ \alg regression on GP data with varying sizes of training dataset. (a) Ground truth GP regression with observed data and predicted samples (b) \alg regression with observed data and predicted samples based on the prior trained with 9000 samples. (c) \alg regression with observed data and predicted samples based on the prior trained with 6000 samples. (d) \alg regression with observed data and predicted samples based on the prior trained with 3000 samples. (e) Uncertainty comparison between true GP and \alg predictions based on the prior trained with 9000 samples. (f) Uncertainty comparison between true GP and \alg predictions based on the prior trained with 6000 samples. (g) Uncertainty comparison between true GP and \alg predictions based on the prior trained with 3000 samples.}
    \label{fig:prior_size}
\end{figure*}

As depicted in Fig. \ref{fig:prior_size}, the prior trained with 3000 samples showed reasonable performance. Although there was an observable increase in the error of the mean prediction, the uncertainty was correctly captured. This demonstrates that even a substantially reduced dataset size retains predictive utility. For the dataset of 6000 samples, we observed that the regression performance closely aligned with the ground truth and matched the performance of the prior trained with 30000 samples shown in the main text. This finding highlights that a training dataset size of 6000 is nearly as effective as one five times larger, significantly reducing data-collection demand while maintaining high fidelity in the predictions. Moreover, increasing the dataset to 9000 samples further enhanced both the accuracy and the precision of uncertainty estimation, closely approximating the performance achieved with the original dataset size of 30000.

These results underscore the point that while larger datasets can enhance \alg's performance, comparable performances are achievable with much smaller datasets. In the context of GP regression with \alg, training datasets ranging from 3000 to 6000 samples, which is about 10\% to 20\% of the original training dataset size, should suffice to train a robust and effective prior. Last, it's worth exploring using an untrained \alg as a prior for estimating the posterior distribution of specific observations, where no training data is required. This approach is known as Deep Probabilistic Imaging~\citep{sun_deep_2020}. Utilizing \alg in this manner may significantly broaden its applicability, particularly in scenarios where collecting training data is challenging~\citep{aigrain_gaussian_2023}.

\subsection{Regression with codomain \alg}
\begin{figure*}[h]
    \centering
    \hspace{-.4cm}
    \begin{subfigure}{0.3\textwidth}
        \includegraphics[height=.73\textwidth]{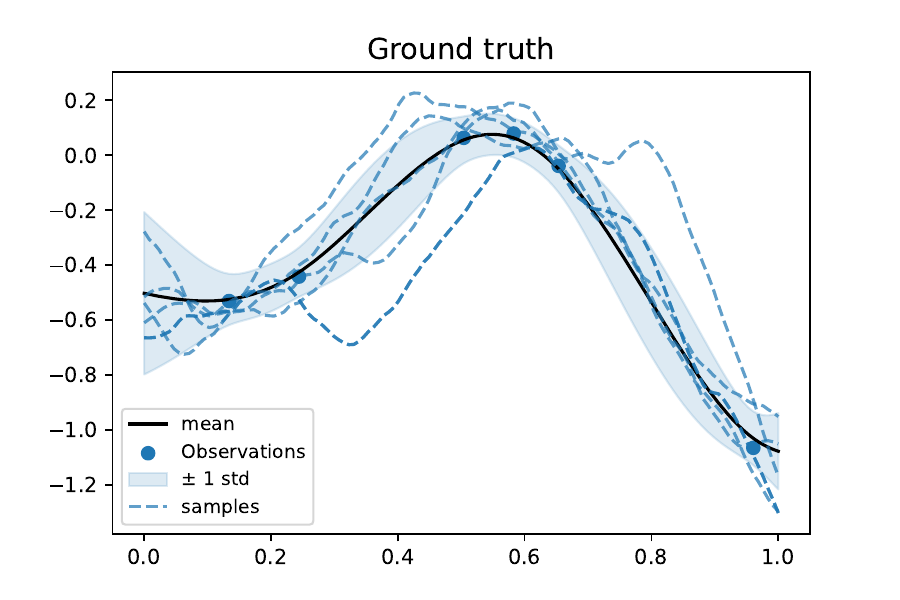}
        \vspace*{-.6cm}
        \caption{}
    \end{subfigure}
    \quad
    \begin{subfigure}{0.3\textwidth}
        \includegraphics[height=.73\textwidth]{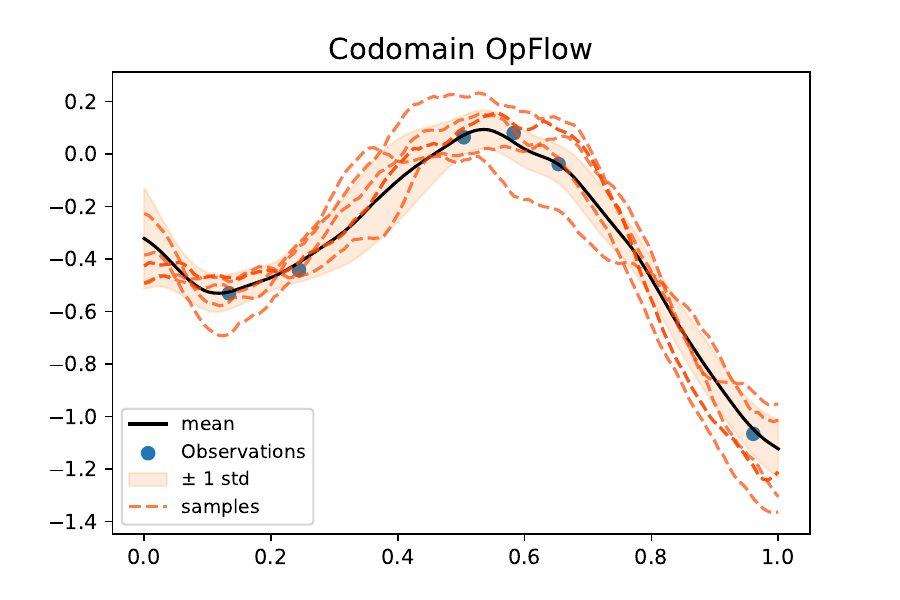}
        \vspace*{-.6cm}
        \caption{}
    \end{subfigure}
    \quad
    \begin{subfigure}{0.3\textwidth}
        \includegraphics[height=.73\textwidth]{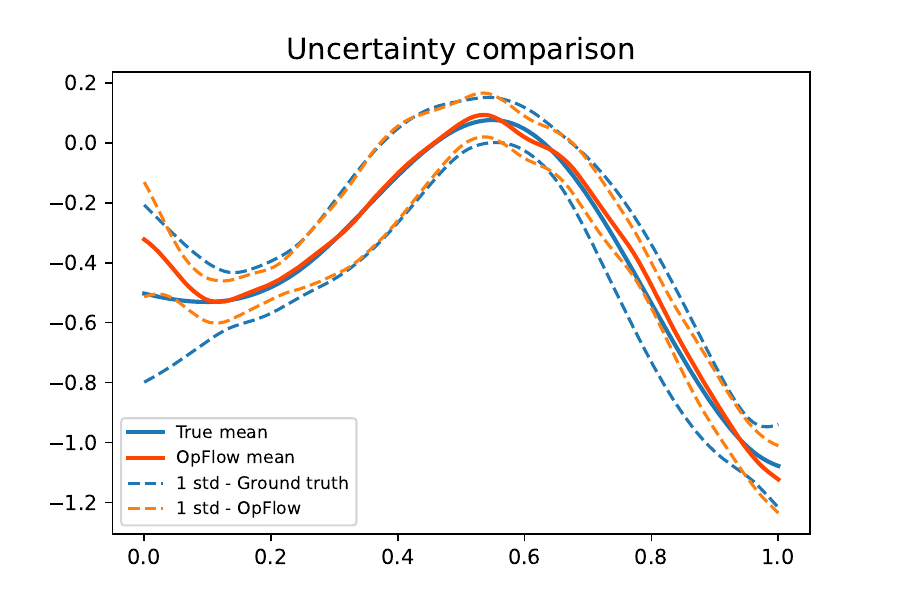}
        \vspace*{-.6cm}
        \caption{}
    \end{subfigure}
    \caption{ Codomain \alg regression on GP data. (a) Ground truth GP regression with observed data and predicted samples. (b) OpFlow regression with observed data and predicted samples. (c) Uncertainty comparison between true GP and OpFlow predictions.}
    \label{fig:GP_codomain_opflow_reg}
\end{figure*}

For the training of the codomain \alg prior, we selected hyperparameters $l_{\mathcal{U}} = 0.5, \nu_{\mathcal{U}}=1.5$ for the observed space and $l_{\mathcal{A}} = 0.2, \nu_{\mathcal{A}}=1.5$ for the Gaussian Process (GP) in the latent space. In our regression task, a new realization of $u \in \mathcal{U}$ was generated, with observations taken at six randomly chosen positions within the domain $\mathcal{D}$. The goal was to infer the posterior distribution of $u$ across 128 positions. Further details on regression parameters available in Table \ref{tab:reg_table}. Additionally, we only show the regression results from the first channel of \alg generated samples. As depicted in Fig. \ref{fig:GP_codomain_opflow_reg}, codomain \alg accurately captures both the mean and uncertainty of the true posterior, as well as generating realistic posterior samples that resemble true posterior samples from GPR. However, as discussed in the previous sections, codomain \alg is not as effective as domain-decomposed \alg for learning the prior, which highly affects the regression performance and may account for the underestimated uncertainty after $x=0.5$ on the x-axis. This experiment confirms codomain \alg can also be used for UFR, which demonstrates the robustness of UFR performance across different \alg architectures.

\end{document}